\pdfoutput=1

\documentclass[11pt]{article}

\usepackage[]{naacl2021}

\usepackage{times}
\usepackage{latexsym}
\usepackage{graphicx}
\usepackage{amsmath} 
\usepackage{amsfonts}
\usepackage{verbatim}
\usepackage{caption}
\usepackage{subcaption}
\usepackage{float}
\usepackage{multirow}
\usepackage[T1]{fontenc}
\usepackage{chngcntr}

\usepackage[utf8]{inputenc}

\usepackage{booktabs}

\usepackage{microtype}

\usepackage{todonotes}
\newcommand{\Note}[2]{}
\renewcommand{\Note}[2]{\todo[color=#1,size=\small, inline=true]{#2}} 

\title{On the Impact of Random Seeds on the Fairness of Clinical Classifiers}

\author{Silvio Amir \and Jan-Willem van de Meent \and Byron C. Wallace \\
  Northeastern University \\
  \texttt{\small{\{s.amir,j.vandemeent,b.wallace\}@northeastern.edu}} \\
  }

\begin{document}
\maketitle
\begin{abstract}




Recent work has shown that fine-tuning large networks is surprisingly sensitive to changes in random seed(s). 
We explore the implications of this phenomenon for model fairness across demographic groups in clinical prediction tasks over electronic health records (EHR) in MIMIC-III ––– the standard dataset in clinical NLP research.
Apparent subgroup performance varies substantially for seeds that yield similar overall performance, although there is no evidence of a trade-off between overall and subgroup performance. However, we also find that the small sample sizes inherent to looking at intersections of minority groups and somewhat rare conditions limit our ability to accurately estimate disparities. Further, we find that jointly optimizing for high overall performance and low disparities does not yield statistically significant improvements.
Our results suggest that fairness work using MIMIC-III should carefully account for variations in apparent differences that may arise from stochasticity and small sample sizes.
\end{abstract}

\section{Introduction}
\label{sec:intro}

Fine-tuning pre-trained transformers \cite{vaswani2017attention} such as BERT \cite{devlin2019bert} has become the dominant paradigm in NLP, owing to their performance across a range of downstream tasks.
Clinical NLP --- in which we often aim to make predictions on the basis of notes in electronic health records (EHRs) --- is no exception \cite{alsentzer2019publicly}. 
However, fine-tuning large networks is a stochastic process. 
Performance can vary considerably as a function of hyperparameter choice, and many parameter sets can yield the same validation accuracy (i.e., the model is not identifiable), and more generally the problem is \emph{underspecified}~\cite{d2020underspecification}.
Recent work has demonstrated that the choice of random seeds alone can have dramatic impact on model performance in NLP and beyond, even when all other hyper-parameters are kept fixed~\cite{phang2018sentence,dodge2020fine,d2020underspecification}. 

\begin{figure}
  \centering
  \includegraphics[trim=0 0 0 30, clip, width=0.45\textwidth]{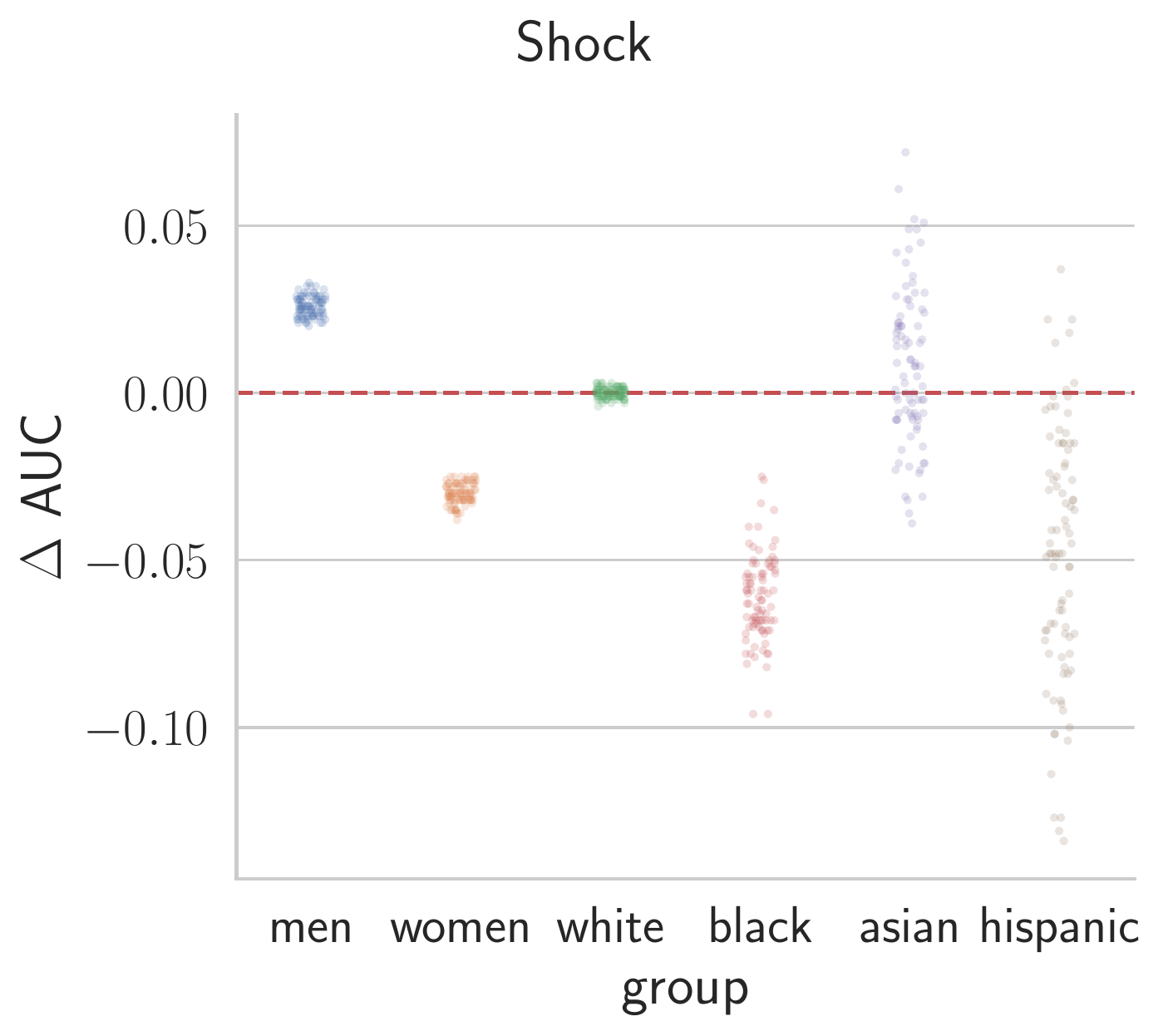}
  \vspace{-0.5\baselineskip}
  \caption{Differences ($\Delta$s) with respect to overall performance as a function of random seeds for demographic subgroups on the \emph{Shock} phenotype classification task. Points represent results from pairs of seeds with similar ($\leq$0.01 difference in AUC) validation performance to the best seeds.} 
  \label{fig:underspec}
\end{figure}

In this work, we explore the intersection of randomness and fairness in the context of clinical NLP. 
Fairness is a particularly acute concern in clinical predictive tasks, given the potential of such models to influence treatment decisions.
This has motivated work investigating biases in predictive models trained over EHR \cite{zhang2020hurtful,chen2018my,chen2019can,chen2020ethical,pfohl2020empirical,chen2020analyzing,tripathi2020fairness}.

We investigate the impact of random seeds on the fairness of fine-tuned classifiers with respect to demographic characteristics such as gender and ethnicity.
There are many definitions of algorithmic fairness which formalize different desired properties \cite{mehrabi2019survey}.  
Following prior work, here we adopt a simple measure: The \emph{mean differences in model performance across demographic subgroups}~\cite{chen2019can}. We find that, on the popular MIMIC-III dataset \cite{johnson2016mimic}, seeds with comparable validation performance can give rise to large variations in disparities across demographic subgroups (Figure \ref{fig:underspec}).

\section{Data and Methods}

We investigate the variability of overall model performance and fairness across random seeds for a set of clinical prediction tasks derived from the Multiparameter Intelligence Monitoring in Intensive Care (MIMIC-III) set of Electronic Health Records (EHRs; \citealt{johnson2016mimic}). 
For each task, we train a classifier on top of the contextualized representations of a BERT \cite{devlin2019bert} model pretrained over EHR data \cite{alsentzer2019publicly}.

Following recent work exploring randomness and fine-tuning, we consider the seeds used to shuffle the training data and to initialize the model parameters independently \cite{dodge2020fine}. Specifically, we generate $K=1000$ pairs of shuffling and initialization seeds by sampling from a uniform distribution $\mathcal{U}(0,10000)$. For each seed pair, we measure the overall performance as well as the performance for each demographic subgroup in terms of the Area Under the ROC Curve (AUC).

\subsection{MIMIC-III}
\label{sec:mimic}
MIMIC-III is a database of deidentified EHR comprising over 40k patients admitted to the intensive care unit of the Beth Israel Deaconess Medical Center between 2001 and 2012 \citep{johnson2016mimic}. 
It comprises structured variables including vital sign measurements, lab test results, and medications. 
It also contains clinical notes (e.g., doctor and nursing notes, radiology reports, and discharge summaries), which are the focus of our analysis. 

MIMIC-III contains demographic information, including potentially sensitive attributes such as ethnicity/race, sex, spoken language, religion, and insurance status (which may be seen as a proxy for socioeconomic status \cite{chen2019can}). 
We are interested in the interaction between randomness and fairness in clinical predictions.
Following recent prior work \cite{zhang2020hurtful} on the latter, we focus our analyses on two benchmark tasks proposed by \citet{harutyunyan2019multitask}:

\paragraph{In-hospital Mortality (IHM)} Predict risk of in-hospital mortality based on the first 48 hours of an ICU stay.
    
\paragraph{Phenotype Classification (PC)} Classify which of 25 acute or chronic conditions (e.g., acute cerebrovascular disease, chronic kidney disease) are present in a given patient ICU stay record. 
Similar to \citet{zhang2020hurtful}, we treat each condition as an independent binary classification task. Table \ref{tab:tasks} in the Appendix enumerates the full set of conditions and their respective prevalences. 

We extracted training and test datasets for these tasks using the same pre-processing pipeline as \citet{zhang2020hurtful}.\footnote{\url{https://github.com/MLforHealth/HurtfulWords}} 
We kept the same data splits and reserved 20\% of the training data as validation set per task. For each patient, we collected their clinical notes, as well as their gender and race/ethnicity (as recorded in the EHR). The notes were filtered according to the categories \emph{Nurse}, \emph{Physician} and \emph{Nursing/Other} to avoid notes of poor semantic quality, as suggested by \citet{zhang2020hurtful}. Patients without relevant clinical notes were discarded, resulting in $11384$/$2591$ and $22033/4919$ training/test examples for the IHM and PC tasks, respectively. It should be noted that these datasets are highly imbalanced both in terms of labels and demographic distribution with $55\%$ \emph{Male}, $85\%$ \emph{White}, $9\%$ \emph{Black}, $3\%$ \emph{Asian} and $3\%$ \emph{Hispanic} patients. Table \ref{tab:samples} shows the distribution of sample sizes across subgroups for each benchmark.

\begin{table}[tbh]
\centering
\scalebox{0.75}{
\begin{tabular}{l l r r r r r r}
\toprule
\multirow{2}{*}{} & \multirow{2}{*}{}  & \multicolumn{2}{c}{\bf Gender}&\multicolumn{4}{c}{\bf Ethnicity} \\
\cmidrule(lr){3-4} \cmidrule(lr){5-8}
  & & \multicolumn{1}{c}{M} & \multicolumn{1}{c}{F} & \multicolumn{1}{c}{W} & \multicolumn{1}{c}{B} & \multicolumn{1}{c}{H} & \multicolumn{1}{c}{A} \\
 \midrule
 \textbf{IHM} & Train & $6,262$ & $5,122$ & $8,044$ & $1,081$ & $353$ & $251$ \\ 
& Test & $1,438$ & $1,153$ & $1,860$ & $226$ & $84$ & $49$ \\ 
& Val & $1,580$ & $1,268$ & $2,027$ & $2,64$ & $85$ & $62$ \\ 
\midrule
\textbf{PC} & Train & $12,372$ & $9,661$ & $15,652$ & $2,049$ & $728$ & $485$ \\ 
& Test & $2,752$ & $2,167$ & $3,552$ & $451$ & $159$ & $95$ \\ 
& Val & $3,029$ & $2,480$ & $4,002$ & $521$ & $196$ & $120$ \\ 
\bottomrule
\end{tabular}
}
\caption{Sample sizes across subgroups for the in-hospital mortality and phenotype classification tasks.}
\label{tab:samples}
\end{table}

\subsection{Fine-tuned Classifiers}
\label{sec:models}
We define text classifiers for clinical tasks that map clinical notes corresponding to individual patients to binary labels. 
We extract contextualized embeddings from notes using a pretrained Transformer encoder and then map these to outputs (predictions) via a linear layer.
Transformers are feedforward networks and require fixed-length inputs.
To handle longer sequences, we adopt an approach used in prior works \cite{huang2019clinicalbert, zhang2020hurtful}. Given an input sequence, we: (1) Extract $N$ subsequences with sizes equal to that expected by the Transformer input layer; (2) Make individual predictions on the basis of each subsequence, and; (3) Then aggregate them into a final prediction.

More formally, an encoder $\phi$ operates over inputs of size $E$ with $H$-dimensional hidden layers. 
Given a patient's clinical notes $\mathcal{X}$, we extract a set of $N$ subsequences of length $E$,
\begin{equation*}
x
=
\{\{w_1^1, \ldots, w_{E}^1\}, 
  \ldots,
  \{w_{1}^N, \ldots, w_{E}^N\}\}
\subseteq \mathcal{X}
\end{equation*}
We construct a matrix $\mathbf{Z} \in \mathbb{R}^{H \times N}$ such that the $n$th column represents subsequence $x_n$, ${\mathbf{Z_{[:,n]}} = \phi(x_n) = \sum_j \mathbf{z}_{j}^{x_n}}$ where $\mathbf{z}_{j}^{x_n} \in \mathbb{R}^H$ is the embedding  produced by the last hidden layer of the encoder for token $j$ in the context of $x_n$. 
We then use a linear layer followed by a sigmoid activation to produce a prediction vector $\mathbf{\tilde{Y}}$, encoding the class conditional probabilities for each subsequence. This vector is then used to calculate the final probability as
\begin{align}
\label{eq:classifier}
    P(Y=1|\mathbf{\tilde{Y}}) &= \frac{\mathbf{\tilde{Y}_{\text{max}}} + \mathbf{\tilde{Y}}_{\text{mean}} N/c}{1+N/c}, 
\end{align}
where $c$ is a scaling factor, which we set to $c=2$, following \citet{huang2019clinicalbert}.

We implement classifiers with \texttt{PyTorch} using the Transformer encoders from the \texttt{huggingface}\footnote{\url{https://huggingface.co/}} library \citep{wolf2019huggingface}.
We initialize models to weights from 
ClinicalBERT \cite{huang2019clinicalbert}, which was trained over scientific literature and clinical notes. 
We train classifiers on the most recent $N=10$ subsequences of $E=512$ tokens from the notes associated with each patient.
We train using the ADAM  optimizer \citep{kingma2014adam} for $500$ epochs with early stopping. 
We set the learning rate to $\alpha=0.01$, which we found to have the best validation performance on average across all tasks. 

\section{Results}

\begin{figure*}[bt]
  \centering
  \includegraphics[width=0.725\columnwidth]{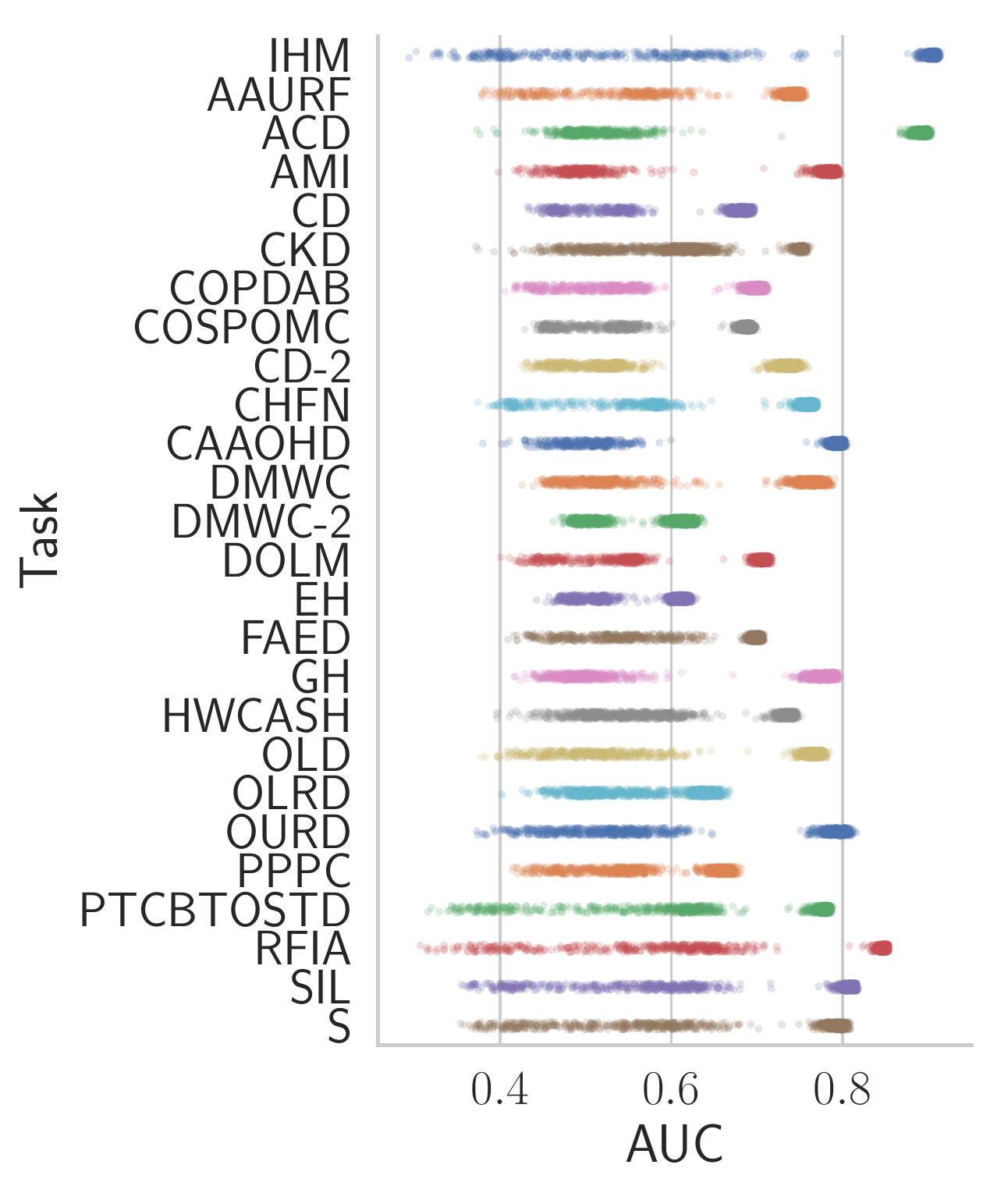}
  \includegraphics[width=0.725\columnwidth]{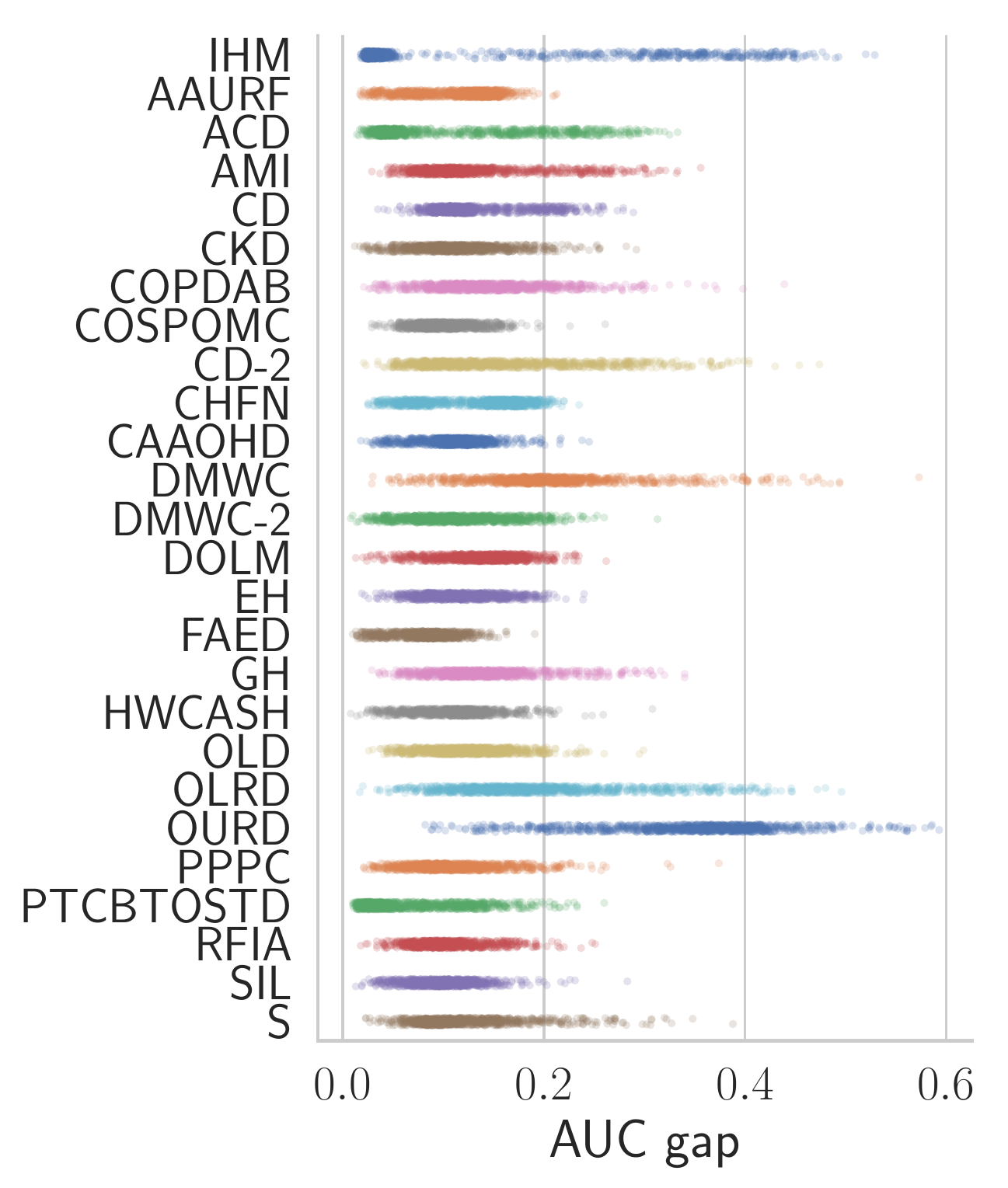}
  \caption{Variation of model performance across random seeds for all tasks (task details in Appendix). \emph{Left:} Overall performance. \emph{Right:} Gap between best and worst subgroup.} 
  \label{fig:gaps}
\end{figure*}

\begin{figure*}
  \begin{minipage}[t]{0.4\textwidth}
    \includegraphics[trim=0 0 0 25, clip, width=\textwidth]{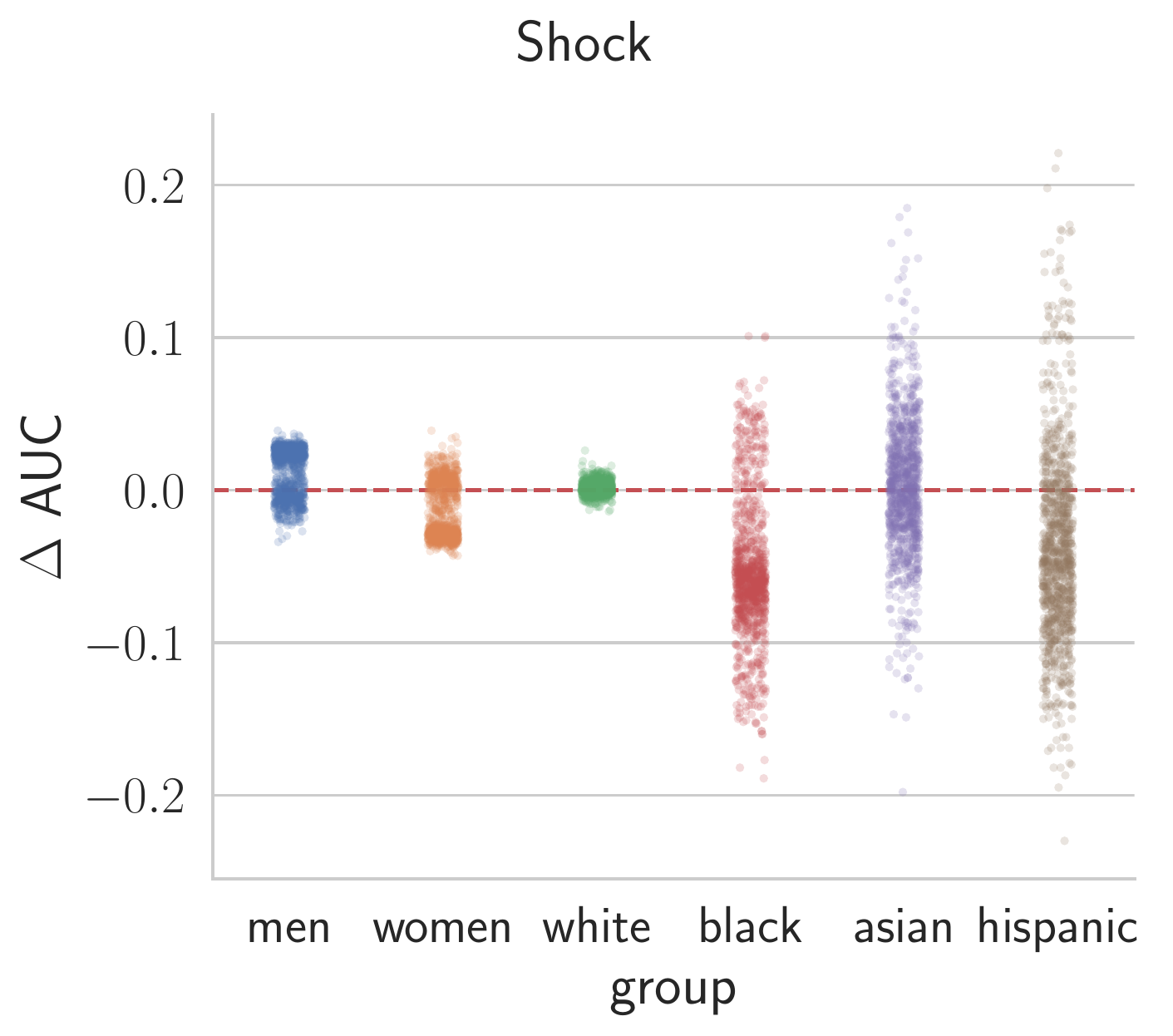}
    \caption{Differences relative to overall AUC as a function of random seeds for subgroups on the \emph{Shock} phenotype classification task.}
    \label{fig:delta_auc}
  \end{minipage}
 \hfill
  \begin{minipage}[t]{0.575\textwidth}
    \includegraphics[trim=0 0 0 25, clip, width=\textwidth]{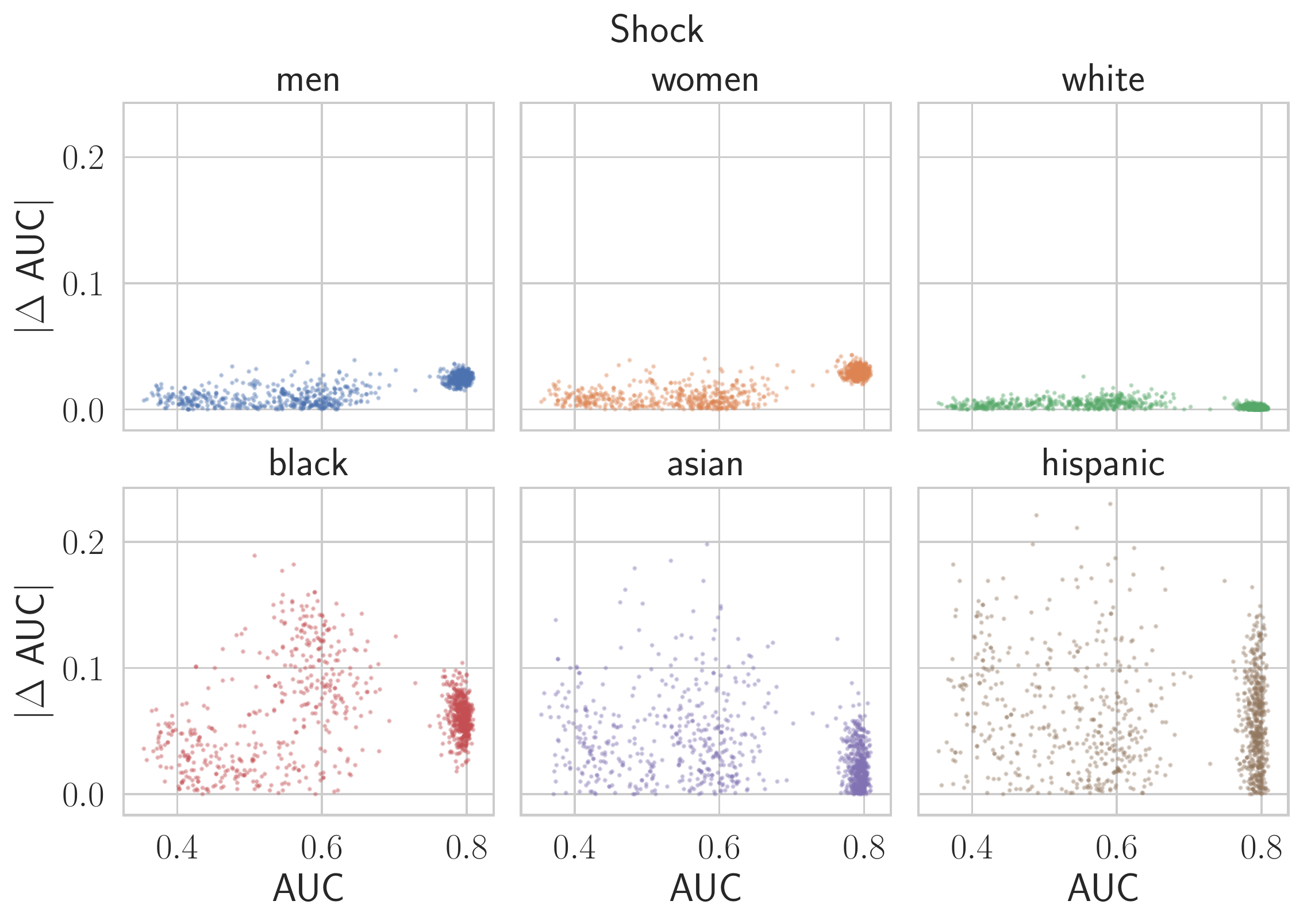}
    \caption{Correlations between overall performance and subgroup performance on the \emph{Shock} phenotype classification task.}
    \label{fig:auc_x_delta_auc}
  \end{minipage}
\end{figure*}

\begin{figure*}[bt]
  \centering
  \includegraphics[width=0.7\columnwidth]{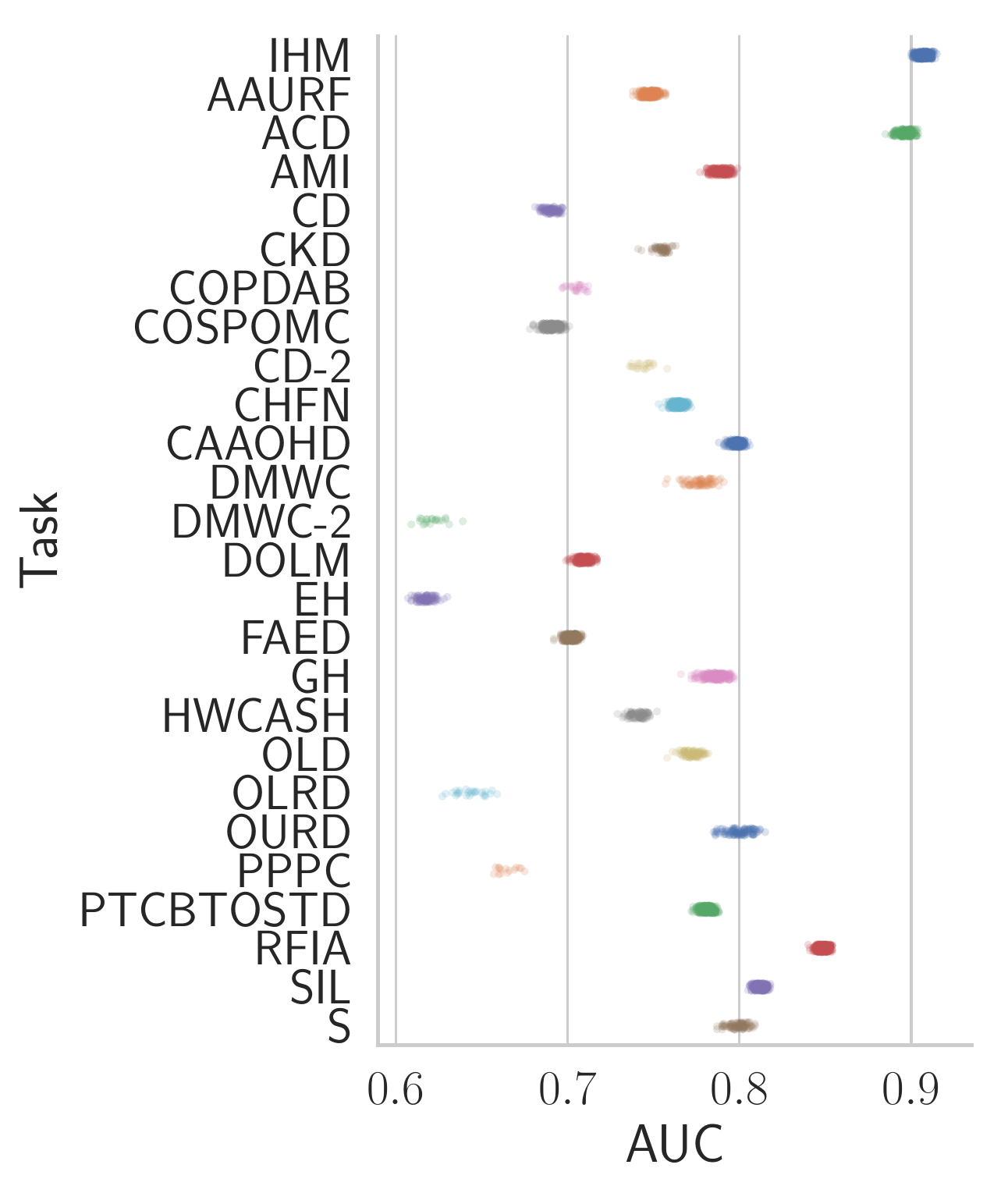}
  \includegraphics[width=0.7\columnwidth]{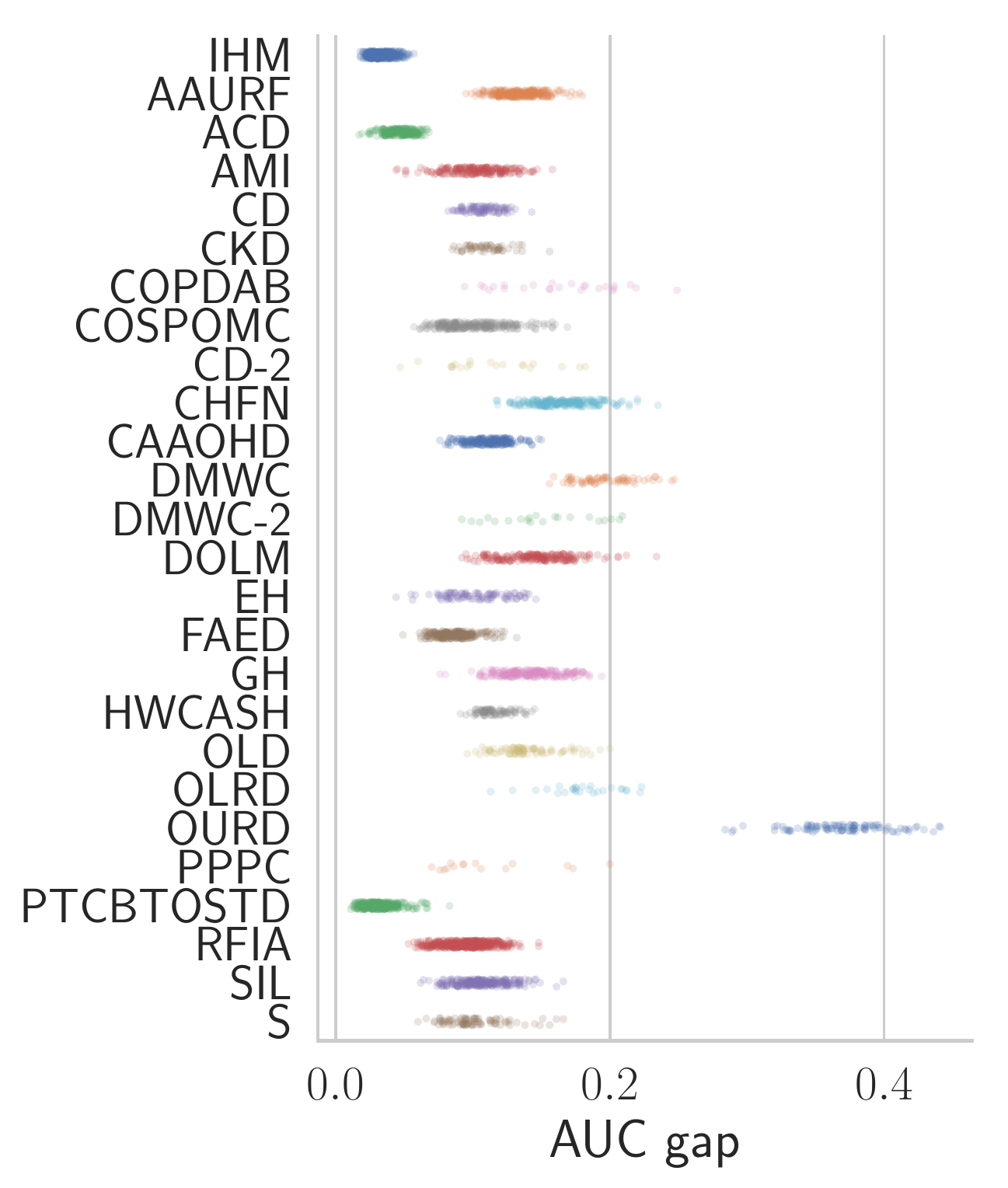}
  \caption{Variation of model performance for seeds with validation performance similar to the best seeds, for all tasks. \emph{Left:} Overall performance. \emph{Right:} Gap between best and worst subgroup.} 
  \label{fig:underspec_gaps}
\end{figure*}

We compare the overall performance with the performance for each subgroup as a function of random seeds. 
Figure \ref{fig:gaps} shows the overall performance (left) along with the gap between the best and worst observed subgroup AUCs (right), across tasks. 
We observe a large variance in both the overall performance and the gap. 
The former observation corroborates previous findings \cite{dodge2020fine}. 

To quantify how random seeds affect individual subgroups, we measure the the absolute differences ($\Delta$s) between overall and subgroup performances. 
We then evaluate whether there are correlations between overall performance and subgroup $\Delta$s. 
Figures \ref{fig:delta_auc} and \ref{fig:auc_x_delta_auc} present the results for the \emph{Shock} phenotype classification task --- one of the tasks with largest disparities observed in prior work \cite{zhang2020hurtful}. Similar trends were found for the remaining tasks, and we report all results in the Appendix (Figures \ref{fig:deltas_1}-\ref{fig:deltas_2} and \ref{fig:scatters_1}-\ref{fig:scatters_3}).

Figure \ref{fig:delta_auc} shows that the performance of all subgroups varies significantly across random seeds and that variances are higher for minority groups. 
Larger variations in minority subgroups are to be expected, as any empirical estimate will have a variance that is inversely proportional to the sample size of a group. 
In Figure \ref{fig:auc_x_delta_auc}, we observe that there seem to be two distinct clusters of seeds: One corresponding to high performing models (right of plots), and another to suboptimal models.\footnote{\citet{dodge2020fine} also found that some seeds performed consistently well across all the evaluated tasks, while others always performed poorly.} 
While the best performing models tend to have a lower variance of subgroup performance, there is otherwise no clear relationship between overall and subgroup performance. Indeed, we find that many models with similar overall performance correspond to widely different subgroup $\Delta$s, particularly for the minority groups. 


To explore the implications of this phenomenon, we simulate a grid search over all the random seeds on the validation set. We select the best seed along with all other seeds with similar performance (i.e., within a difference of $\epsilon=0.01$ absolute AUC). Figure \ref{fig:underspec} shows the test set subgroup performance $\Delta$s, for the best validation seeds, in the \emph{Shock} phenotype classification task (see Figures \ref{fig:underspec_1}-\ref{fig:underspec_2} for the other tasks). 
Figure \ref{fig:underspec_gaps} summarizes the overall performance (left) along with the subgroup performance gap (right) across tasks.

We can see that selecting seeds on the basis of overall performance helps to reduce the subgroup performance gap (compare the right subplots in Figures \ref{fig:gaps} and \ref{fig:underspec_gaps}). 
However, the top performing models show disparities with respect to both gender and ethnicity, suggesting that these models maximize performance for some groups at the expense of others. 
Moreover, we find \emph{multiple seeds with similar levels of validation performance that correspond to very different subgroup $\Delta$s}. 

Since we have not encoded any model selection preferences into the pipeline this variance may reflect a form of \emph{underspecification}. 
Can we define criteria that explicitly accounts for subgroup performance? We could then ask whether it is possible to maximize both fairness and overall performance with respect to random seeds. We repeated the grid search experiments with simple criteria that incorporate some notion of subgroup performance, such as selecting the seeds that: (a) maximize subgroup macro-average performance; (b) minimize the average subgroup $\Delta$; and (c) maximize the overall performance minus the average subgroup $\Delta$.
To account for the effect the sample sizes on the apparent subgroup performance, we directly compare subgroup $\Delta$s for each random seed on the validation set and the test set. 
We find that correlations between validation set and test set fairness are either non-existent or very weak in most tasks. 

These findings imply that the same pipeline may produce models with similar validation performance but very different levels of apparent `fairness' as a result of varying the random seed alone. However, the fact that training-set and validation-set fairness are not reliable indicators of test-set fairness suggests that variance due to small subset sizes may be significant. 

\begin{figure}
  \centering
  \includegraphics[trim=0 0 0 30, clip, width=0.45\textwidth]{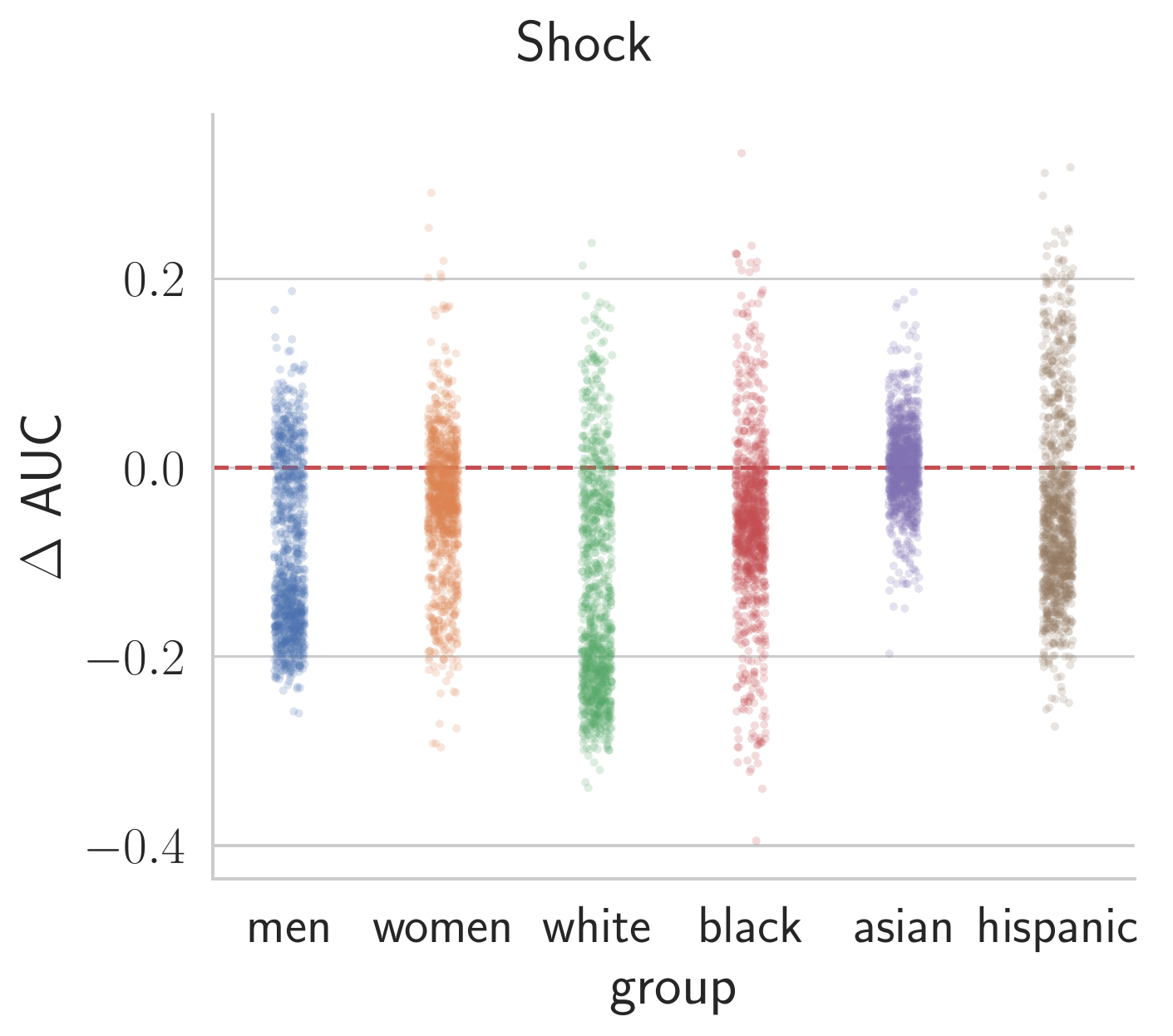}
  \vspace{-0.5\baselineskip}
  \caption{Differences ($\Delta$s) estimated from a balanced subset of the test data with equal sample sizes for all demographic subgroups. As in Figure~\ref{fig:delta_auc}, we show deviations relative to the overall AUC for the \emph{Shock} phenotype classification task.}  
  \label{fig:subsample_deltas}
\end{figure}

This is in some sense not surprising, given the combination of pronounced class imbalance and small subgroup samples in this data (see Tables \ref{tab:samples} and \ref{tab:tasks}). To confirm this, we repeat the experiments on a subset of the test data containing the same number of examples (equal to the smallest subgroup) for \emph{all} groups, including majority groups. Evaluating all subgroups using small samples yields similarly high variances in performance $\Delta$ (Figure \ref{fig:subsample_deltas} and Appendix Figure \ref{fig:subsample_scatters}), which confirms that the sample size is a significant factor in the variation of apparent model performance across random seeds.

These findings suggest that work investigating the fairness of fine-tuned classifiers should be careful to account for: a) model variability due the choice of random seeds; and b) variance in performance estimates due to small sample sizes. 
See Appendix Section \ref{sec:finetunings} for an illustrative example. 
These observations are relevant for research using MIMIC-III, and for any corpora with similar properties, namely the combination of class imbalance and comparatively small subgroup sizes, which is likely to be present in EHR data where conditions are relatively rare and one is interested in fairness to minority groups (which are smaller by definition).

\section{Conclusions} 
\vspace{-.5em}
\label{sec:conclusions}

We have investigated the impact of random seeds on the fairness of fine-tuned pre-trained models for clinical tasks. 
Specifically, we measured gaps in performance across gender and racial subgroups as a function of the choice of random seeds for data shuffling and parameter initialization. 
In line with prior work, we found that classifiers trained on MIMIC-III data are often biased with respect to demographic subgroups.
The contribution of this work is the empirical confirmation that choice of random seed alone significantly affects the apparent bias: Seeds that yield comparable performance in aggregate on the validation data correspond to very different performances on subgroups in test data.
Our analyses corroborate \citet{dodge2020fine}'s findings on the importance of carefully chosen random seeds, but also suggest that an equal amount of attention should be payed to the impact of these choices on model fairness. 

However, interpretation of these results is complicated by sample size effects. While MIMIC-III is in itself a large dataset, it also exhibits significant imbalance, both in terms of subgroups of patients and the prevalence of medical conditions. These imbalances compound when considering subsets of patients in the context of specific prediction tasks, which often leads to small sample sizes for minority subgroups.
While we observed higher apparent variances for demographic minorities, our results also suggest that these variances can in large part be explained by the smaller sample sizes. Indeed, we found the variances in subgroup performance to be inversely proportional to the size of the subgroup. 


\section*{Ethical Considerations}

Fairness has rightly been an issue of increasing concern within the NLP community.
This issue is particularly important in clinical NLP, given the potential that such models may ultimately have on patient health.
We have investigated the degree to which different subgroup performances may be observed even fixing the (aggregate) validation data performance; we find wide variances across subgroups.
That said, this work also highlights inherent limitations of using MIMIC-III (the standard dataset for clinical NLP) to evaluate the fairness of models, given the relatively small samples of patients that belong to demographic groups of interest. 
We hope these contributions encourage continued research into fairness in the context of clinical NLP. 

\section*{Acknowledgements}

We would like to thank Darius Irani for his contribution in replicating the experiments from \cite{zhang2020hurtful}.
This material is based upon work supported in part by the National Science Foundation under Grant No. 1901117.

\bibliographystyle{acl_natbib}
\bibliography{custom}

\appendix

\counterwithin{figure}{section}
\counterwithin{table}{section}

\section{Clinical Prediction Tasks and Results}

\begin{figure*}
  \begin{minipage}[t]{0.32\textwidth}
    \includegraphics[width=\textwidth]{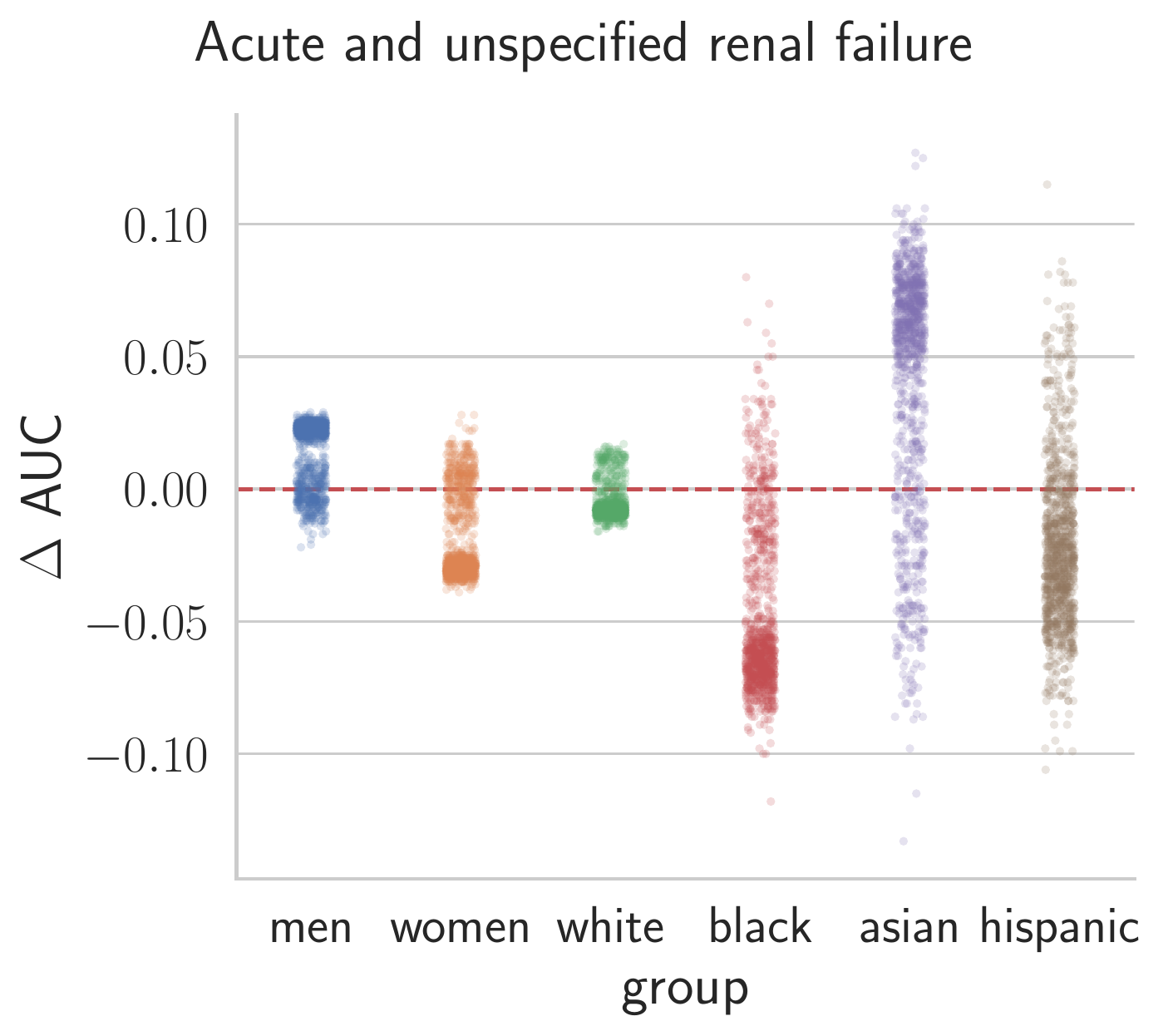}
  \end{minipage}
 \hfill
  \begin{minipage}[t]{0.32\textwidth}
    \includegraphics[width=\textwidth]{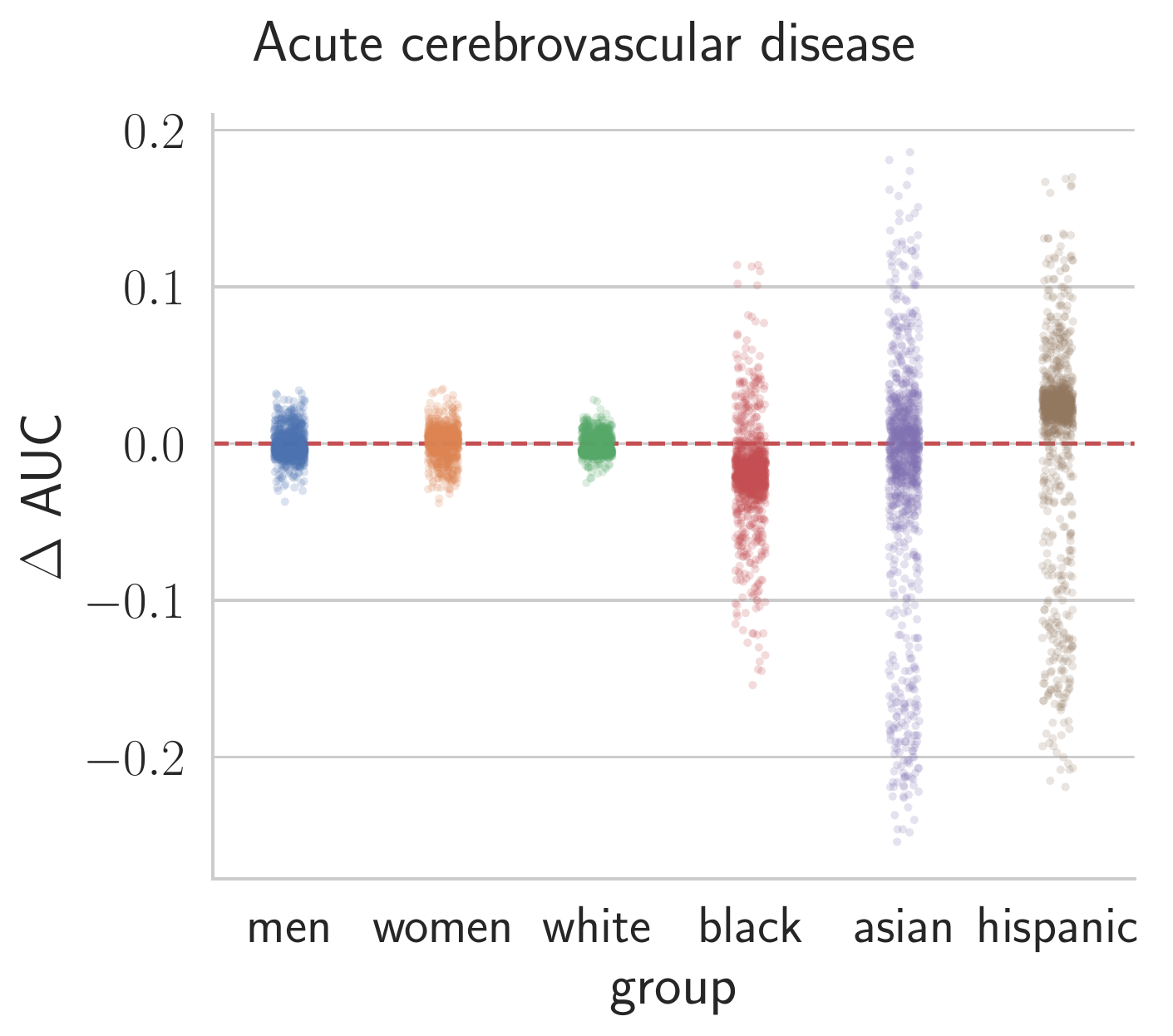}
  \end{minipage}
  \hfill
  \begin{minipage}[t]{0.32\textwidth}
    \includegraphics[width=\textwidth]{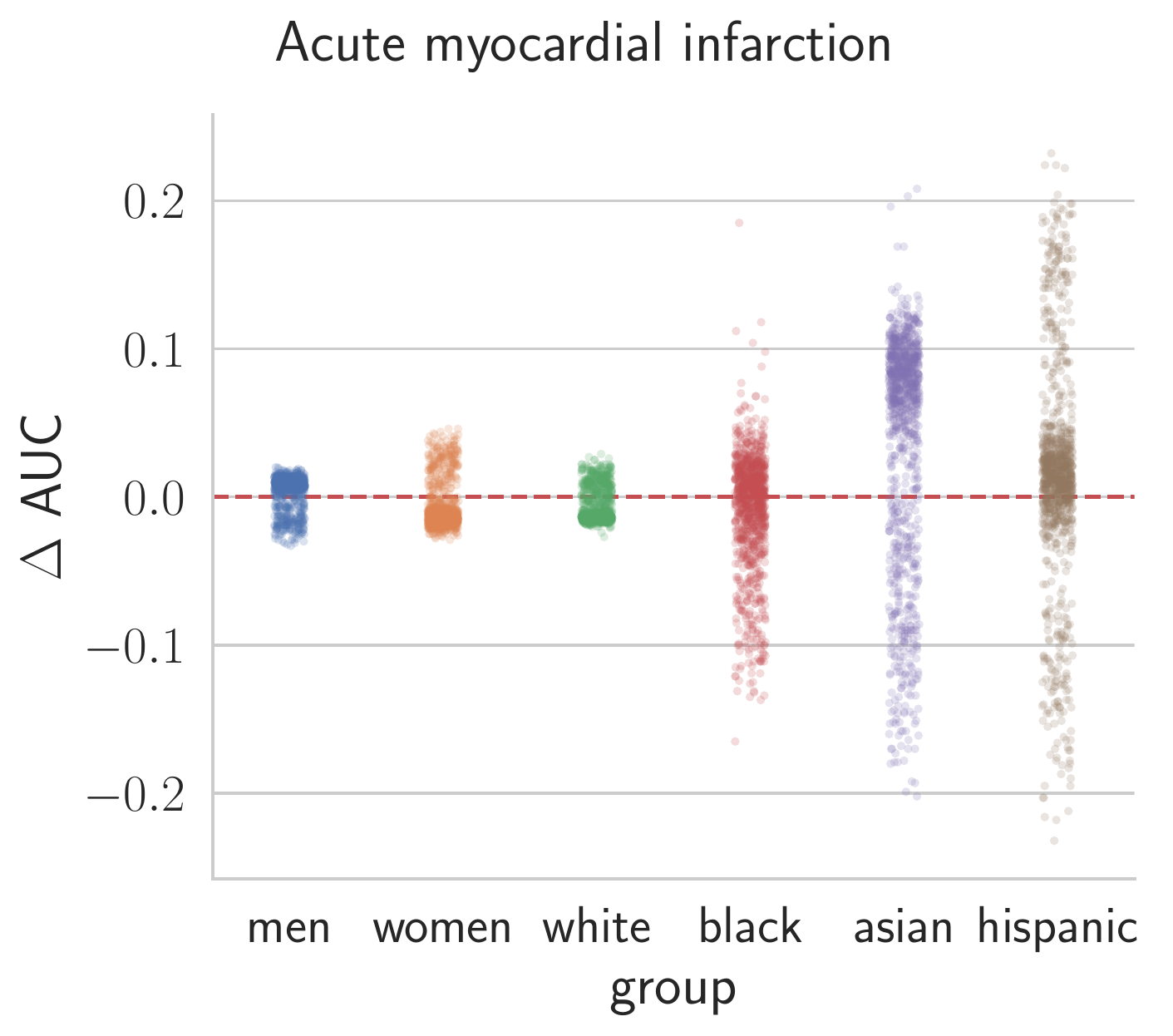}
  \end{minipage}
  \hfill
  \begin{minipage}[t]{0.32\textwidth}
    \includegraphics[width=\textwidth]{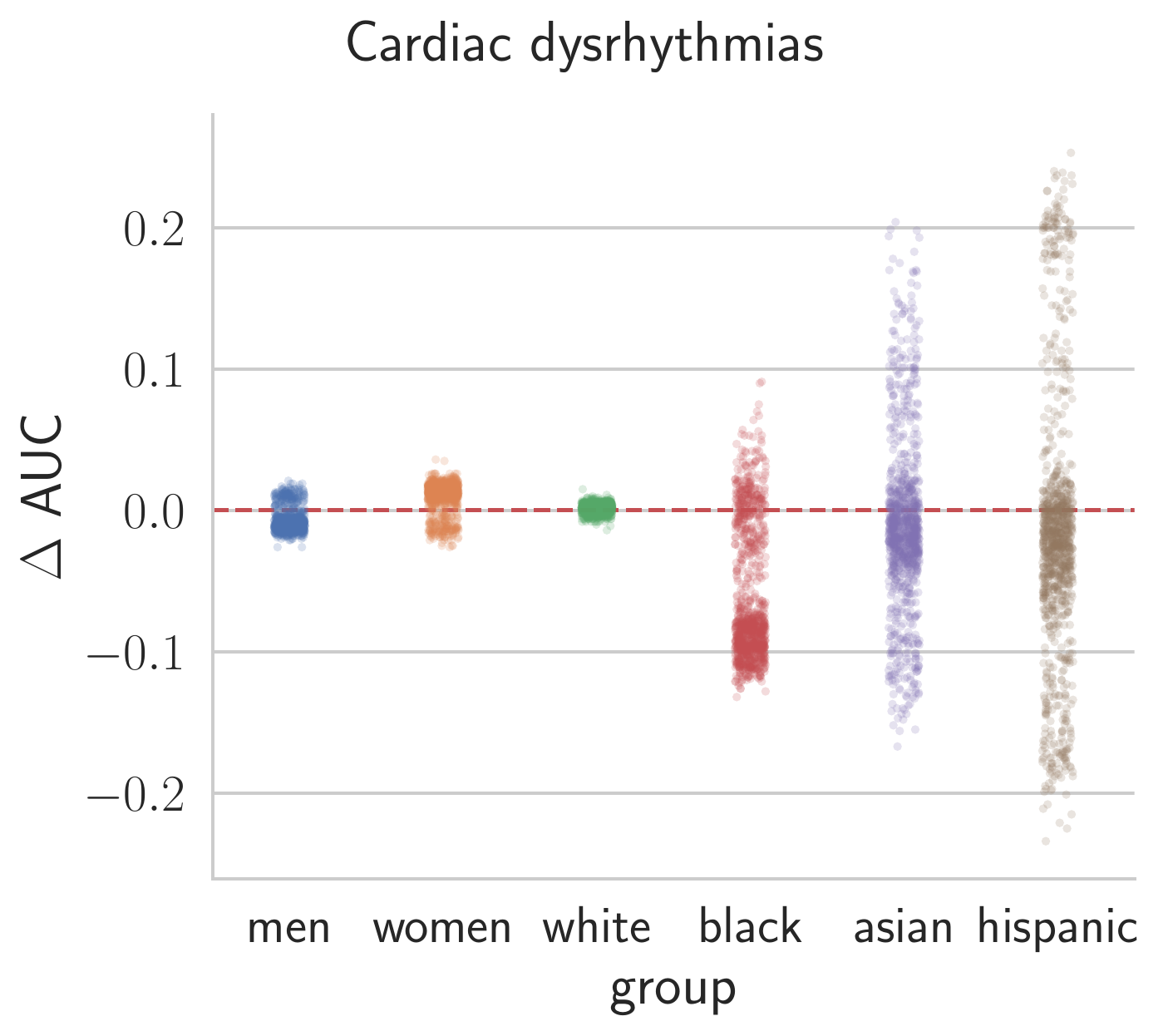}
  \end{minipage}
 \hfill
  \begin{minipage}[t]{0.32\textwidth}
    \includegraphics[width=\textwidth]{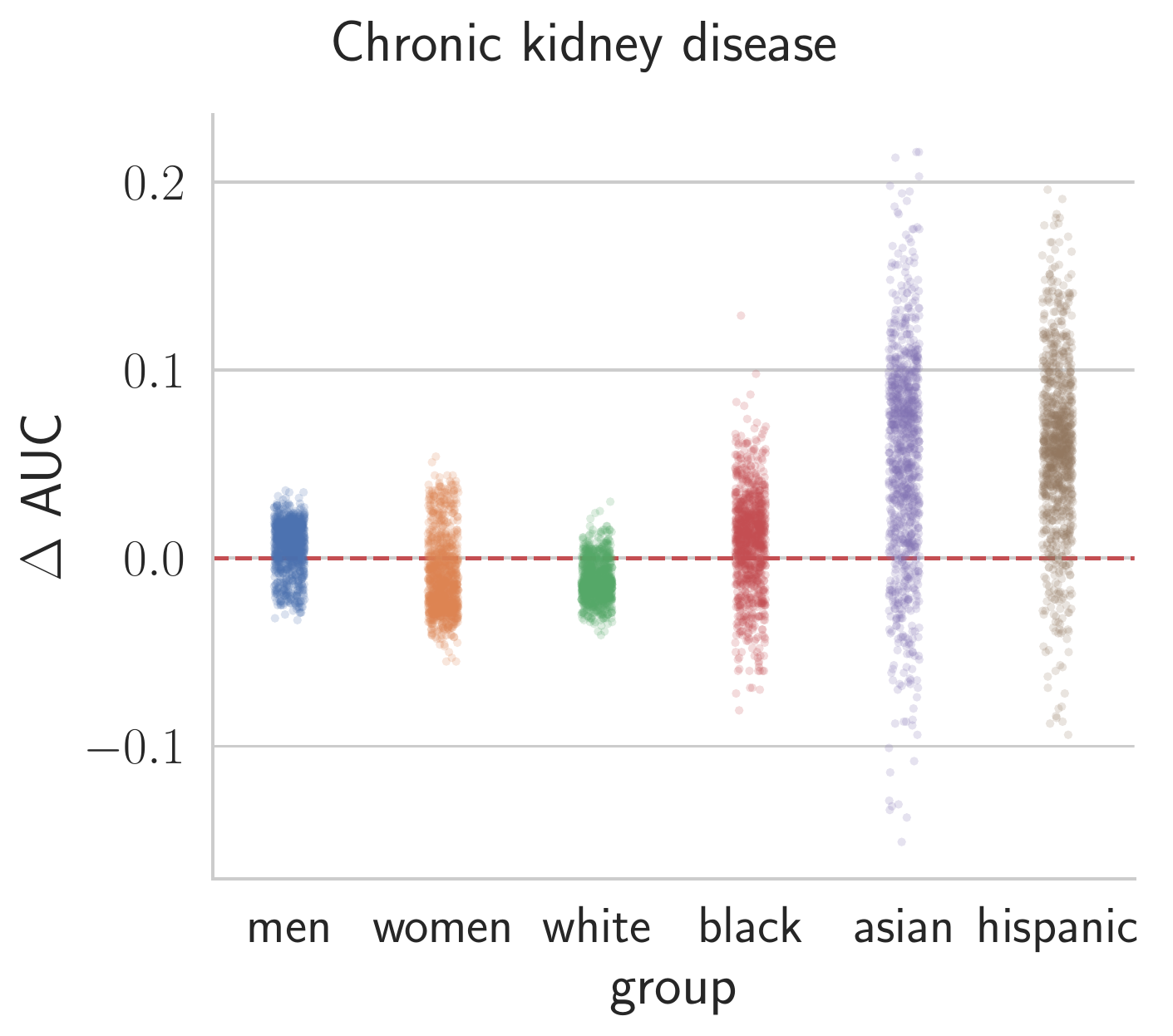}
  \end{minipage}
  \hfill
  \begin{minipage}[t]{0.32\textwidth}
    \includegraphics[width=\textwidth]{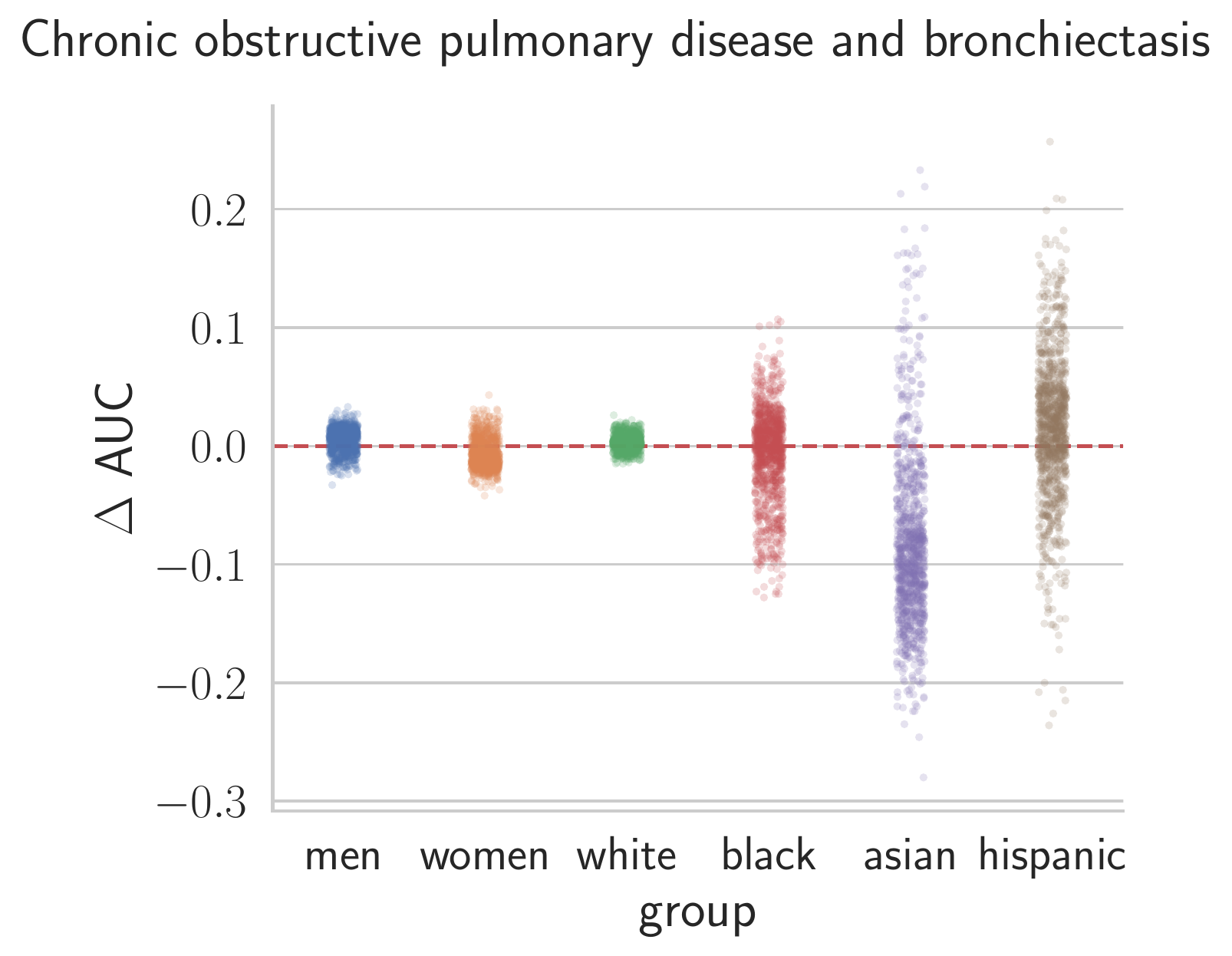}
  \end{minipage}
  \hfill
  \begin{minipage}[t]{0.32\textwidth}
    \includegraphics[width=\textwidth]{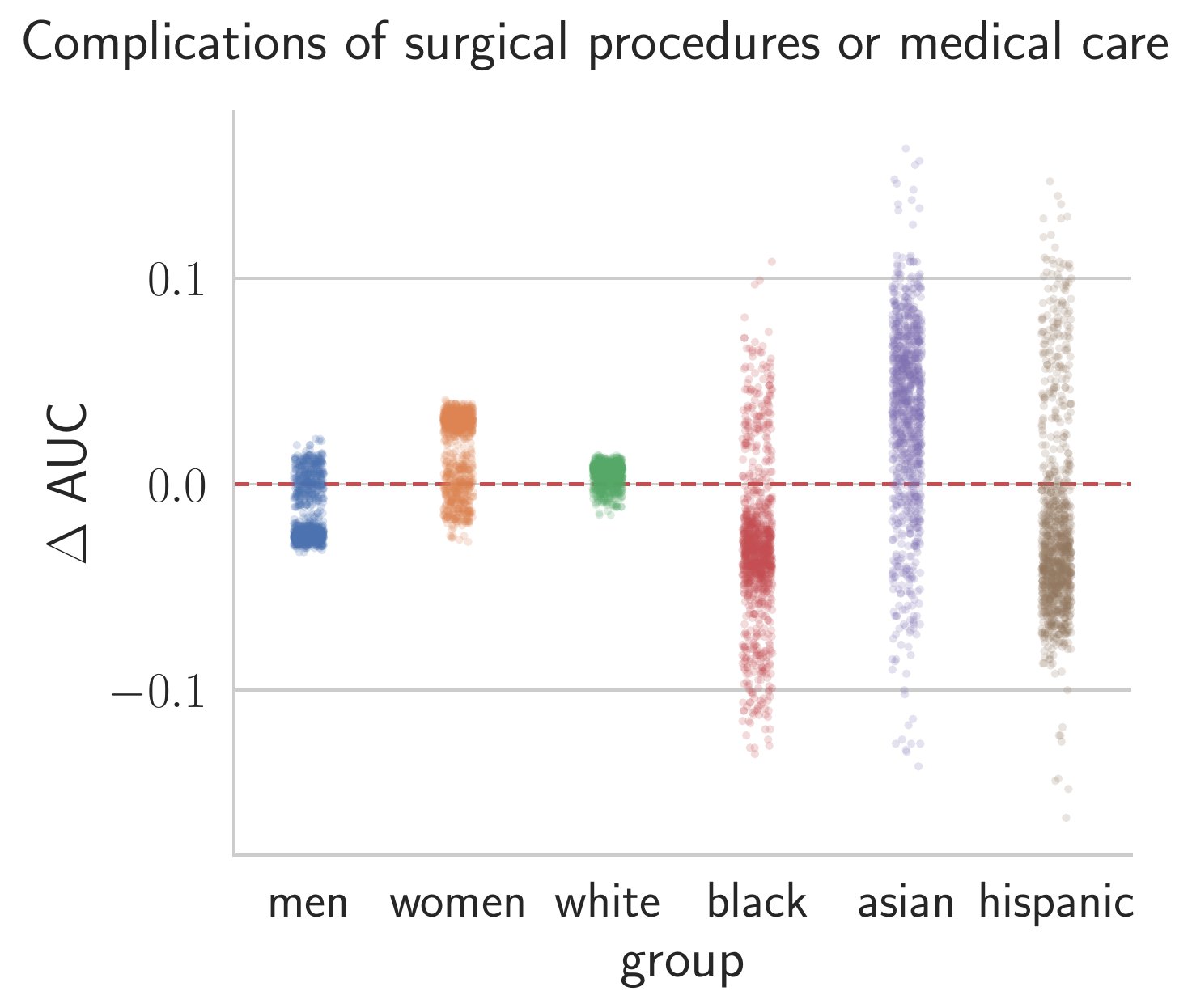}
  \end{minipage}
 \hfill
  \begin{minipage}[t]{0.32\textwidth}
    \includegraphics[width=\textwidth]{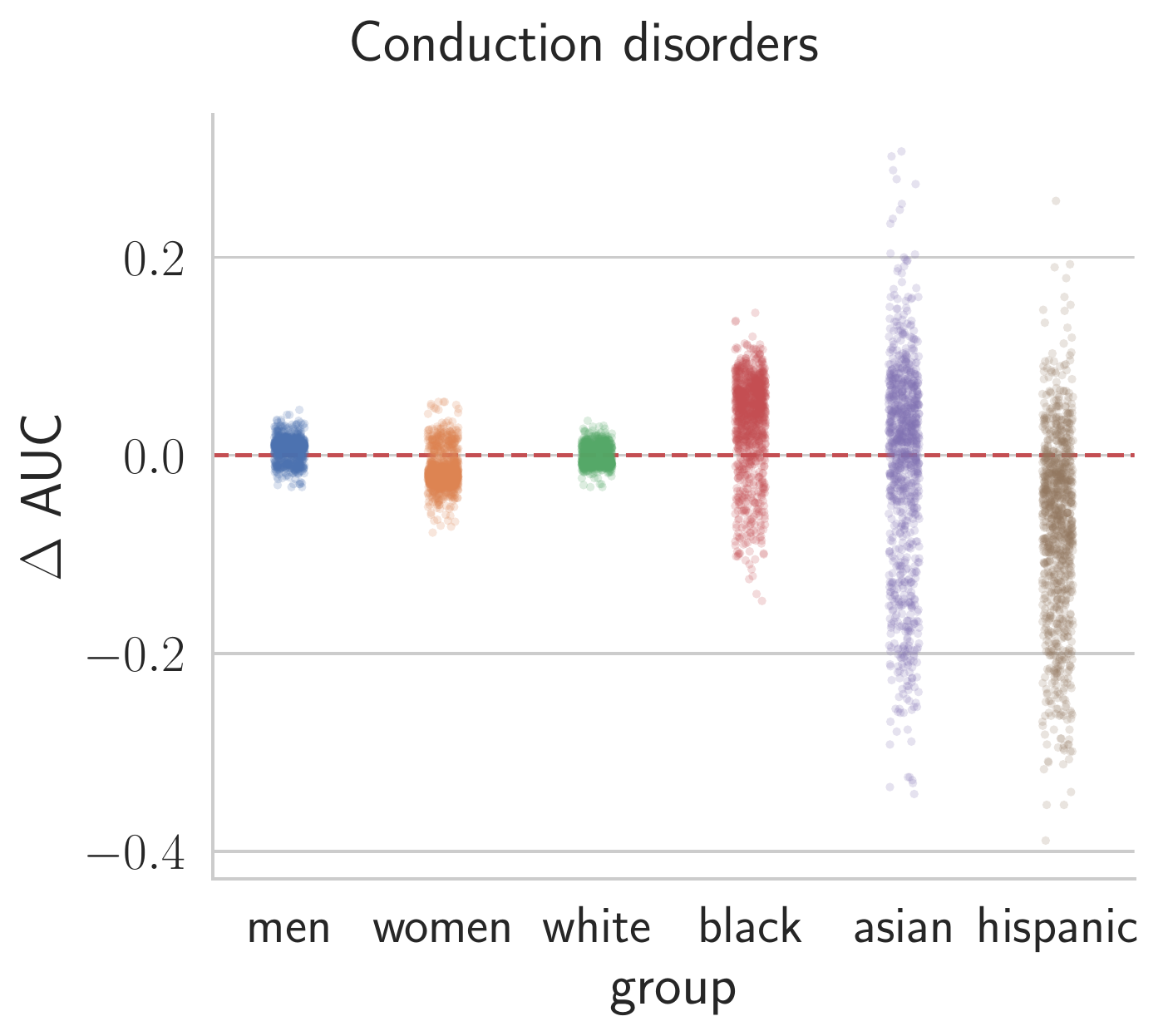}
  \end{minipage}
  \hfill
  \begin{minipage}[t]{0.32\textwidth}
    \includegraphics[width=\textwidth]{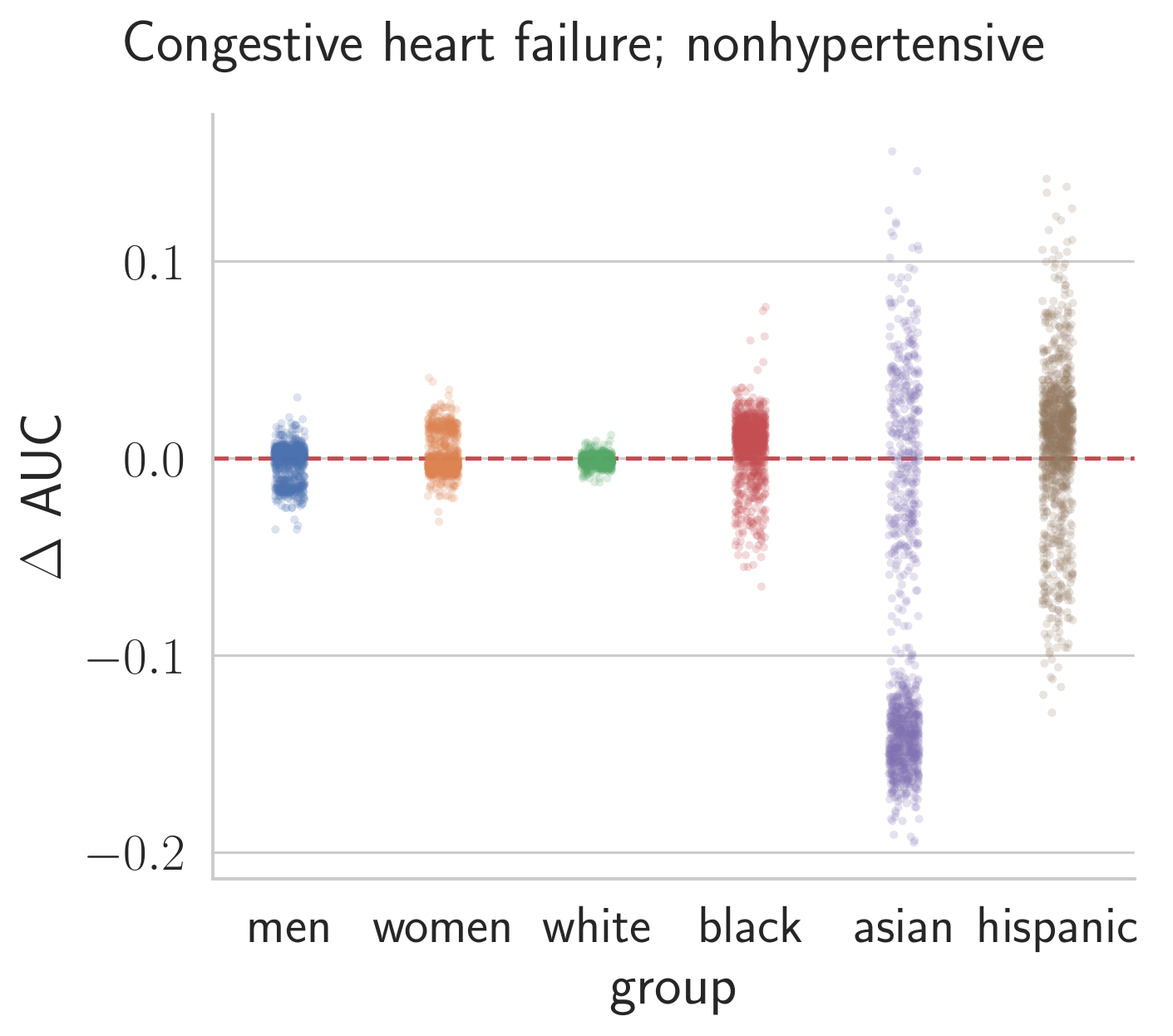}
  \end{minipage}
  \hfill
  \begin{minipage}[t]{0.32\textwidth}
    \includegraphics[width=\textwidth]{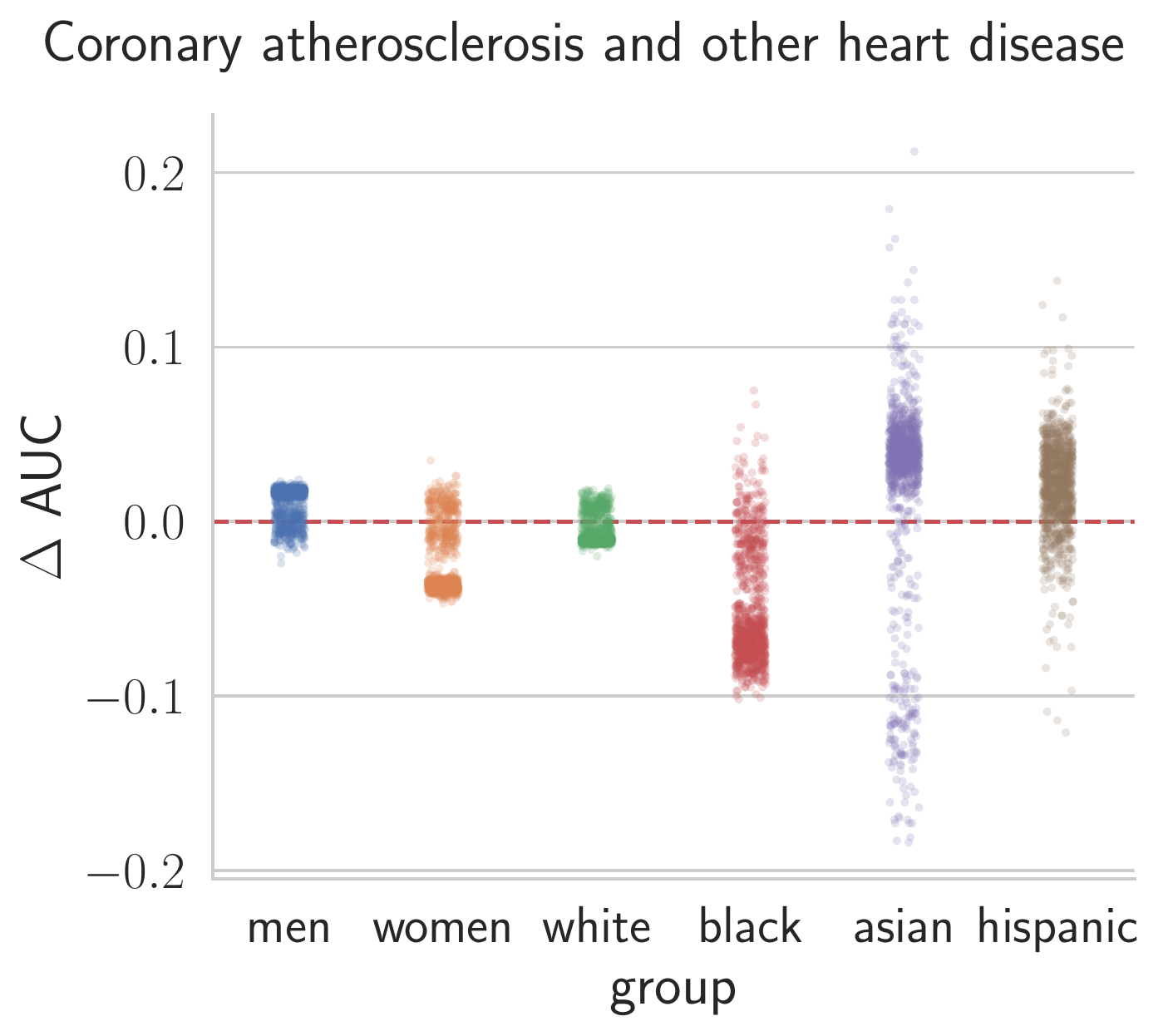}
  \end{minipage}
 \hfill
  \begin{minipage}[t]{0.32\textwidth}
    \includegraphics[width=\textwidth]{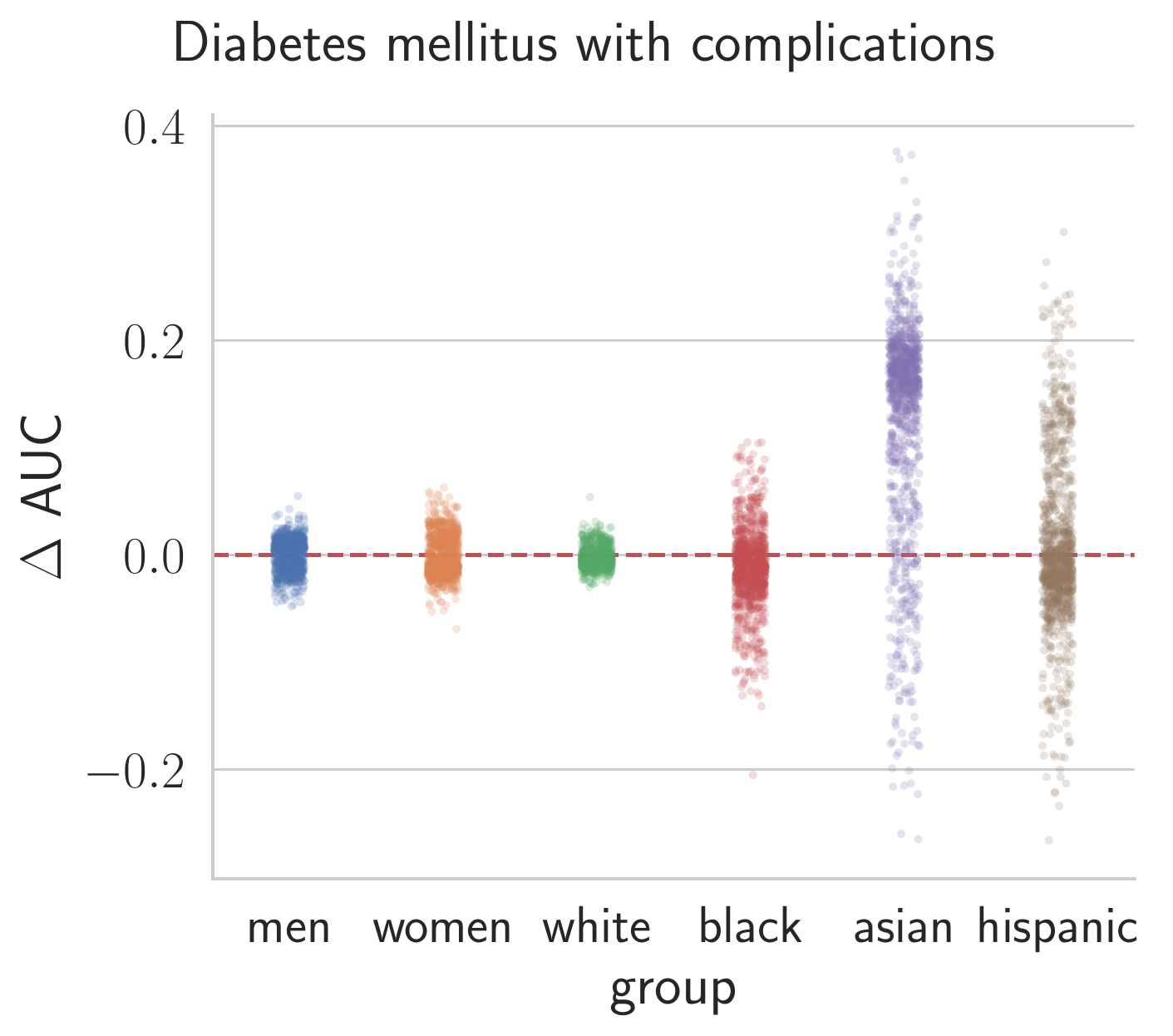}
  \end{minipage}
  \hfill
  \begin{minipage}[t]{0.32\textwidth}
    \includegraphics[width=\textwidth]{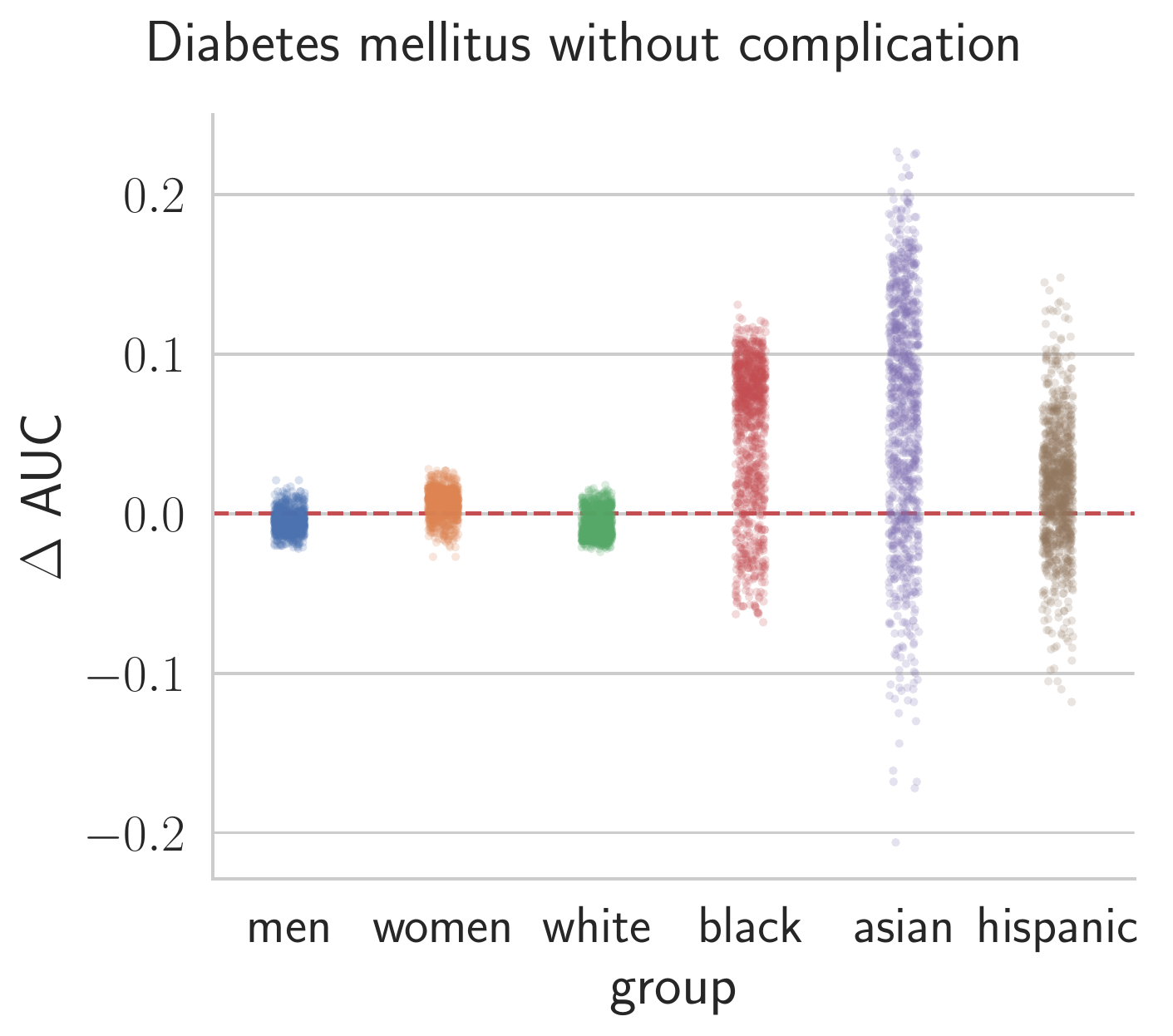}
  \end{minipage}
  \hfill
  \begin{minipage}[t]{0.32\textwidth}
    \includegraphics[width=\textwidth]{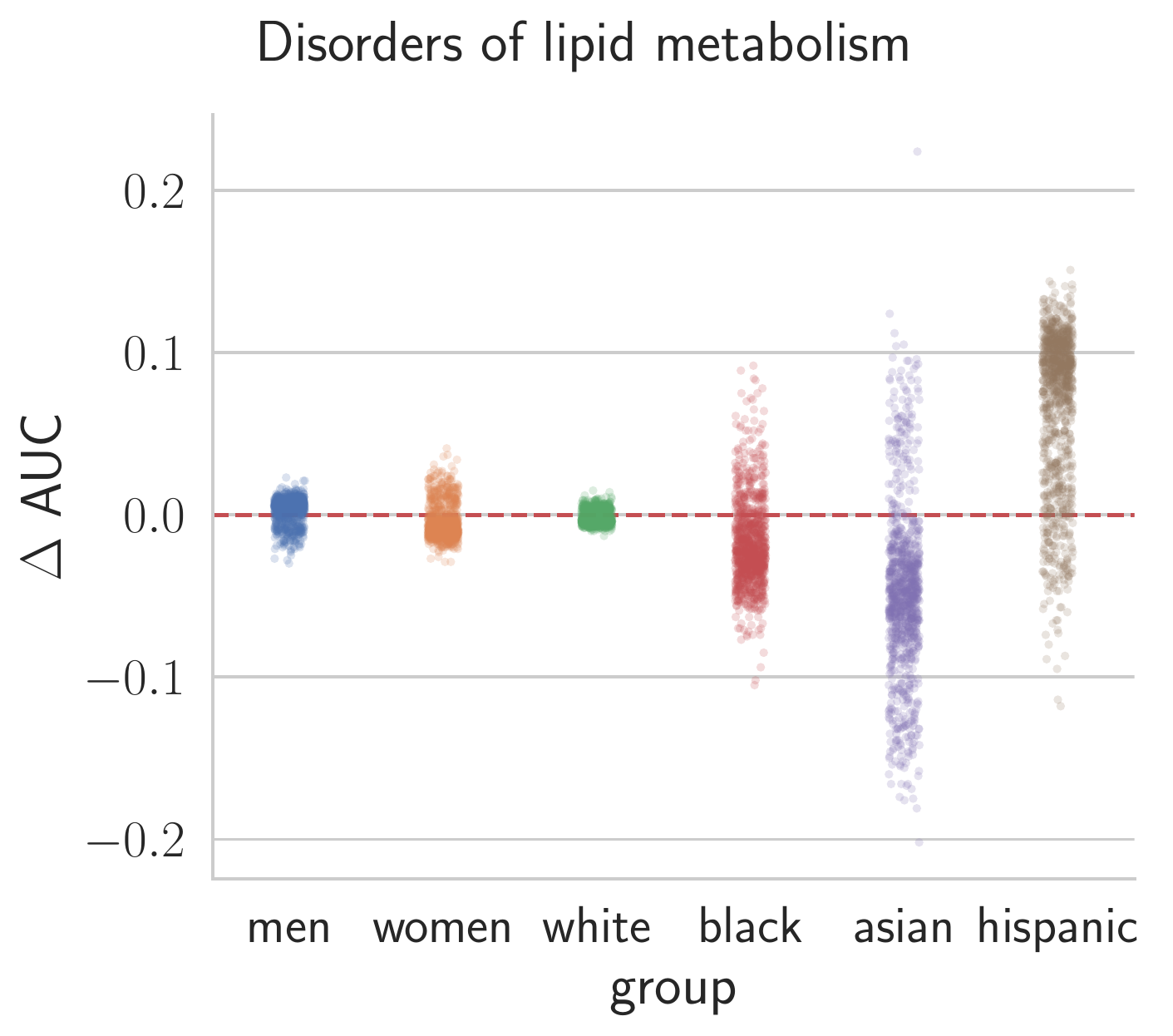}
  \end{minipage}
 \hfill
  \begin{minipage}[t]{0.32\textwidth}
    \includegraphics[width=\textwidth]{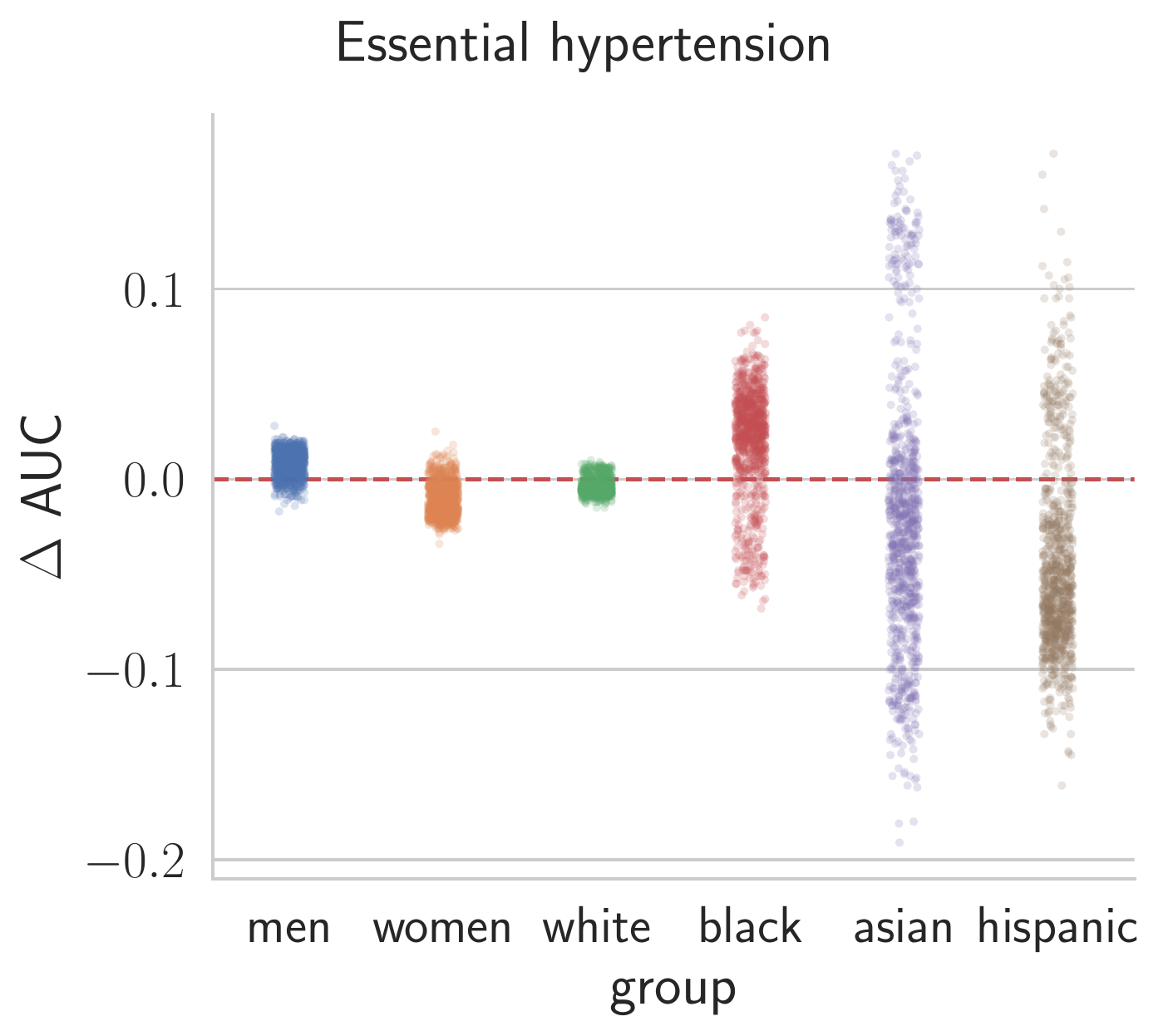}
  \end{minipage}
  \hfill
  \begin{minipage}[t]{0.32\textwidth}
    \includegraphics[width=\textwidth]{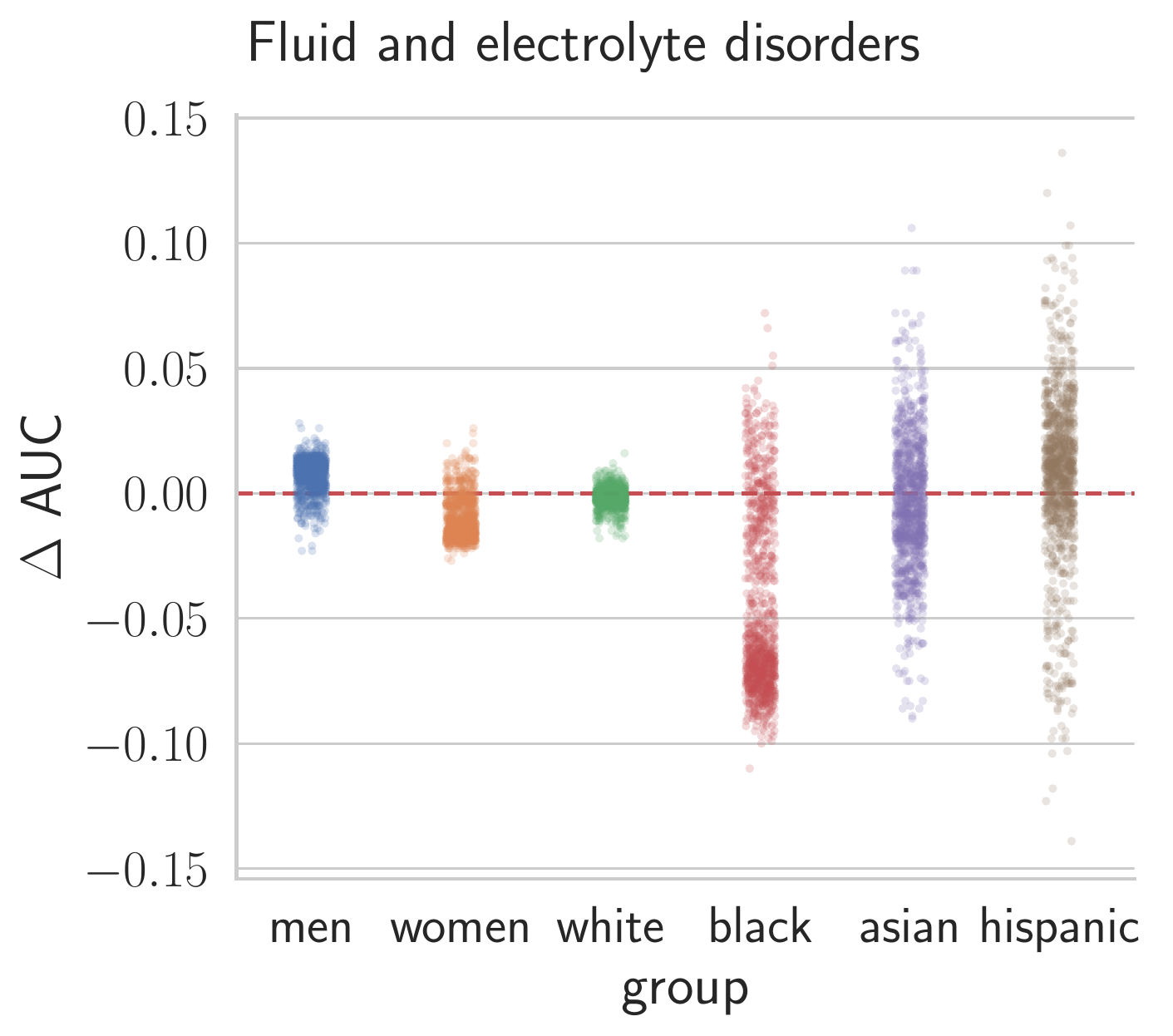}
  \end{minipage}
  \hfill
  \caption{Differences relative to overall performance as a function of random seeds for each subgroup }
  \label{fig:deltas_1}
\end{figure*}

\begin{figure*}
  \begin{minipage}[t]{0.32\textwidth}
    \includegraphics[width=\textwidth]{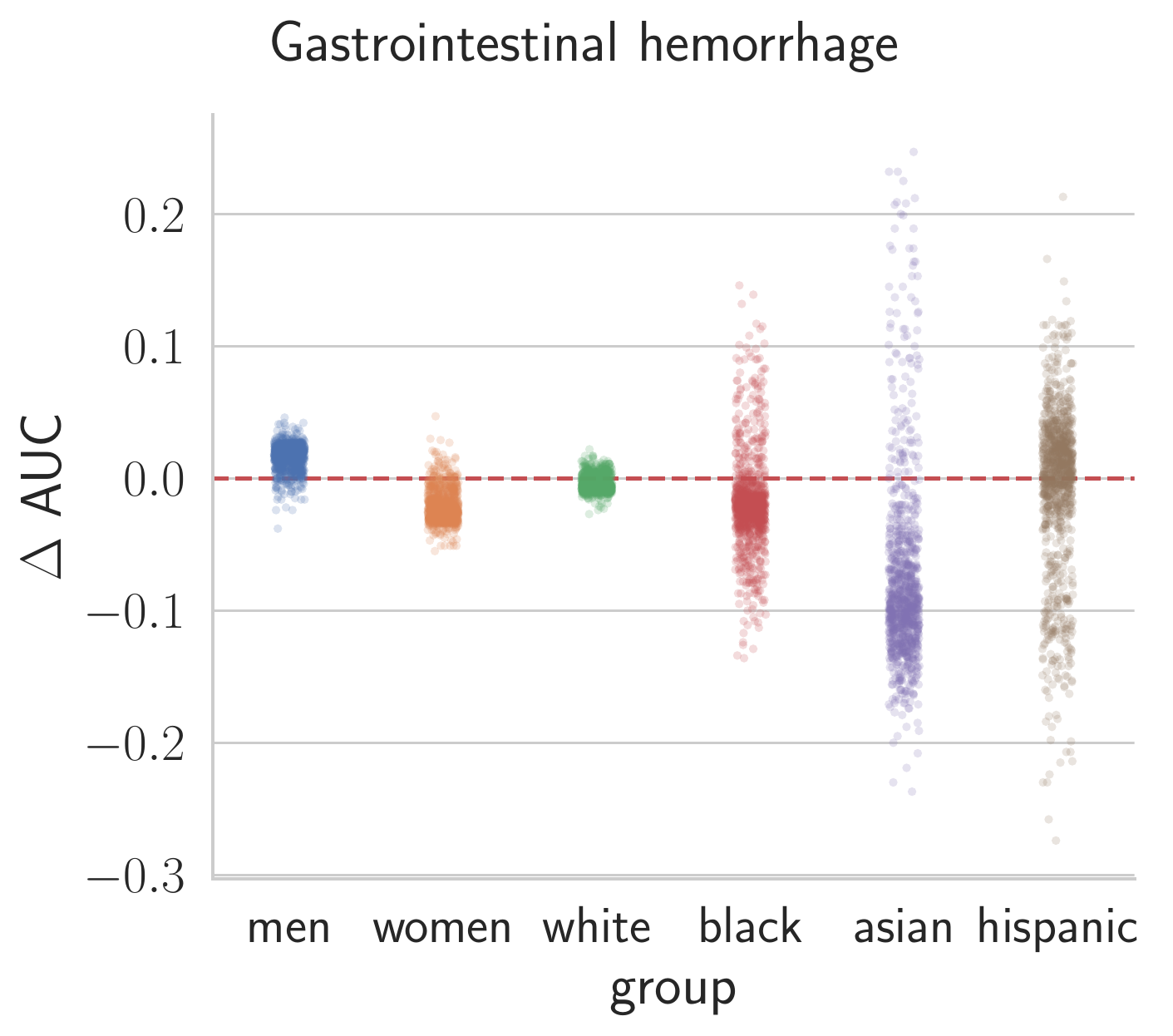}
  \end{minipage}
 \hfill
  \begin{minipage}[t]{0.32\textwidth}
    \includegraphics[width=\textwidth]{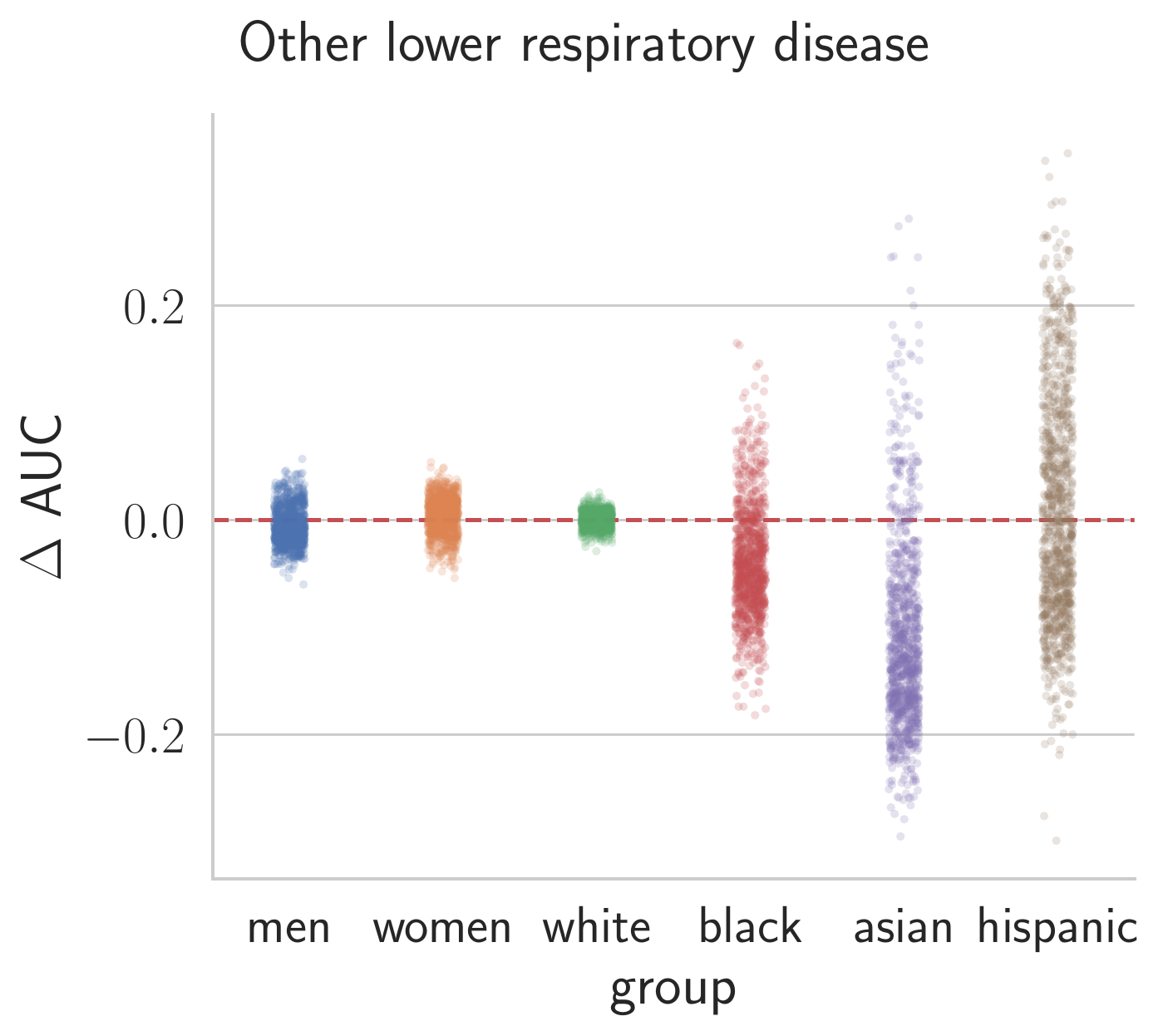}
  \end{minipage}
 \hfill
  \begin{minipage}[t]{0.32\textwidth}
    \includegraphics[width=\textwidth]{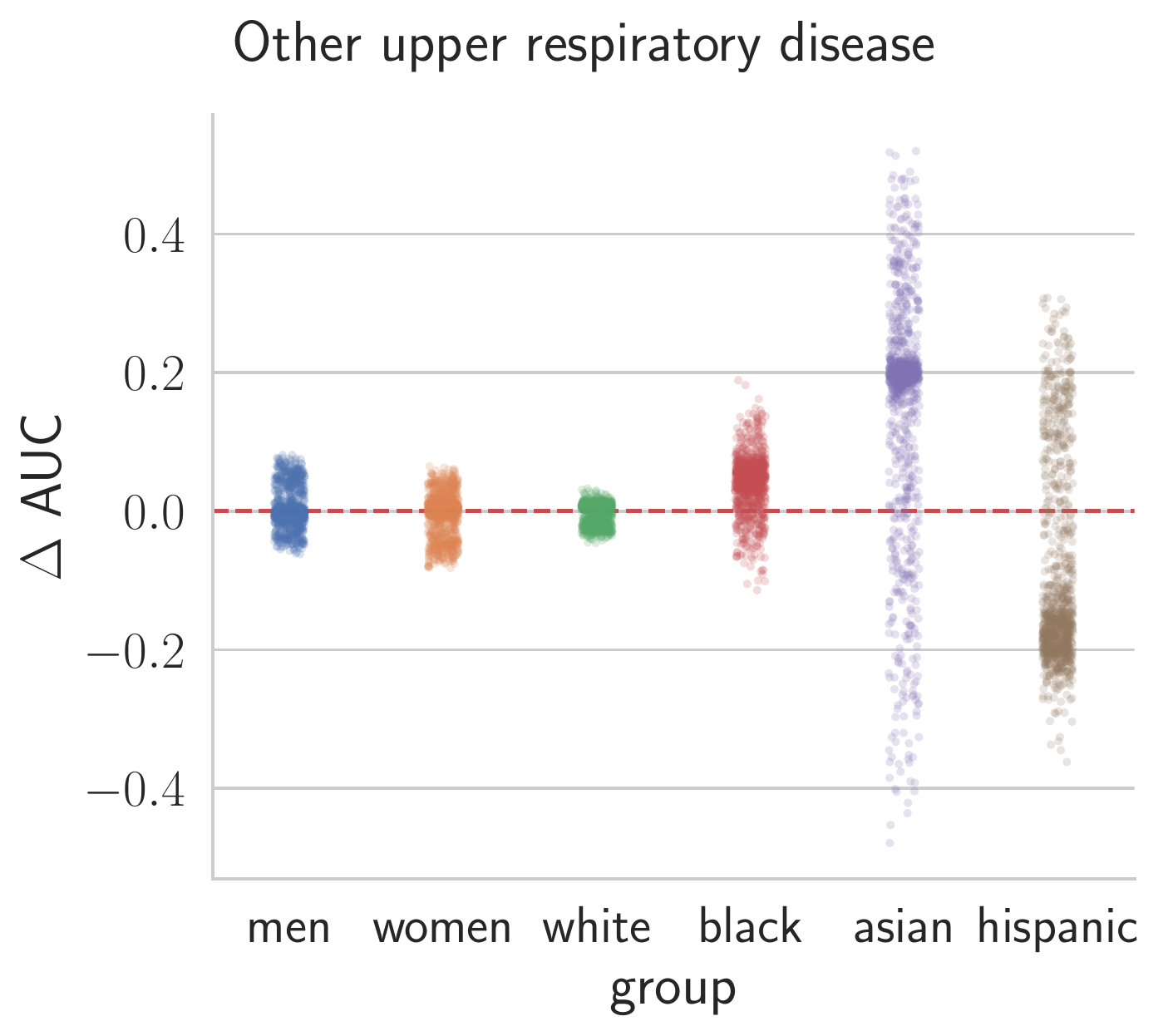}
  \end{minipage}
  \hfill
  \begin{minipage}[t]{0.32\textwidth}
    \includegraphics[width=\textwidth]{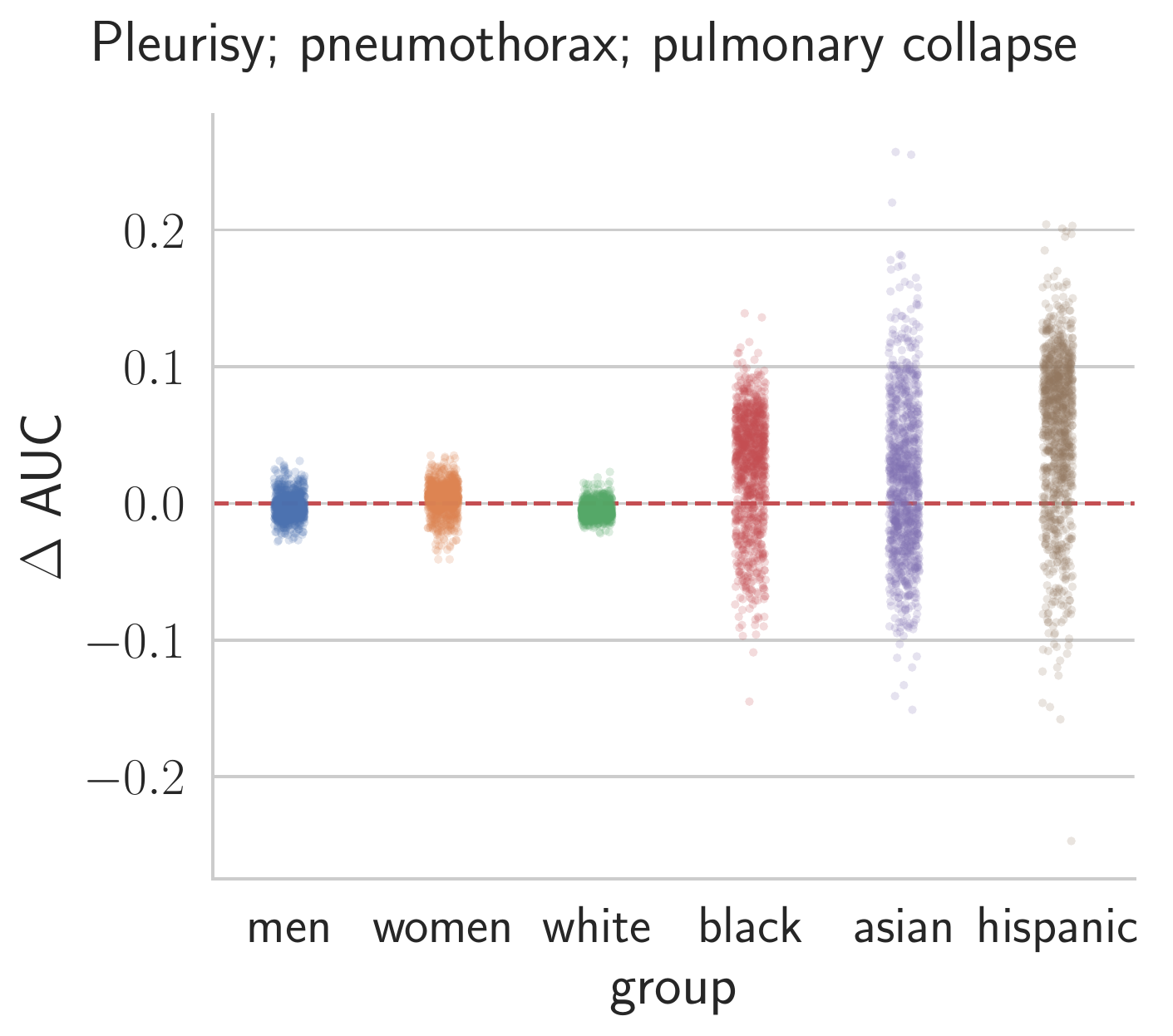}
  \end{minipage}
  \hfill
  \begin{minipage}[t]{0.32\textwidth}
    \includegraphics[width=\textwidth]{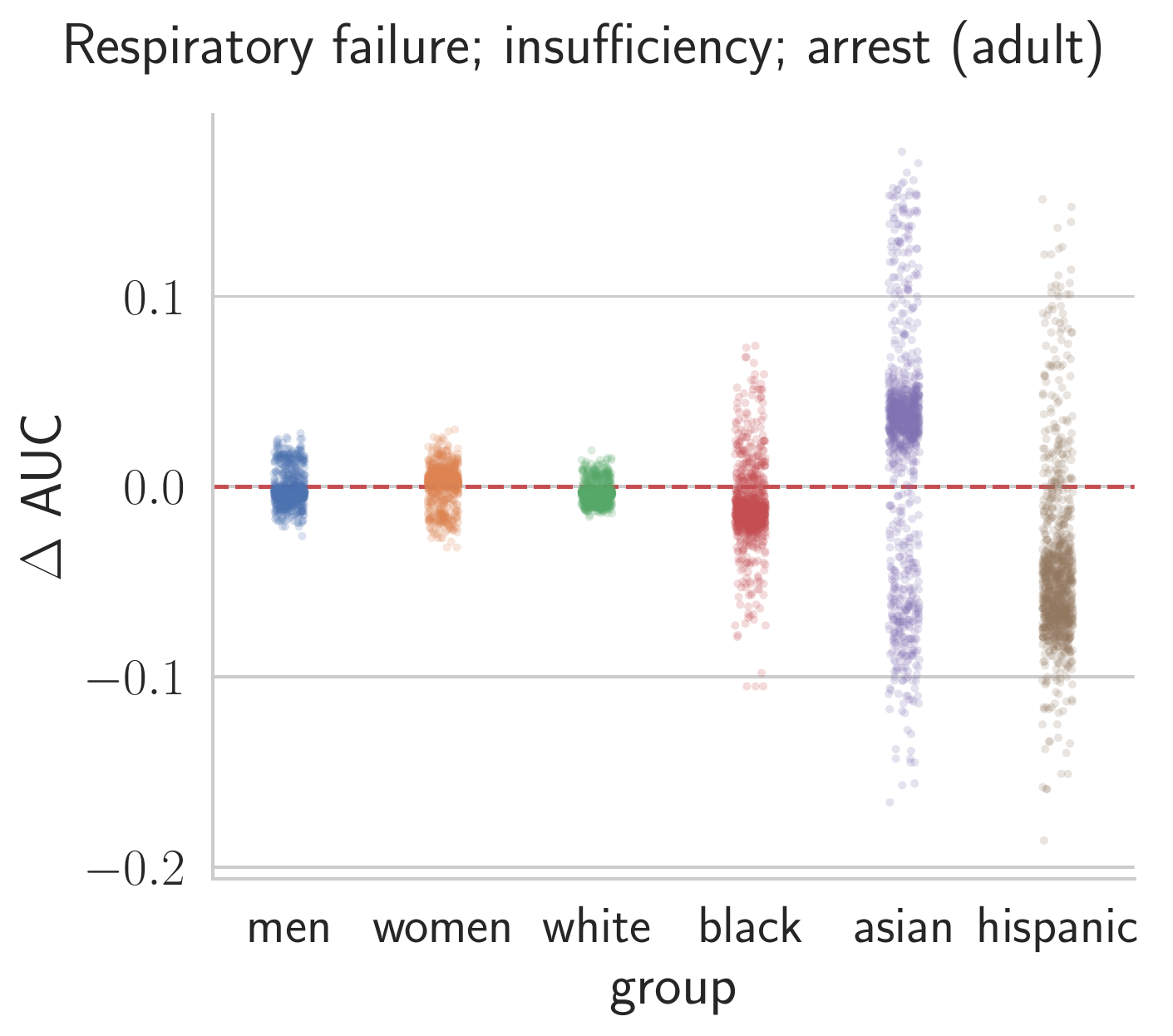}
  \end{minipage}
  \hfill
  \begin{minipage}[t]{0.32\textwidth}
    \includegraphics[width=\textwidth]{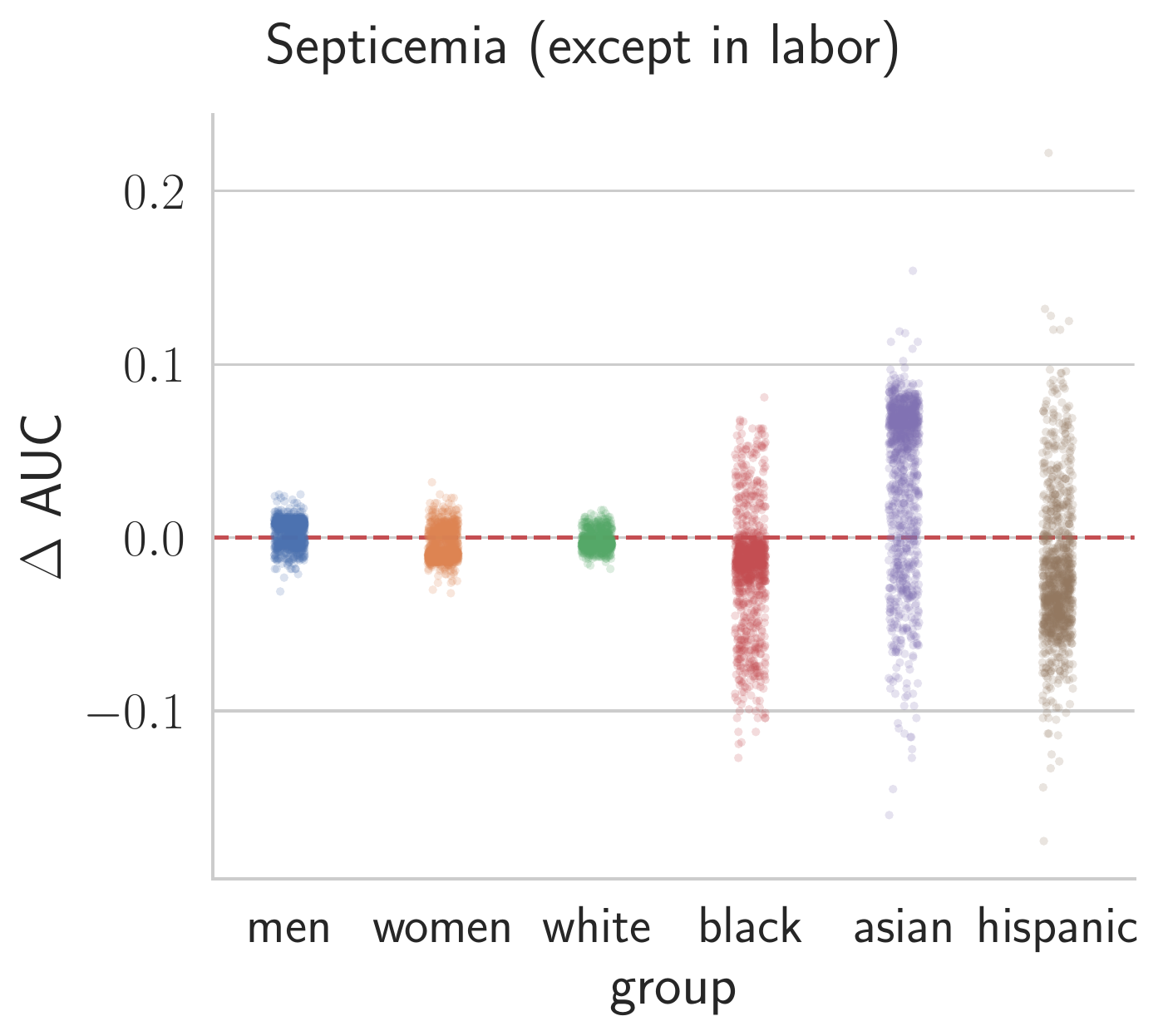}
  \end{minipage}
  \hfill
  \begin{minipage}[t]{0.32\textwidth}
    \includegraphics[width=\textwidth]{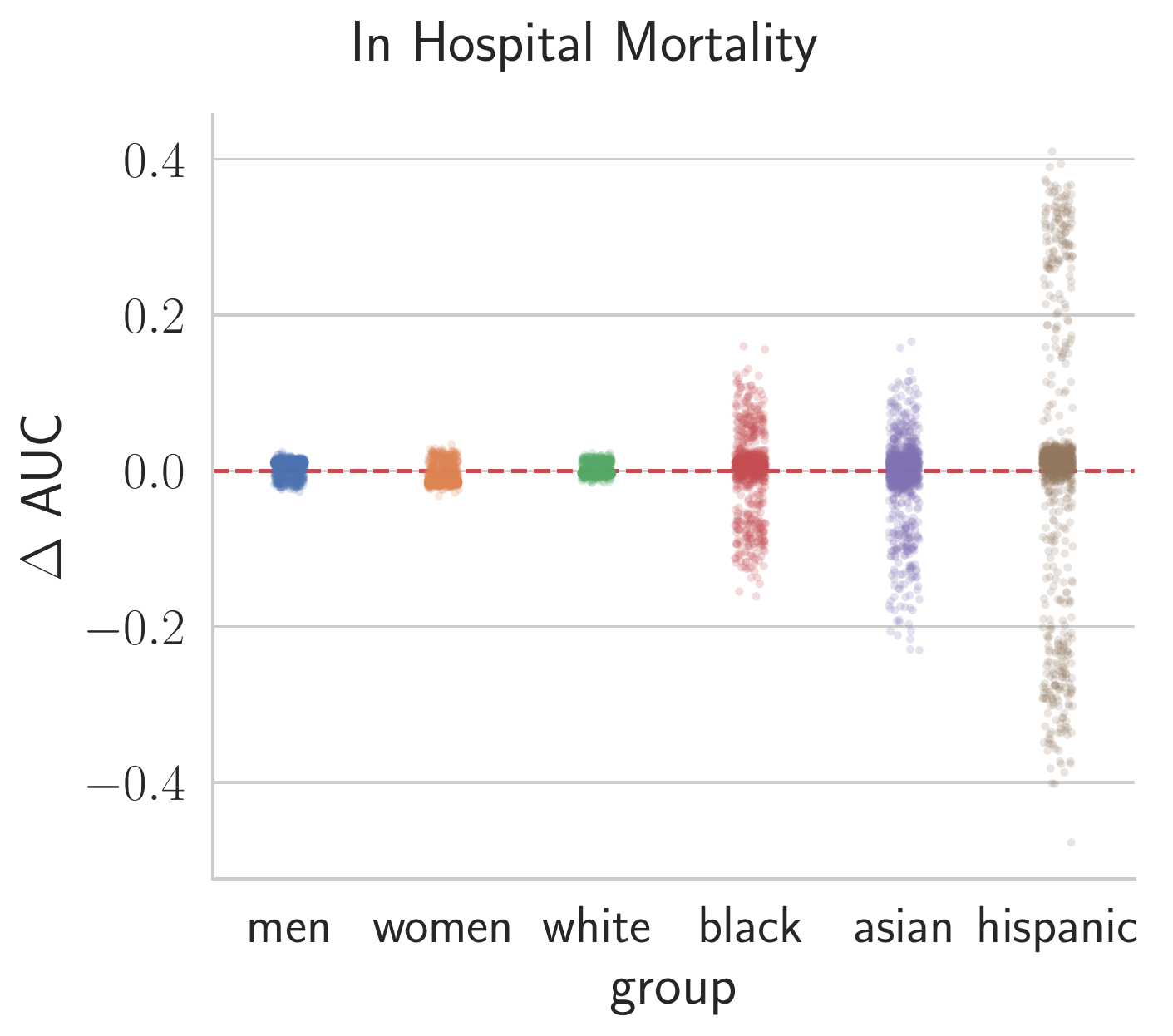}
  \end{minipage}
 \hfill 
 \begin{minipage}[t]{0.42\textwidth}
    \includegraphics[width=\textwidth]{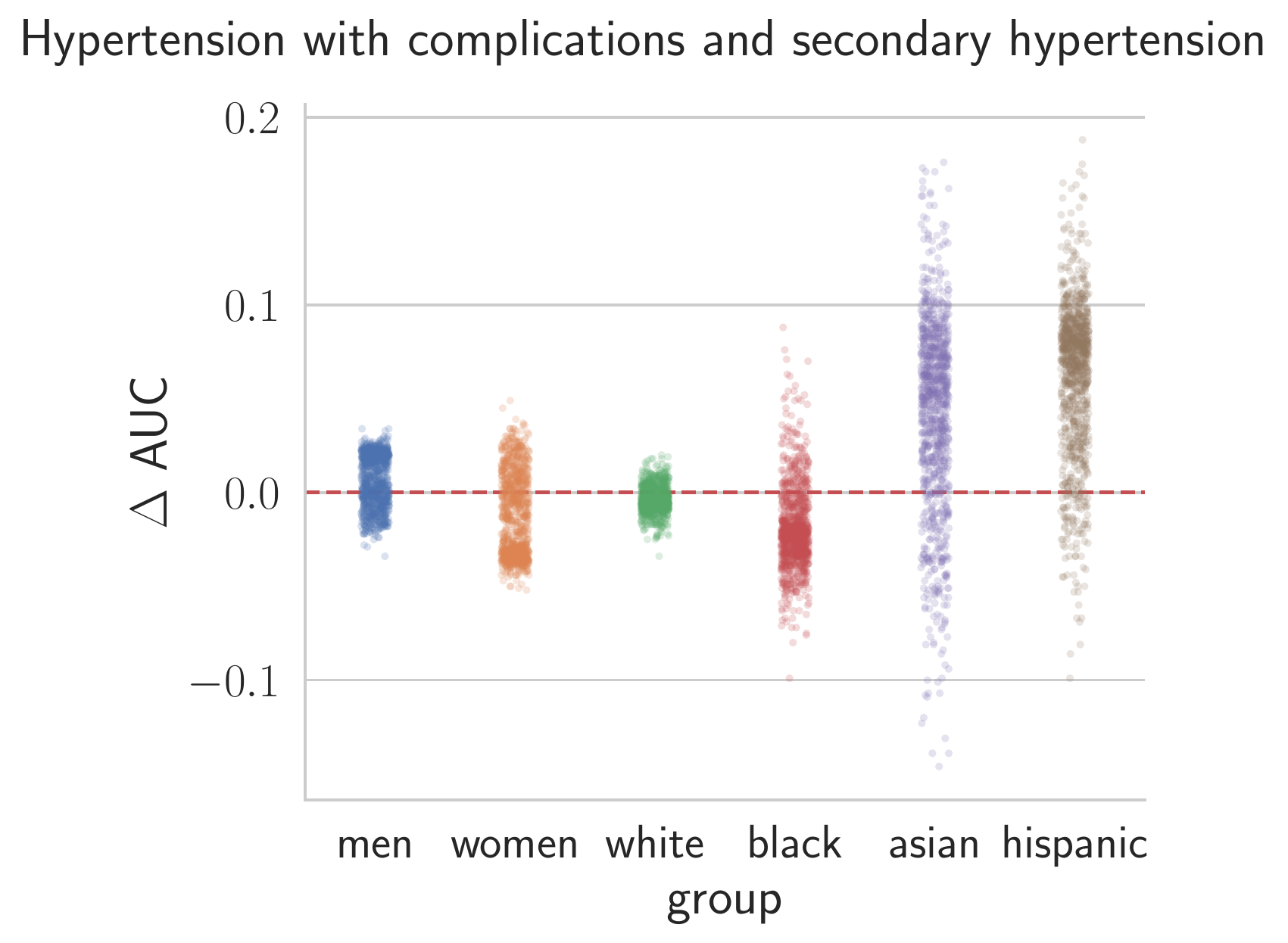}
  \end{minipage}
  \hfill
    \begin{minipage}[t]{0.52\textwidth}
    \includegraphics[width=\textwidth]{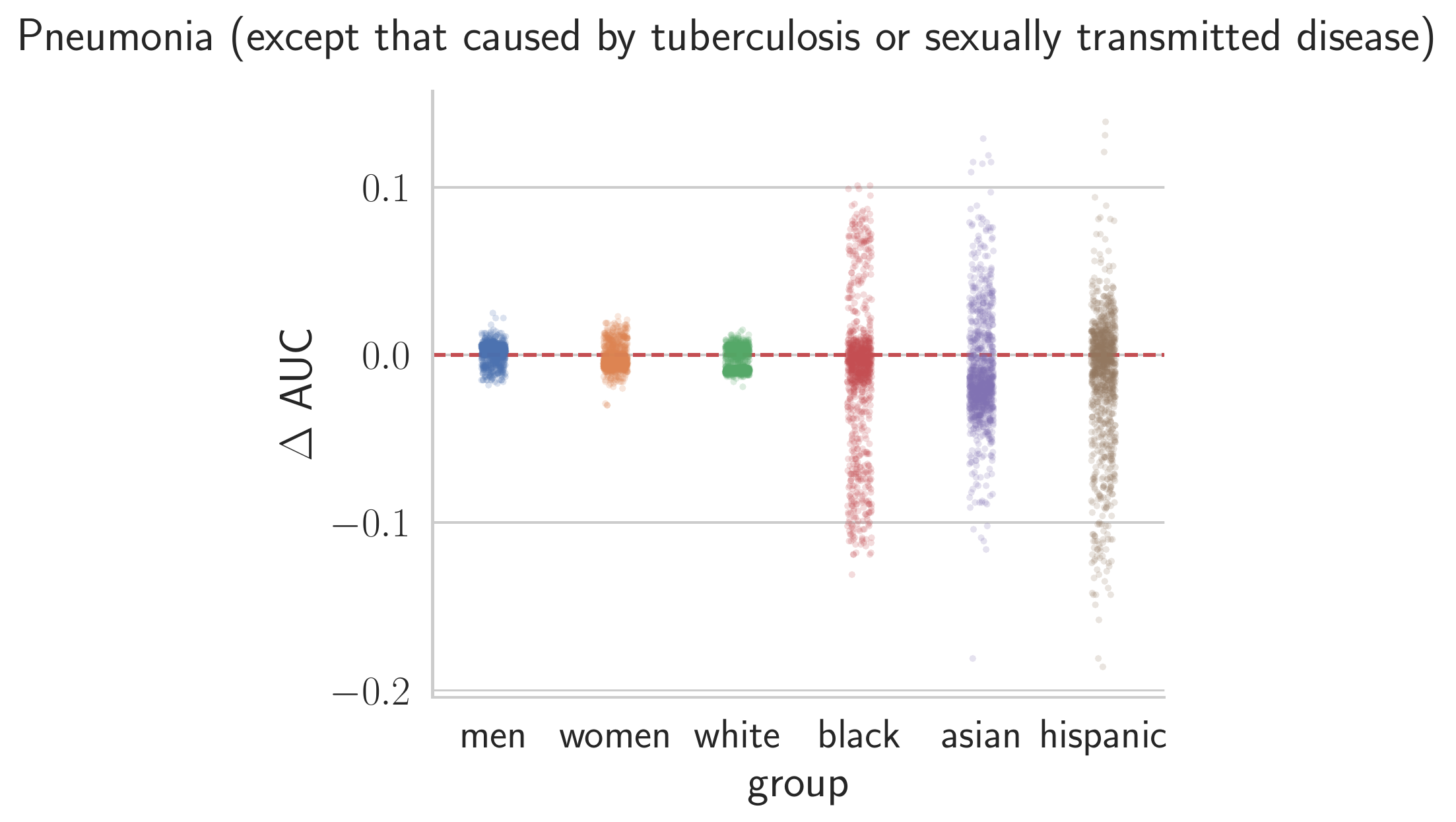}
  \end{minipage}
 \caption{Differences relative to overall performance as a function of random seeds for each subgroup }
 \label{fig:deltas_2}
\end{figure*}

\begin{figure*}
  \begin{minipage}[t]{0.48\textwidth}
    \includegraphics[width=\textwidth]{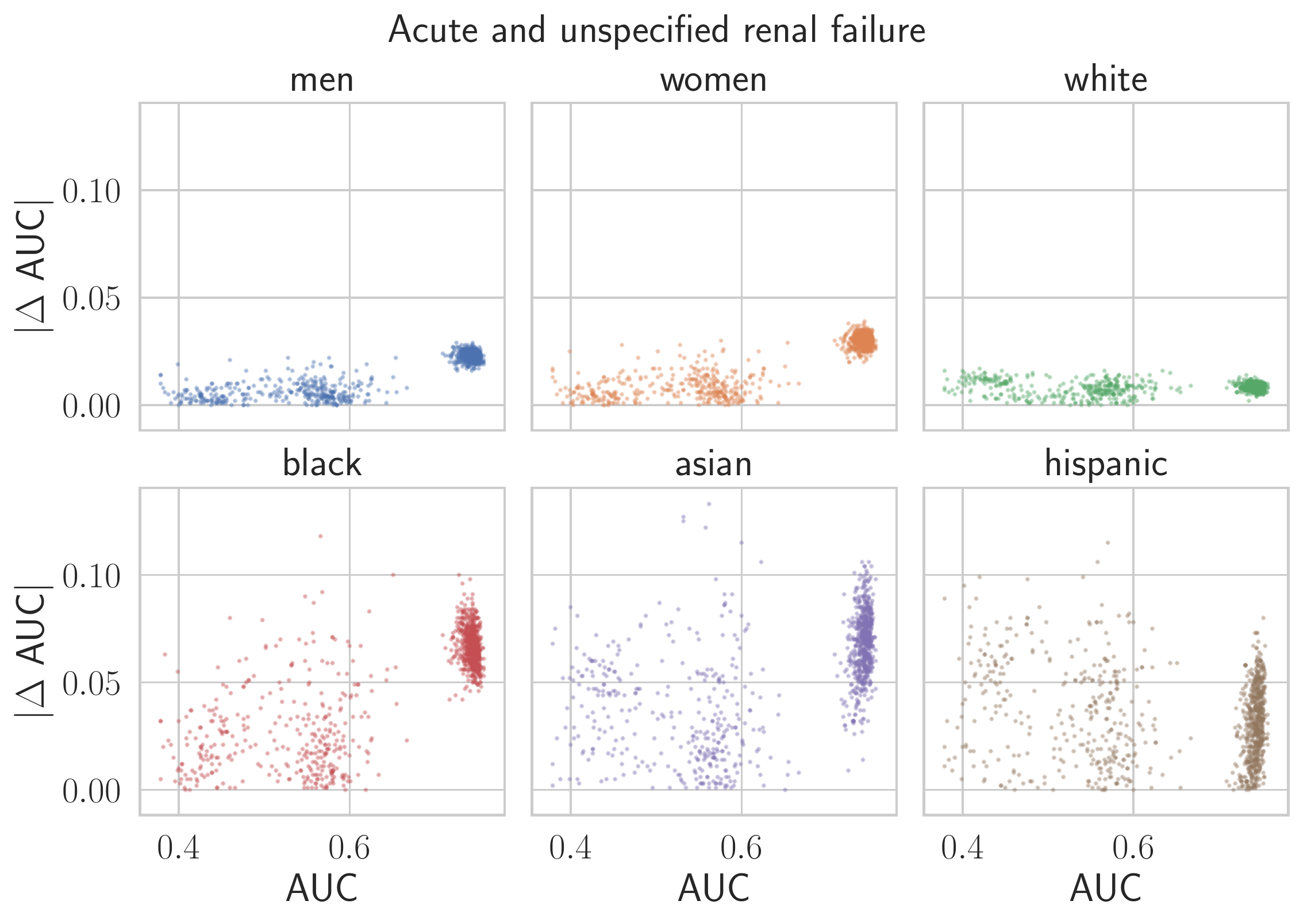}
  \end{minipage}
 \hfill
  \begin{minipage}[t]{0.48\textwidth}
    \includegraphics[width=\textwidth]{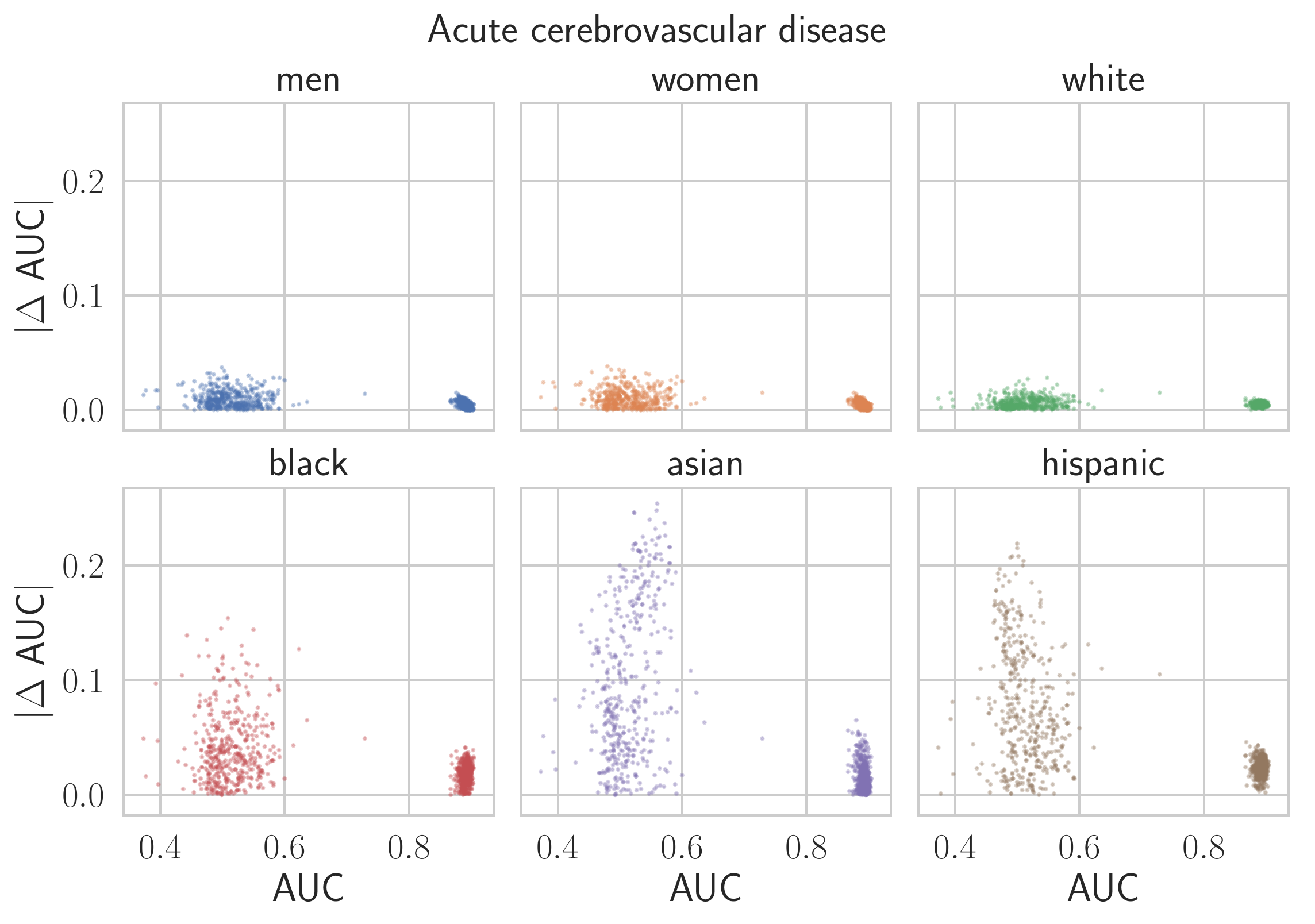}
  \end{minipage}
  \begin{minipage}[t]{0.48\textwidth}
    \includegraphics[width=\textwidth]{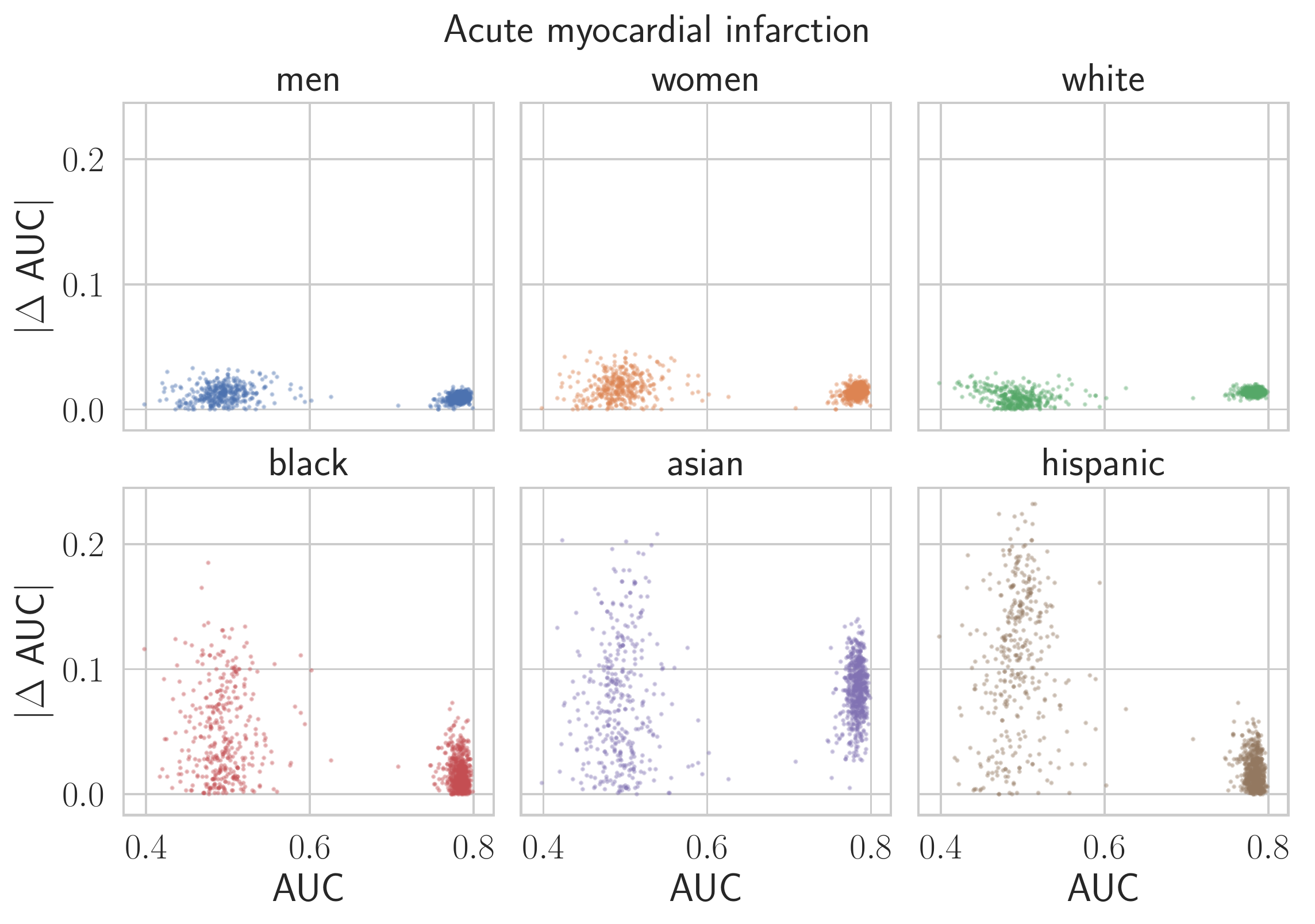}
  \end{minipage}
 \hfill
  \begin{minipage}[t]{0.48\textwidth}
    \includegraphics[width=\textwidth]{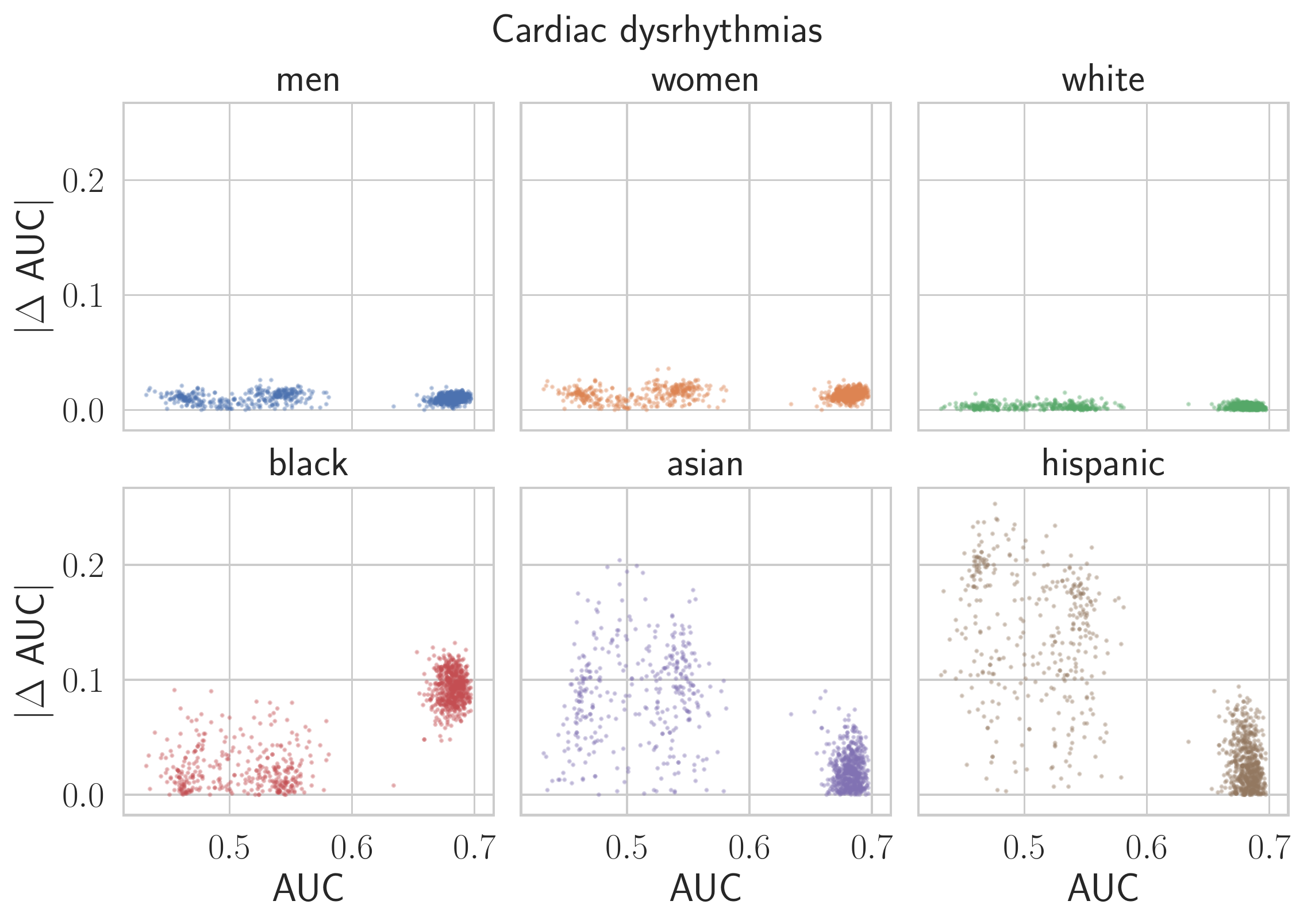}
  \end{minipage}
  \begin{minipage}[t]{0.48\textwidth}
    \includegraphics[width=\textwidth]{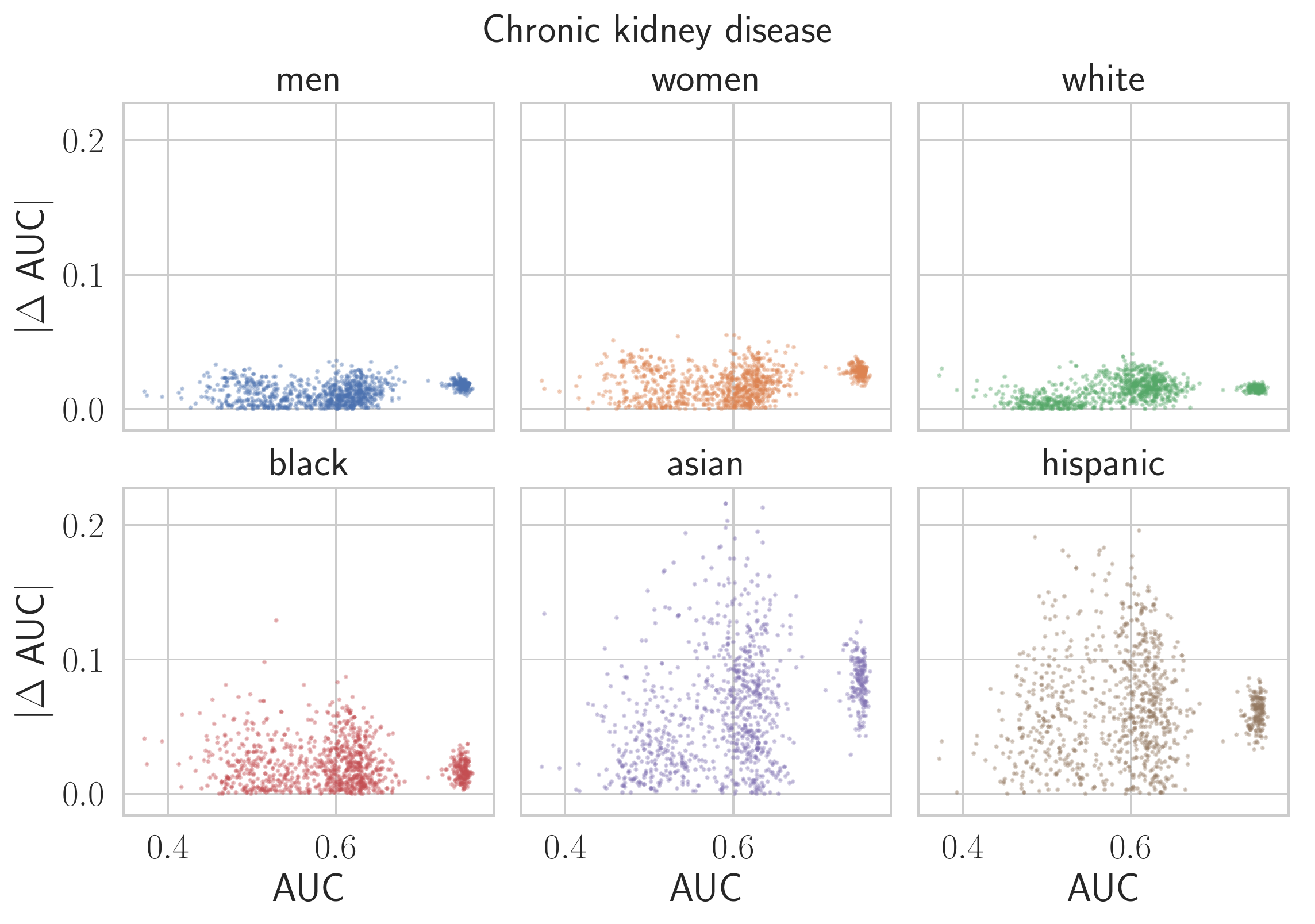}
  \end{minipage}
 \hfill
  \begin{minipage}[t]{0.48\textwidth}
    \includegraphics[width=\textwidth]{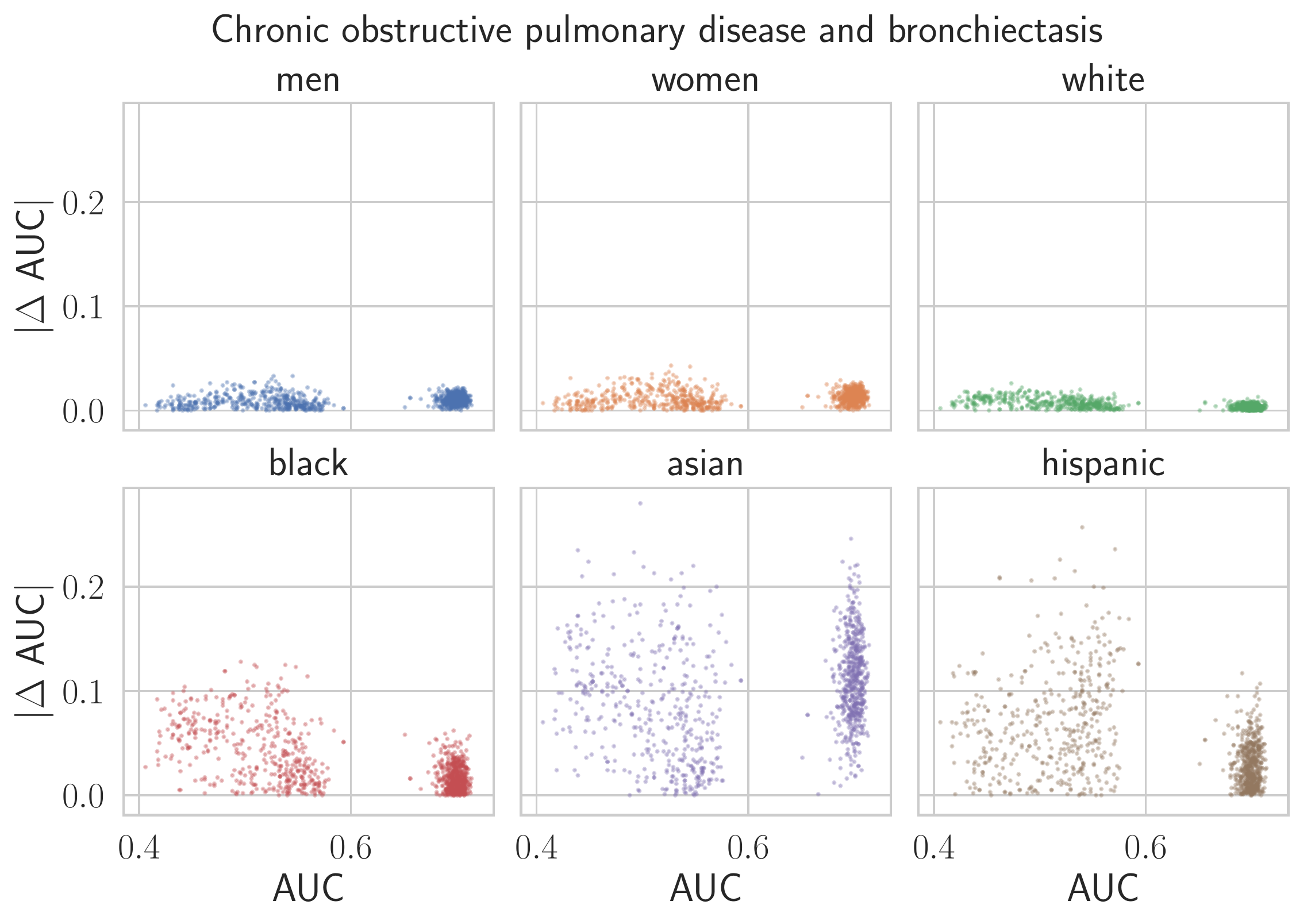}
  \end{minipage}
  \begin{minipage}[t]{0.48\textwidth}
    \includegraphics[width=\textwidth]{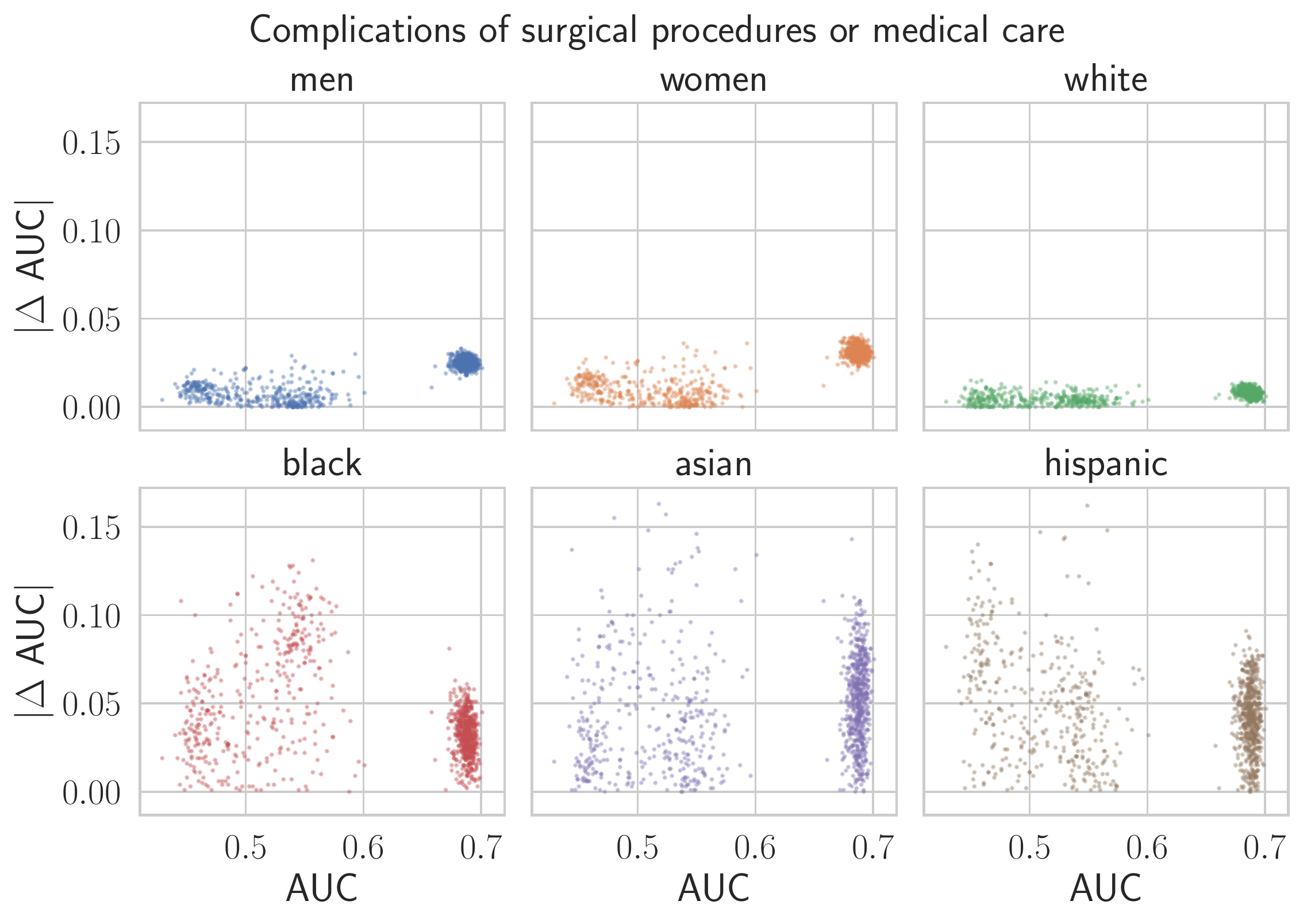}
  \end{minipage}
 \hfill
  \begin{minipage}[t]{0.48\textwidth}
    \includegraphics[width=\textwidth]{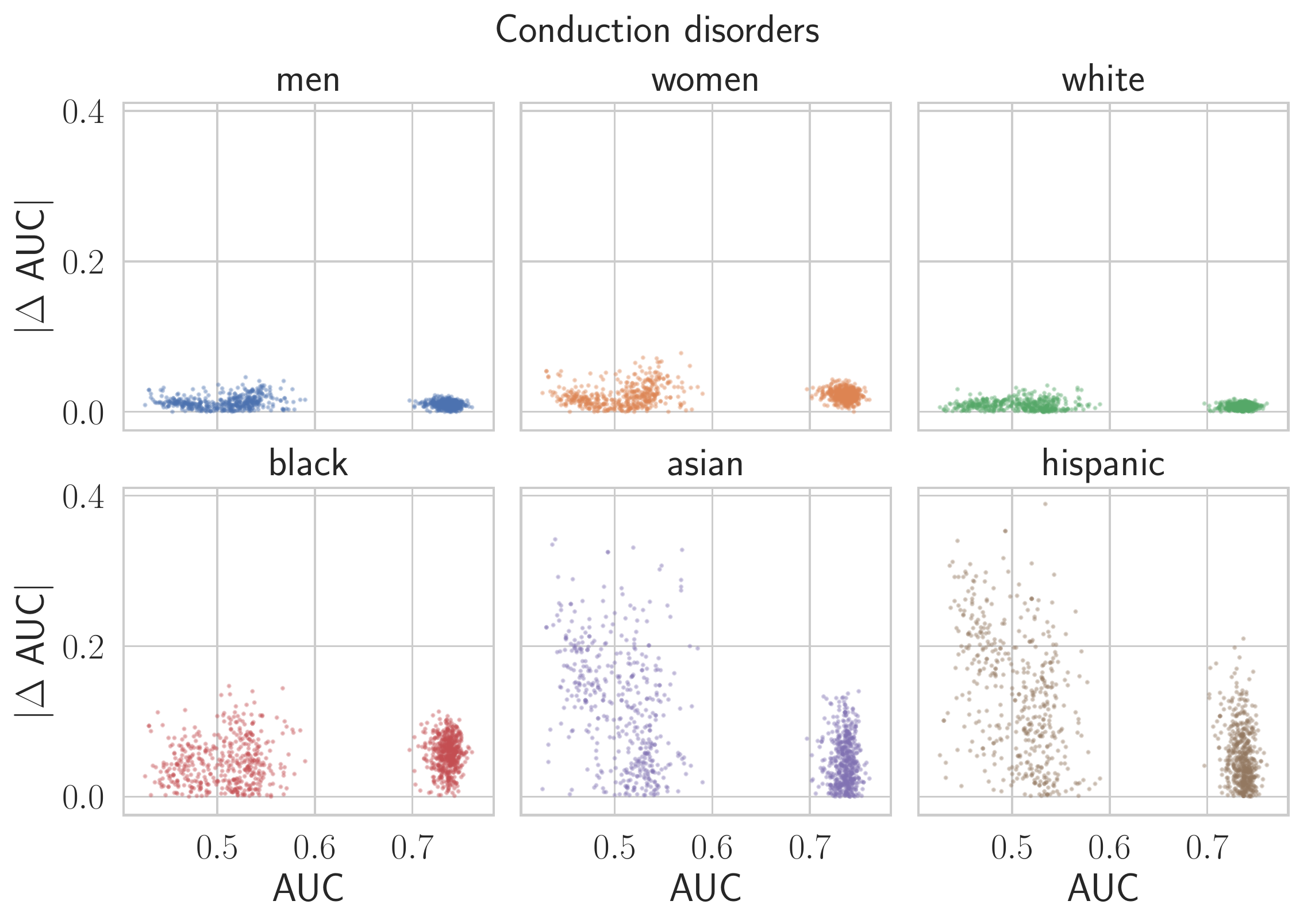}
  \end{minipage}
  \caption{Correlations between overall performance and subgroup performance $\Delta$}
  \label{fig:scatters_1}
\end{figure*}

\begin{figure*}
  \begin{minipage}[t]{0.48\textwidth}
    \includegraphics[width=\textwidth]{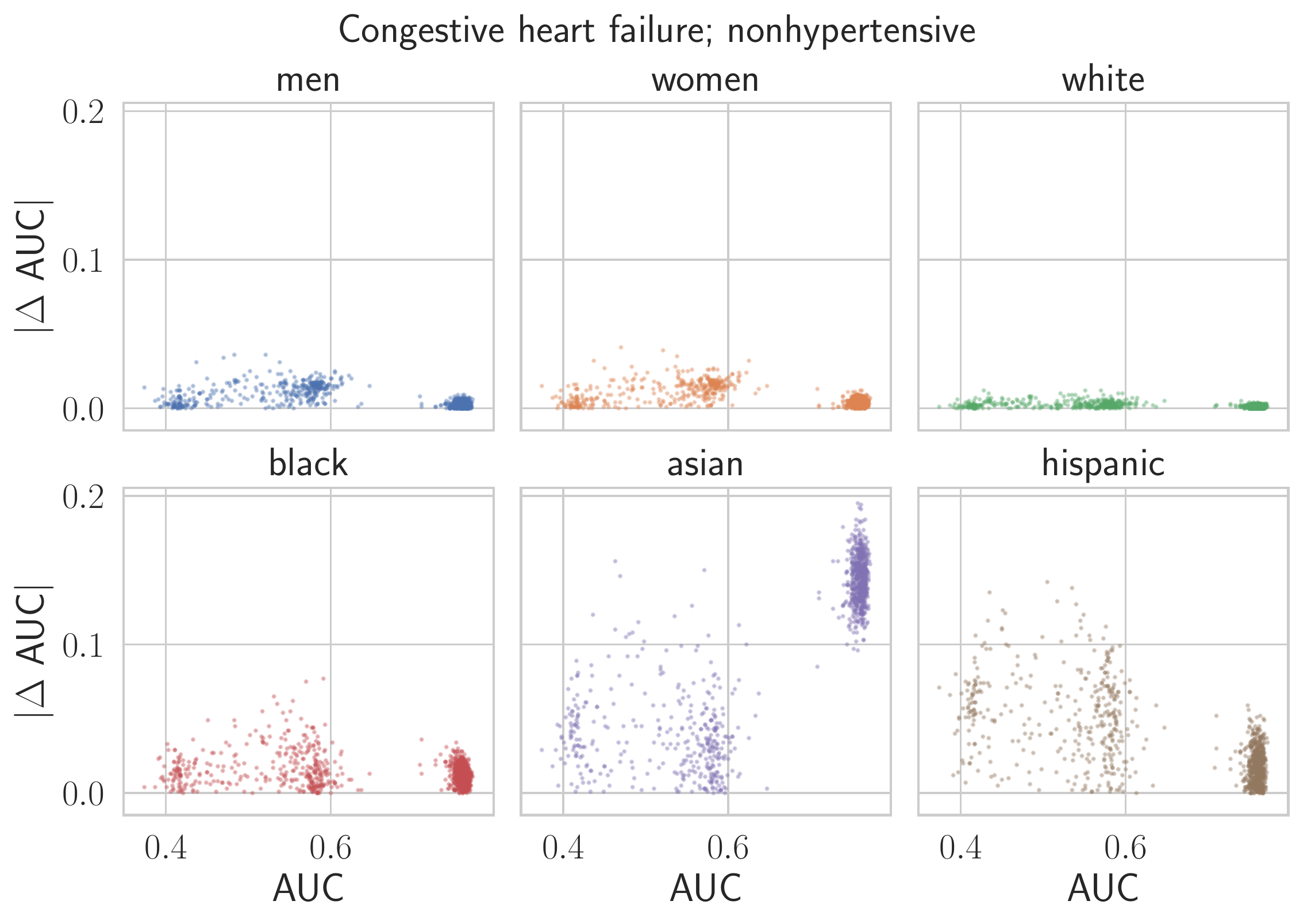}
  \end{minipage}
 \hfill
  \begin{minipage}[t]{0.48\textwidth}
    \includegraphics[width=\textwidth]{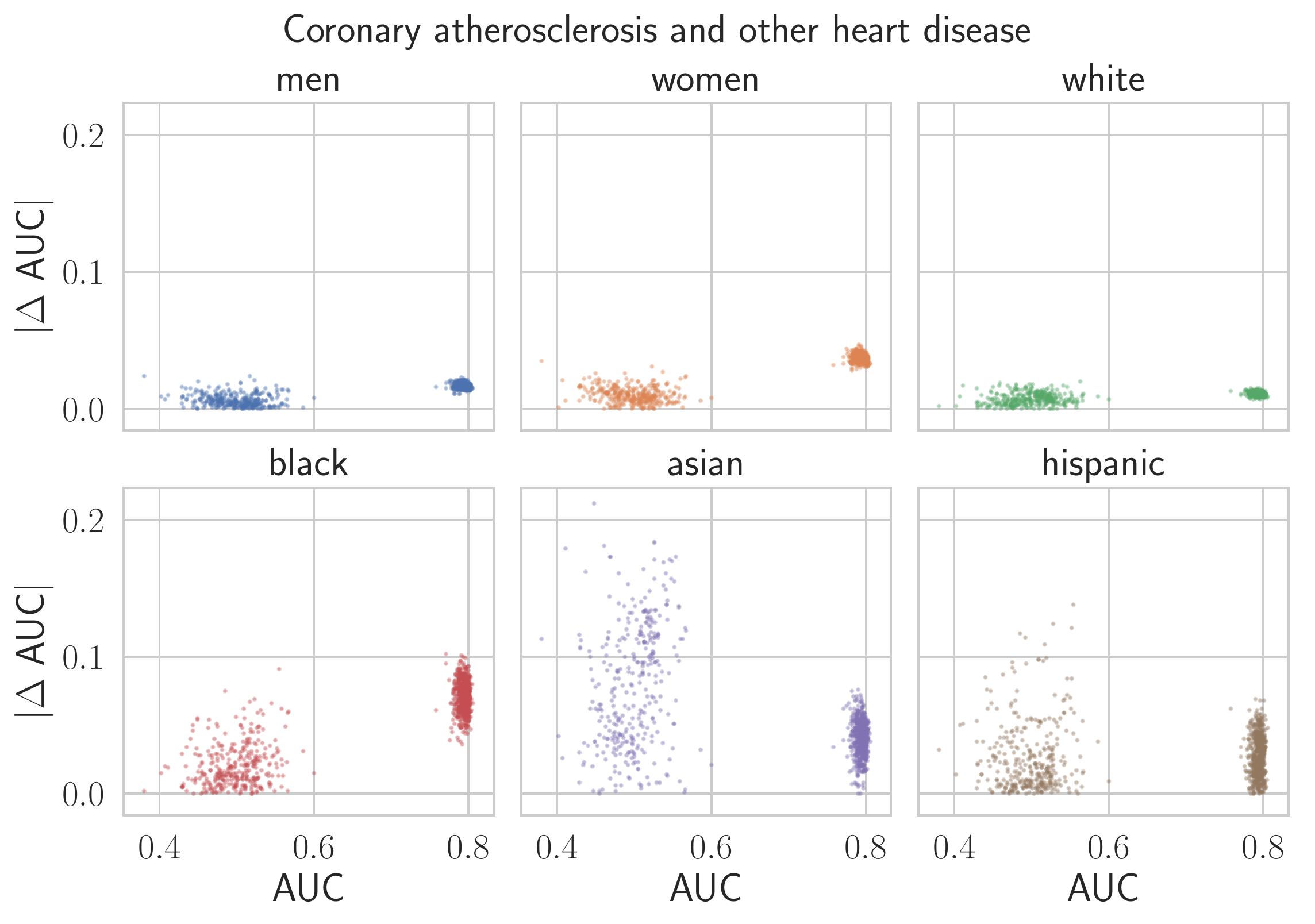}
  \end{minipage}
  \begin{minipage}[t]{0.48\textwidth}
    \includegraphics[width=\textwidth]{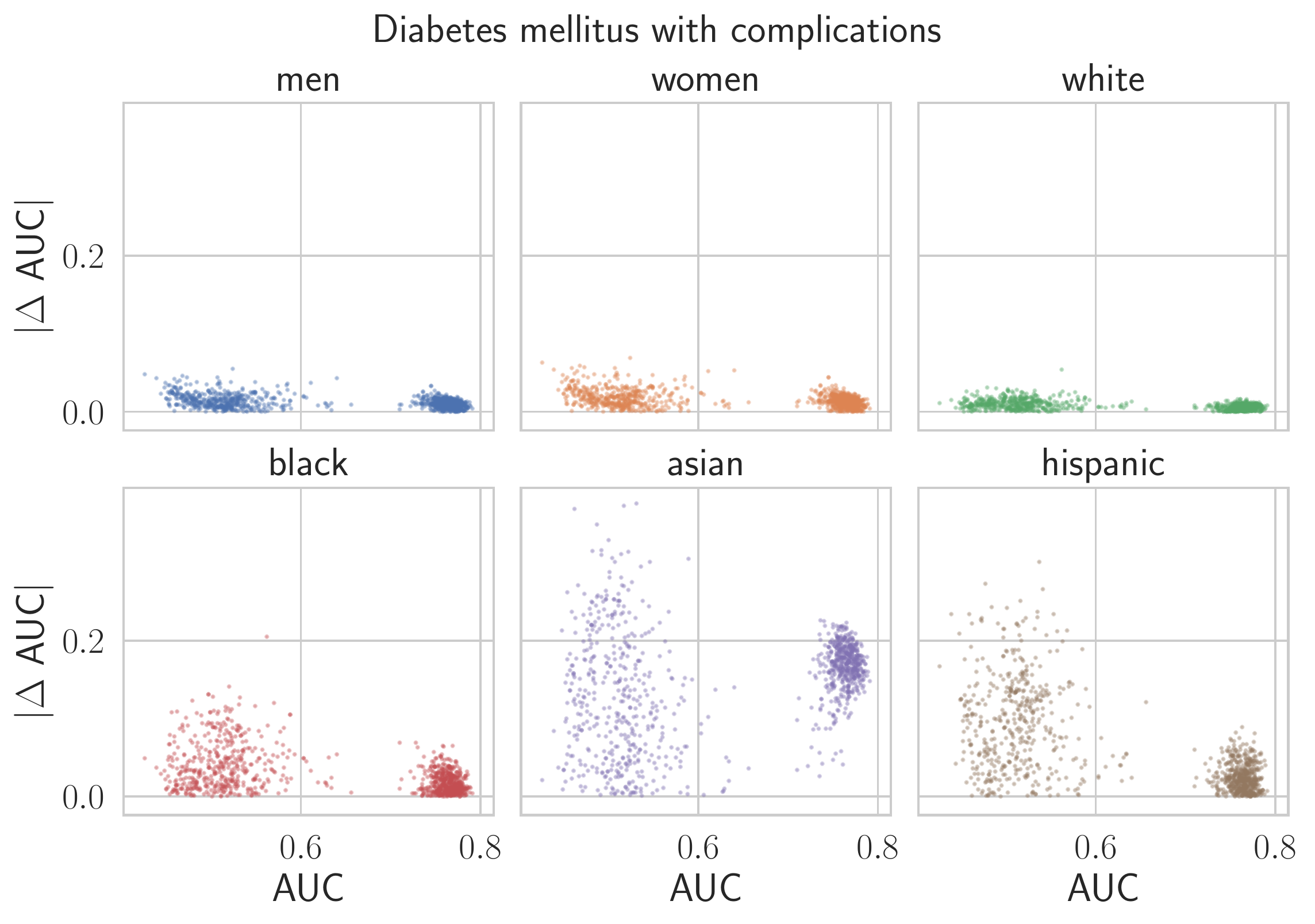}
  \end{minipage}
 \hfill
  \begin{minipage}[t]{0.48\textwidth}
    \includegraphics[width=\textwidth]{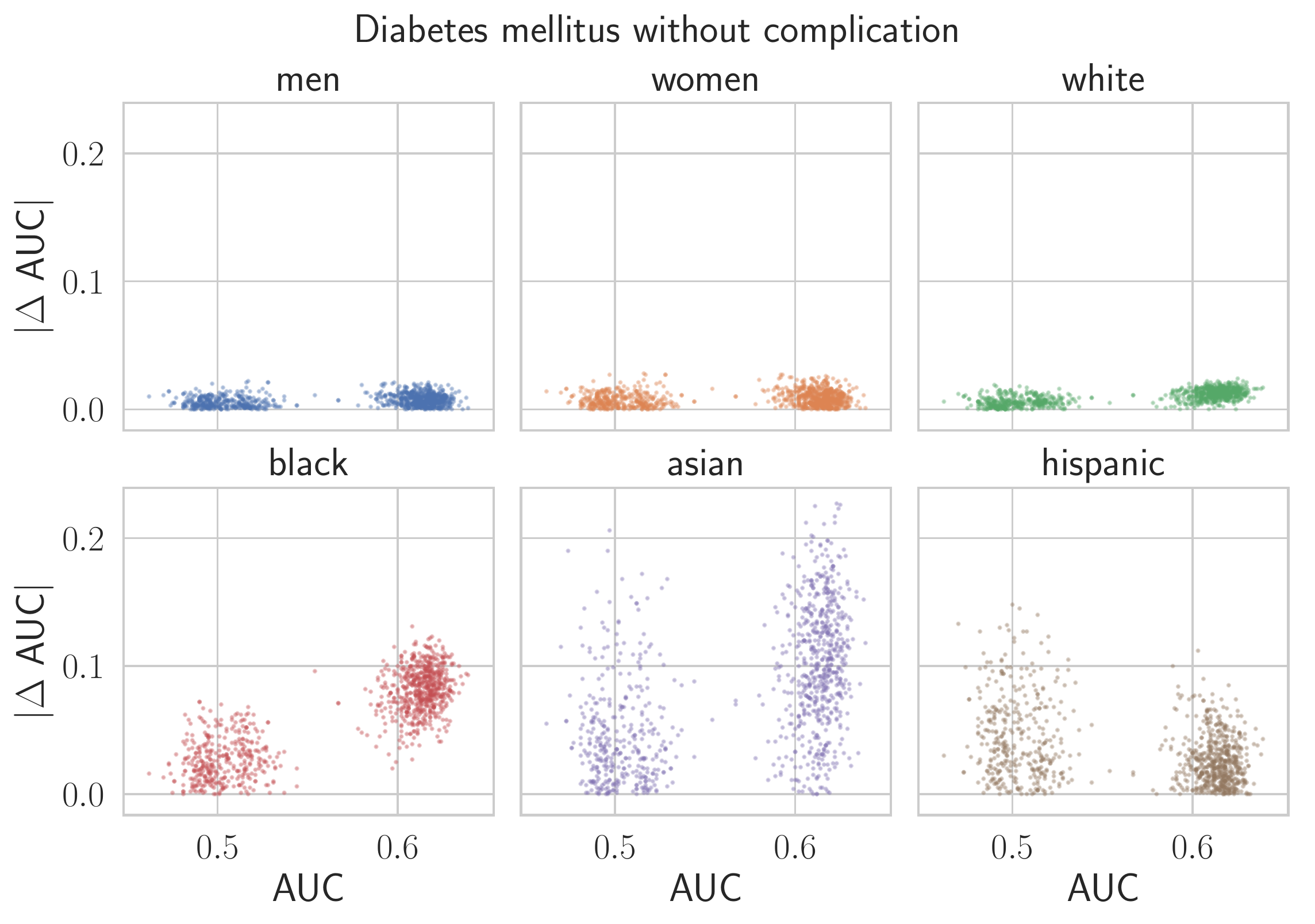}
  \end{minipage}
  \begin{minipage}[t]{0.48\textwidth}
    \includegraphics[width=\textwidth]{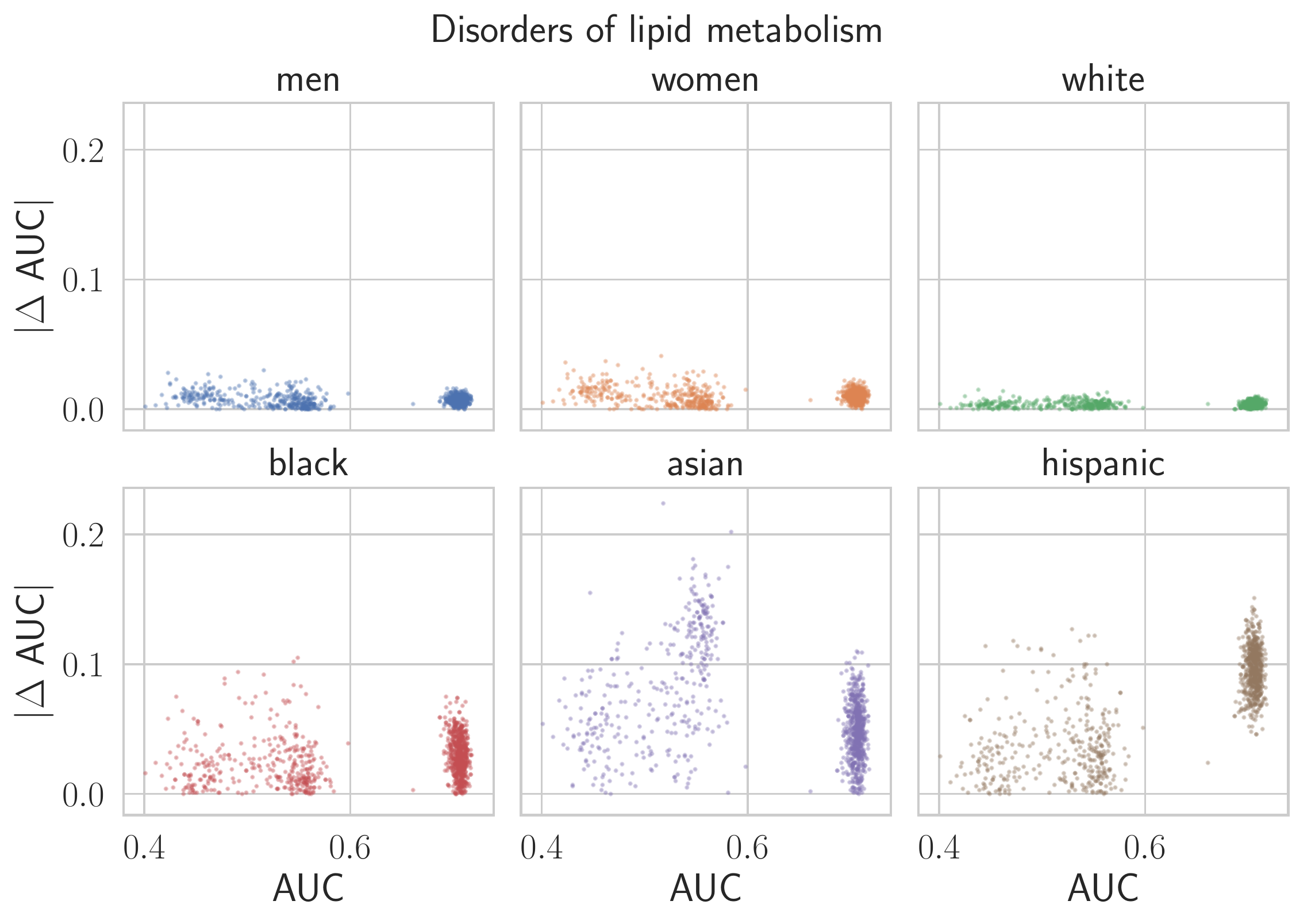}
  \end{minipage}
 \hfill
  \begin{minipage}[t]{0.48\textwidth}
    \includegraphics[width=\textwidth]{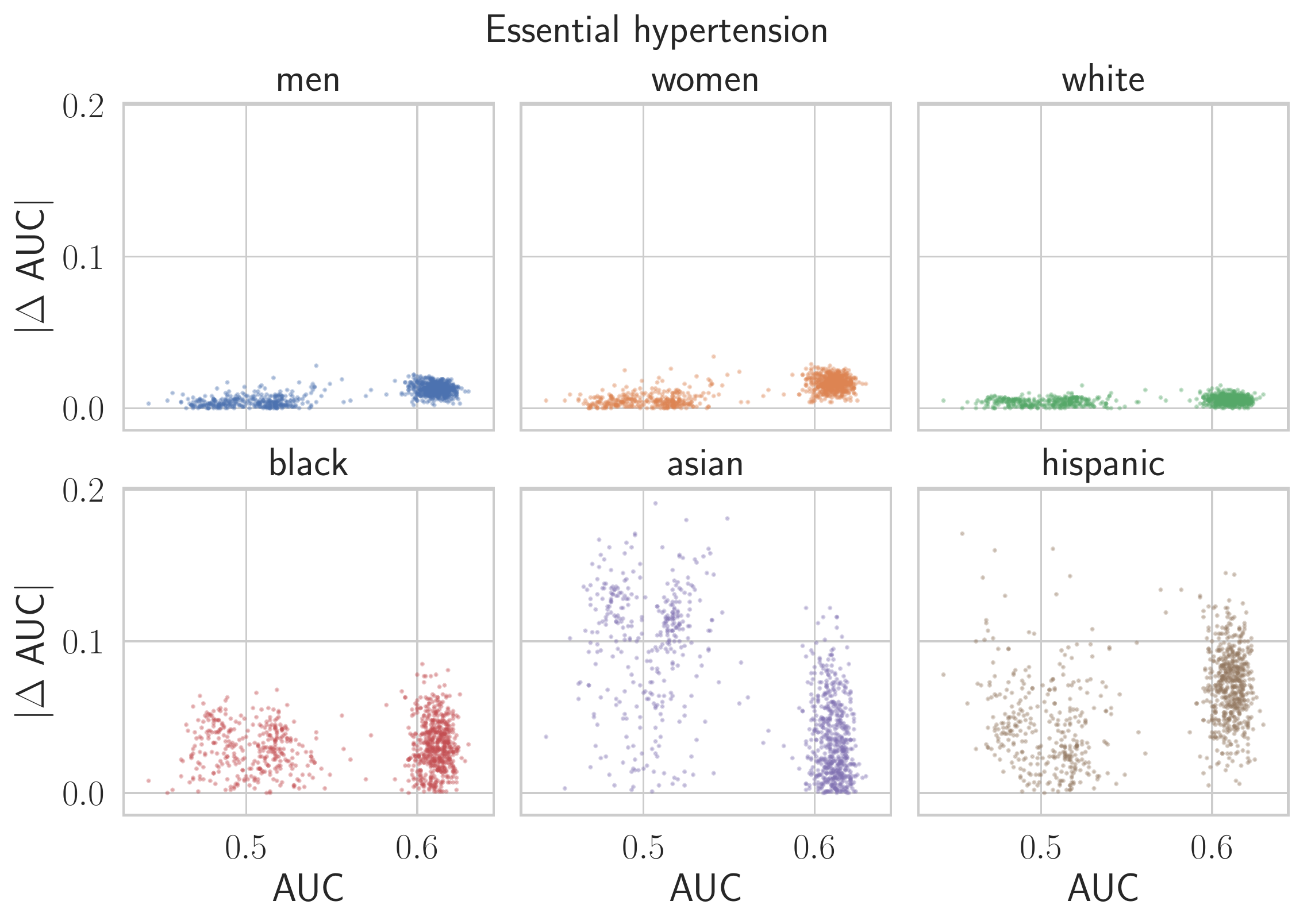}
  \end{minipage}
  \begin{minipage}[t]{0.48\textwidth}
    \includegraphics[width=\textwidth]{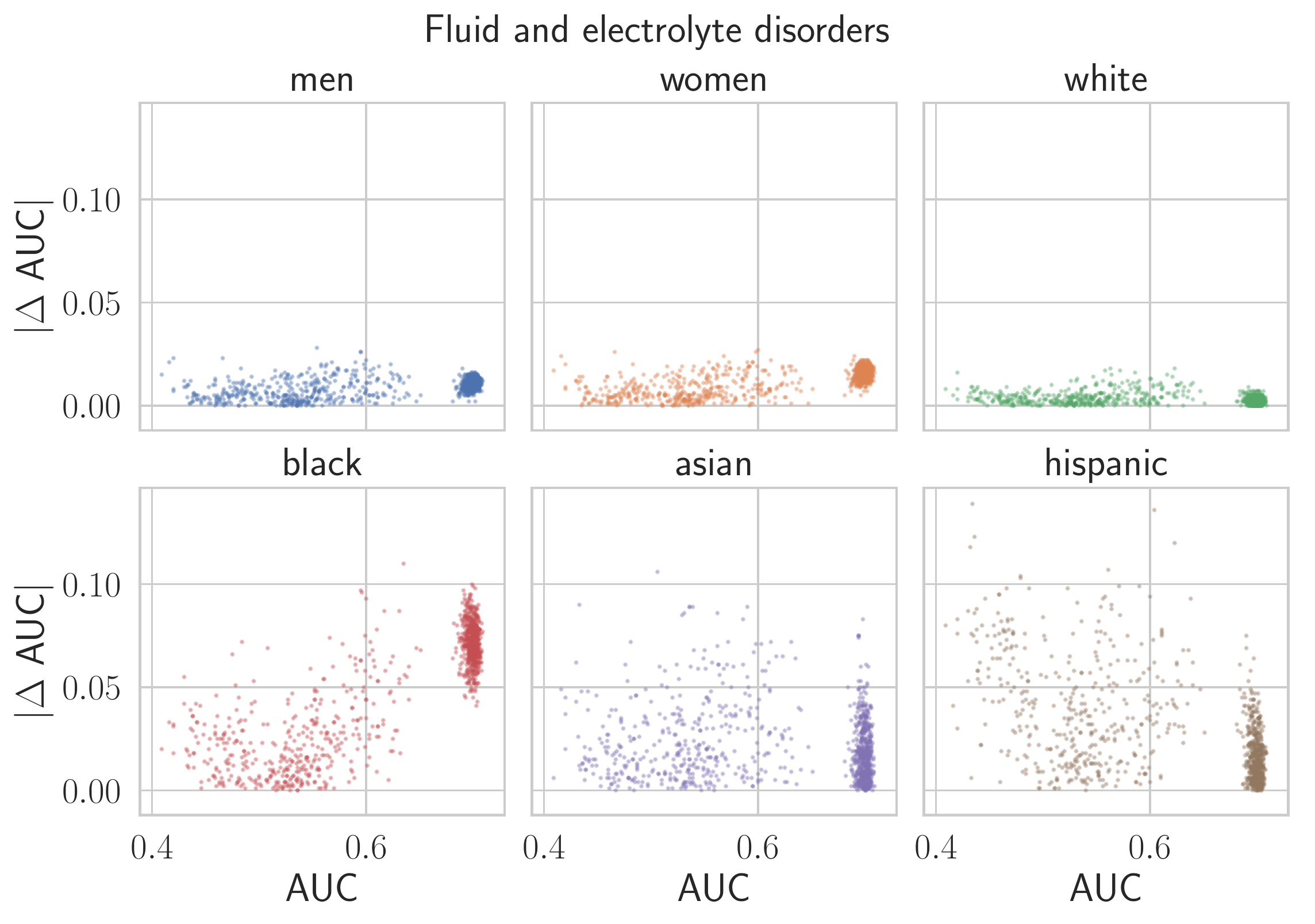}
  \end{minipage}
 \hfill
  \begin{minipage}[t]{0.48\textwidth}
    \includegraphics[width=\textwidth]{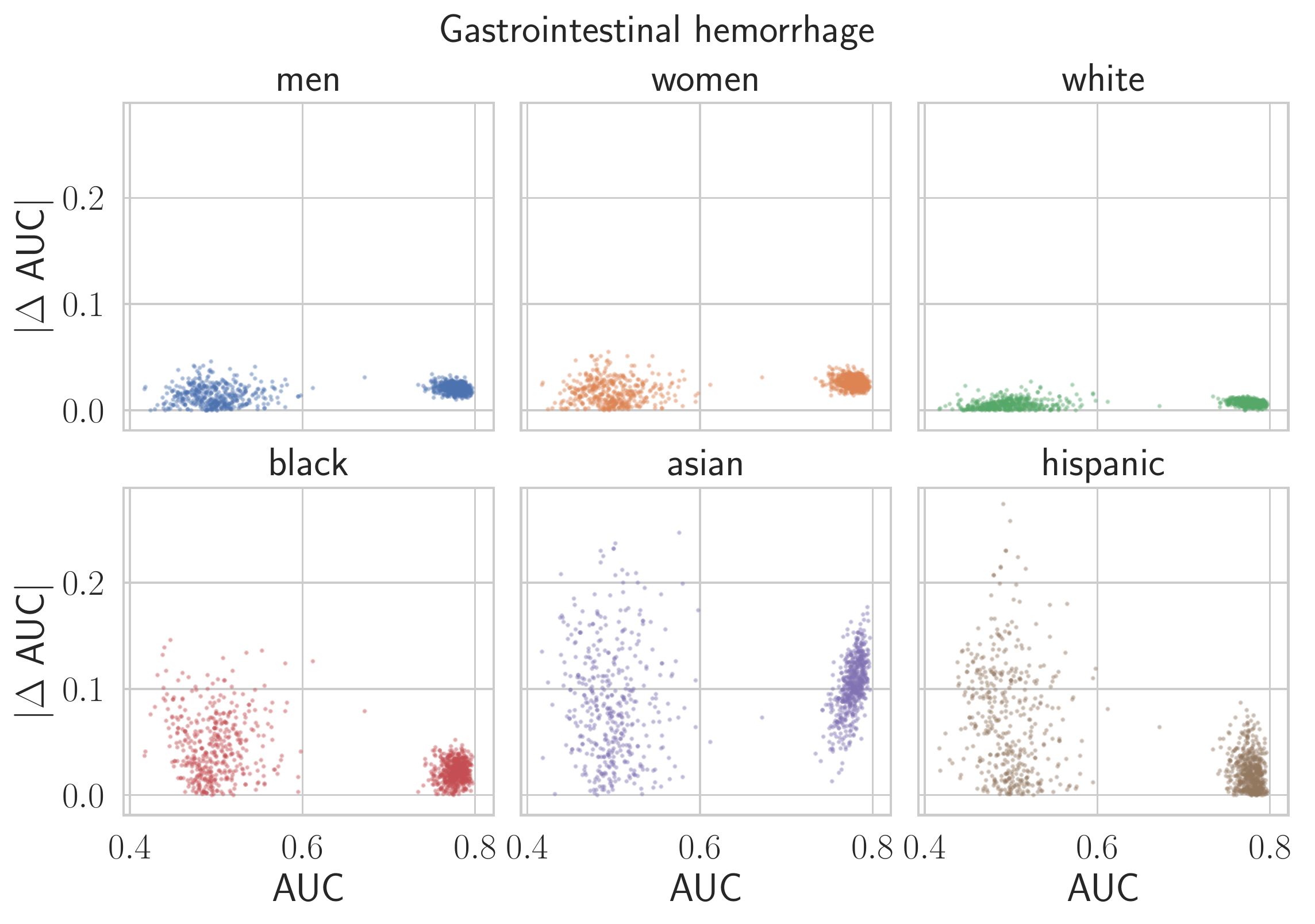}
  \end{minipage}
  \caption{Correlations between overall performance and subgroup performance $\Delta$}
  \label{fig:scatters_2}
\end{figure*}

\begin{figure*}[t!]
  \begin{minipage}[t]{0.48\textwidth}
    \includegraphics[width=\textwidth]{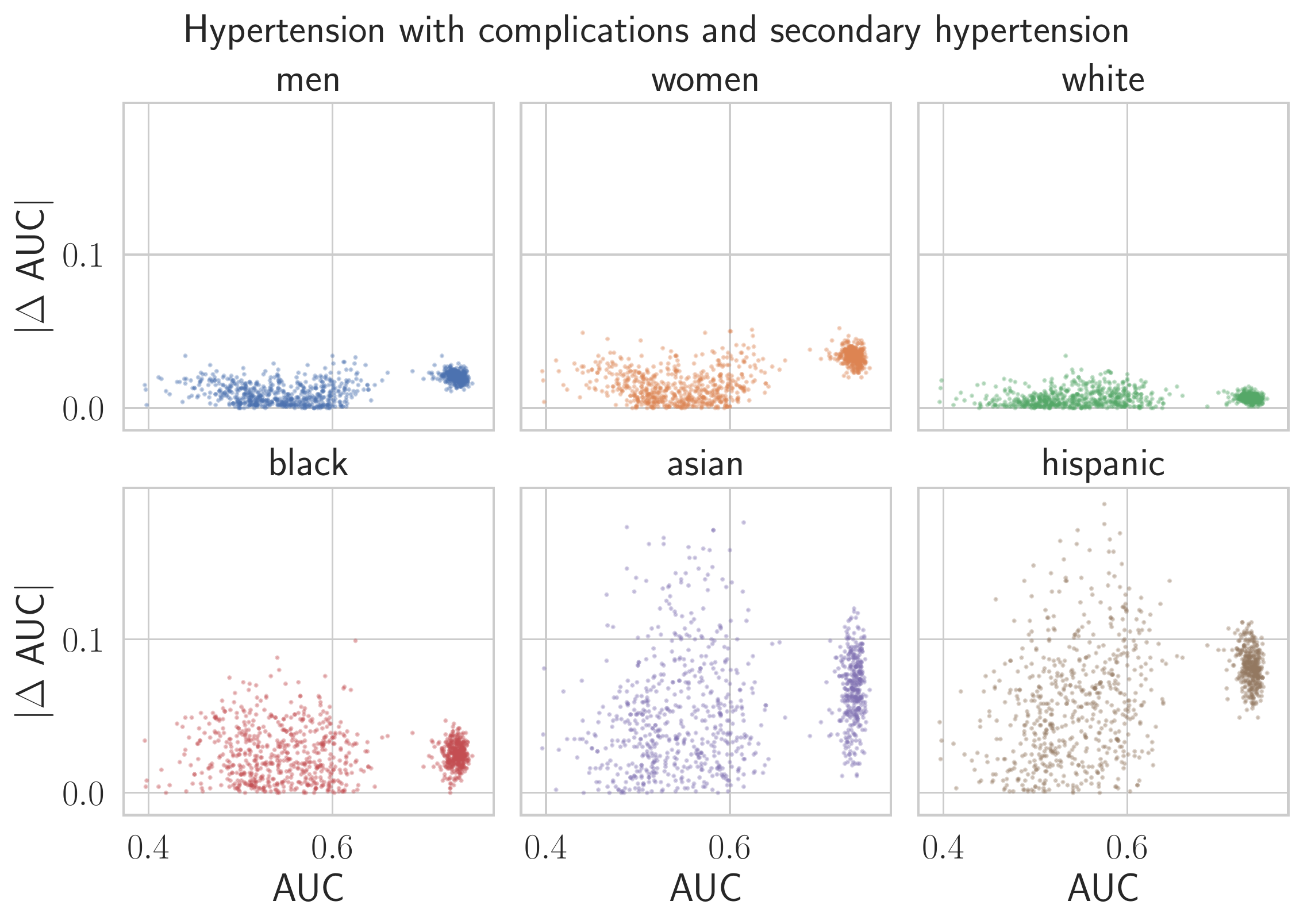}
  \end{minipage}
 \hfill
  \begin{minipage}[t]{0.48\textwidth}
    \includegraphics[width=\textwidth]{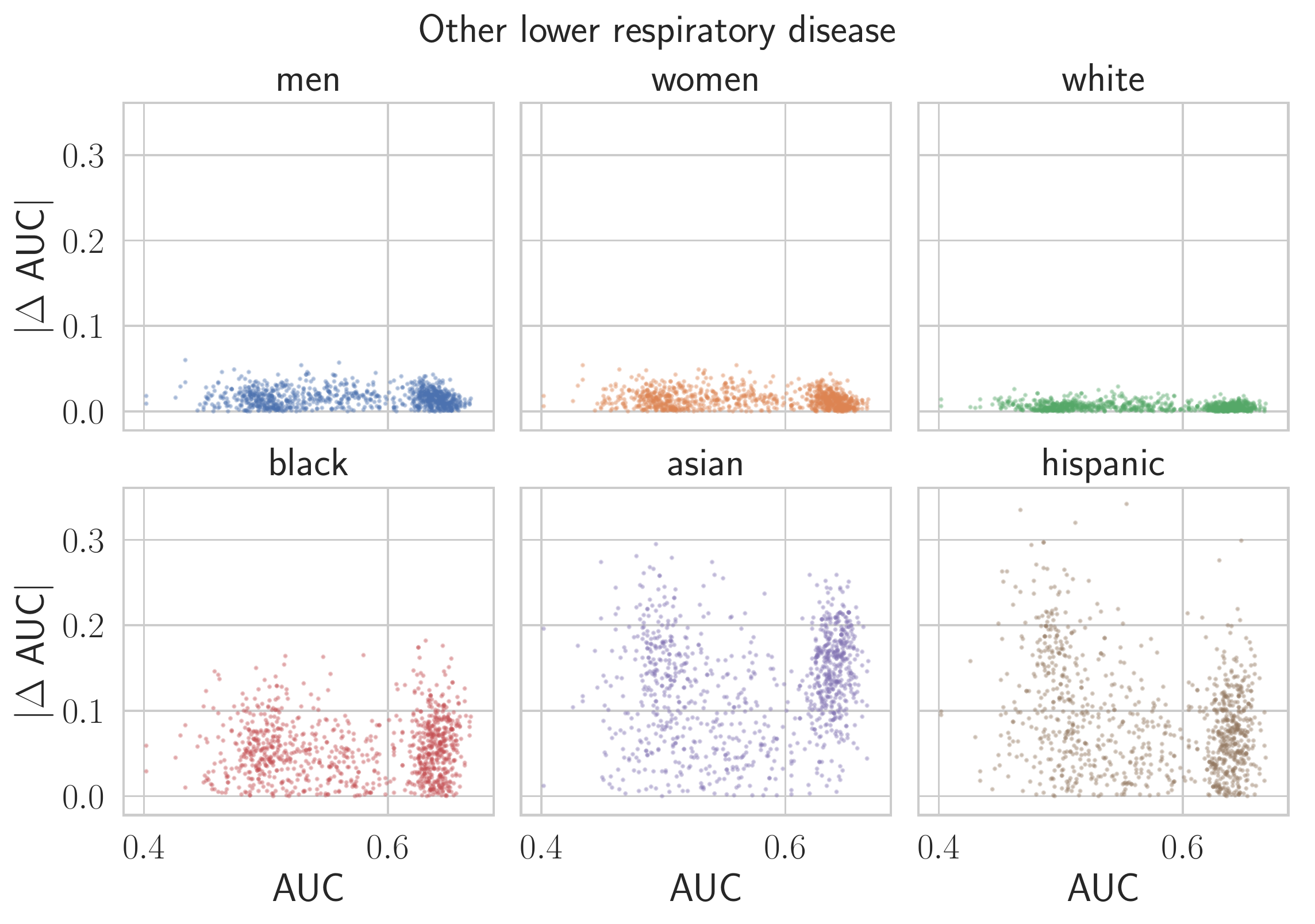}
  \end{minipage}
 \hfill
  \begin{minipage}[t]{0.48\textwidth}
    \includegraphics[width=\textwidth]{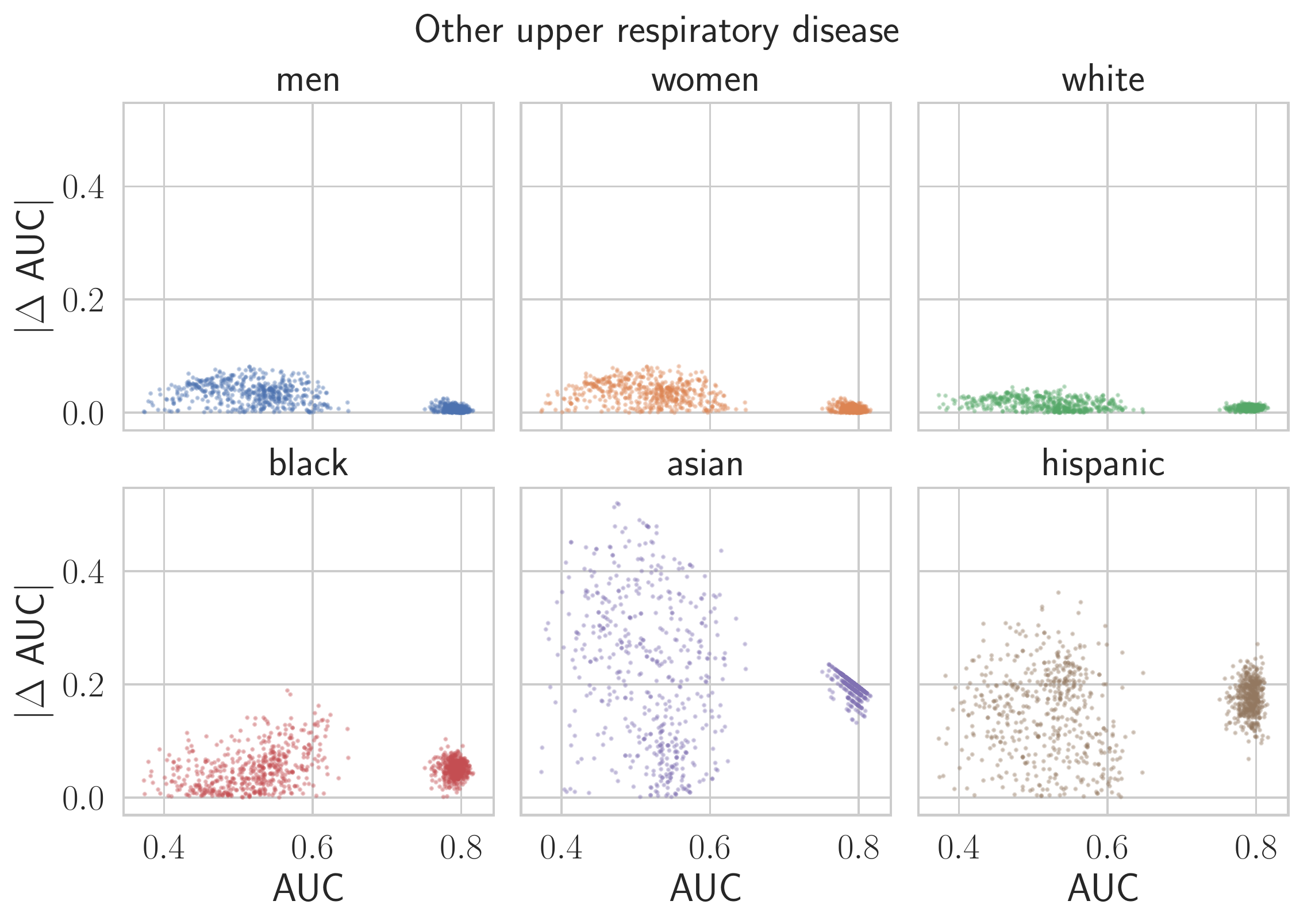}
  \end{minipage}
  \begin{minipage}[t]{0.48\textwidth}
    \includegraphics[width=\textwidth]{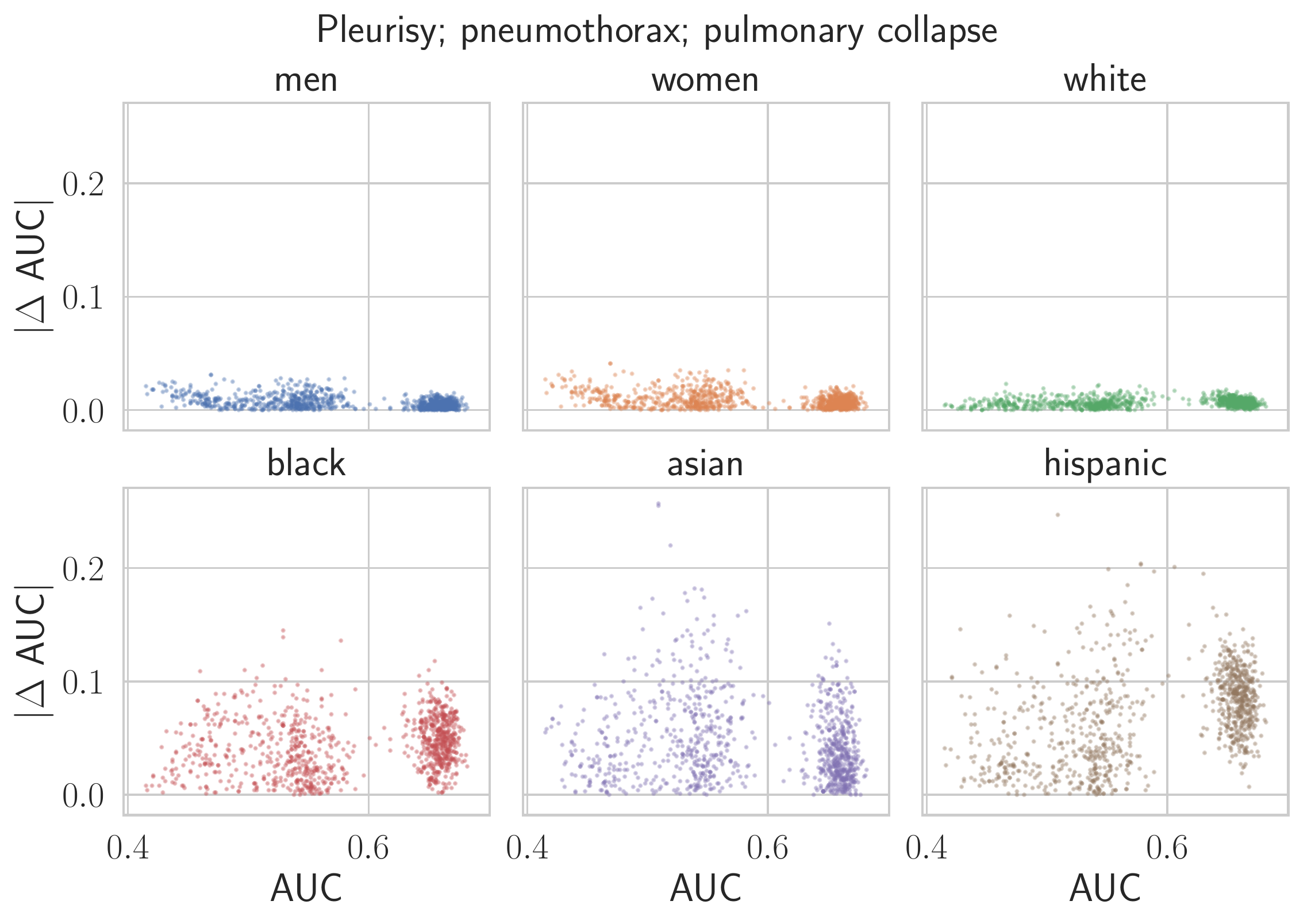}
  \end{minipage}
 \hfill
  \begin{minipage}[t]{0.48\textwidth}
    \includegraphics[width=\textwidth]{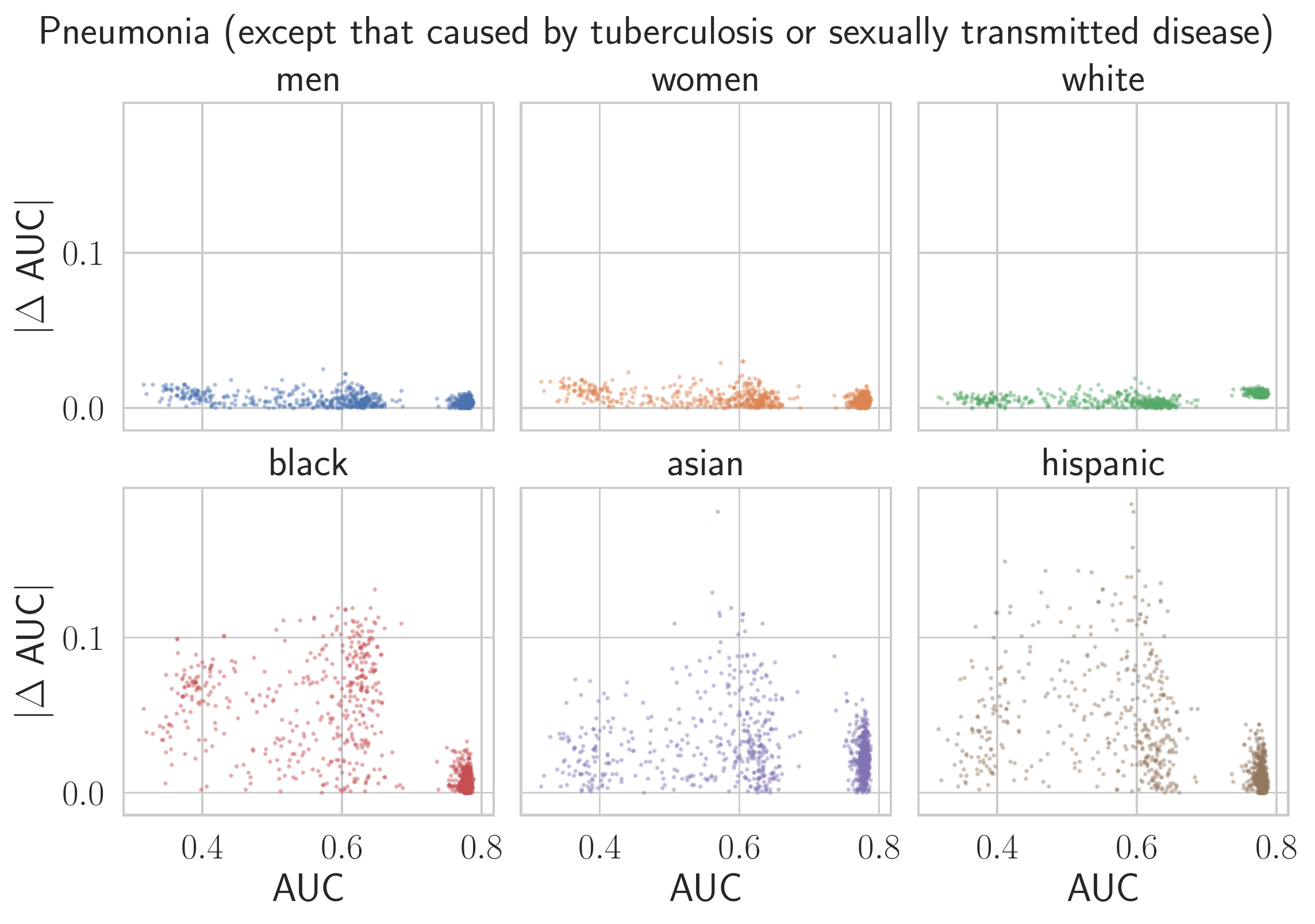}
  \end{minipage}
  \begin{minipage}[t]{0.48\textwidth}
    \includegraphics[width=\textwidth]{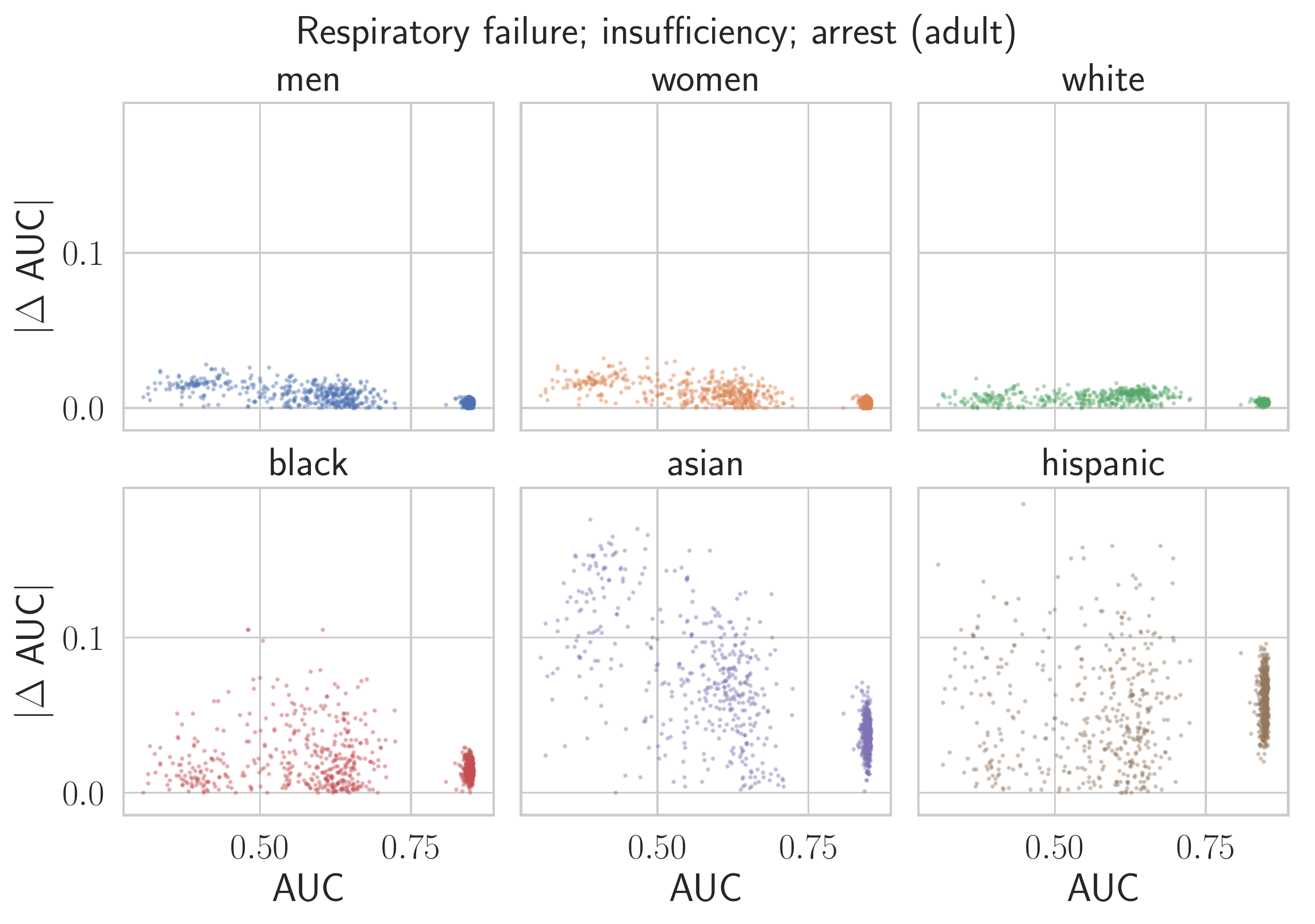}
  \end{minipage}
 \hfill
  \begin{minipage}[t]{0.48\textwidth}
    \includegraphics[width=\textwidth]{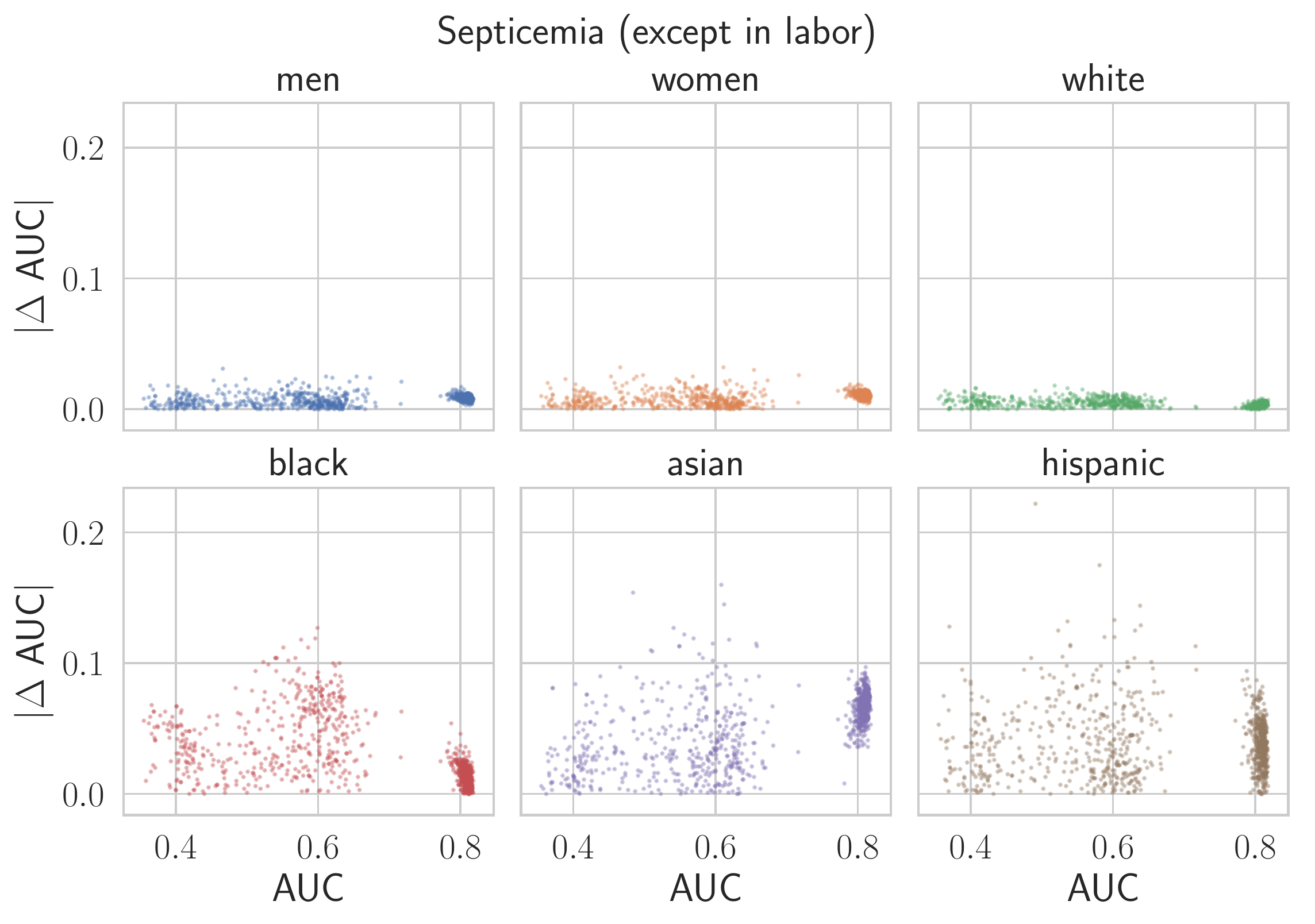}
  \end{minipage}
 \hfill
  \begin{minipage}[t]{0.48\textwidth}
    \includegraphics[width=\textwidth]{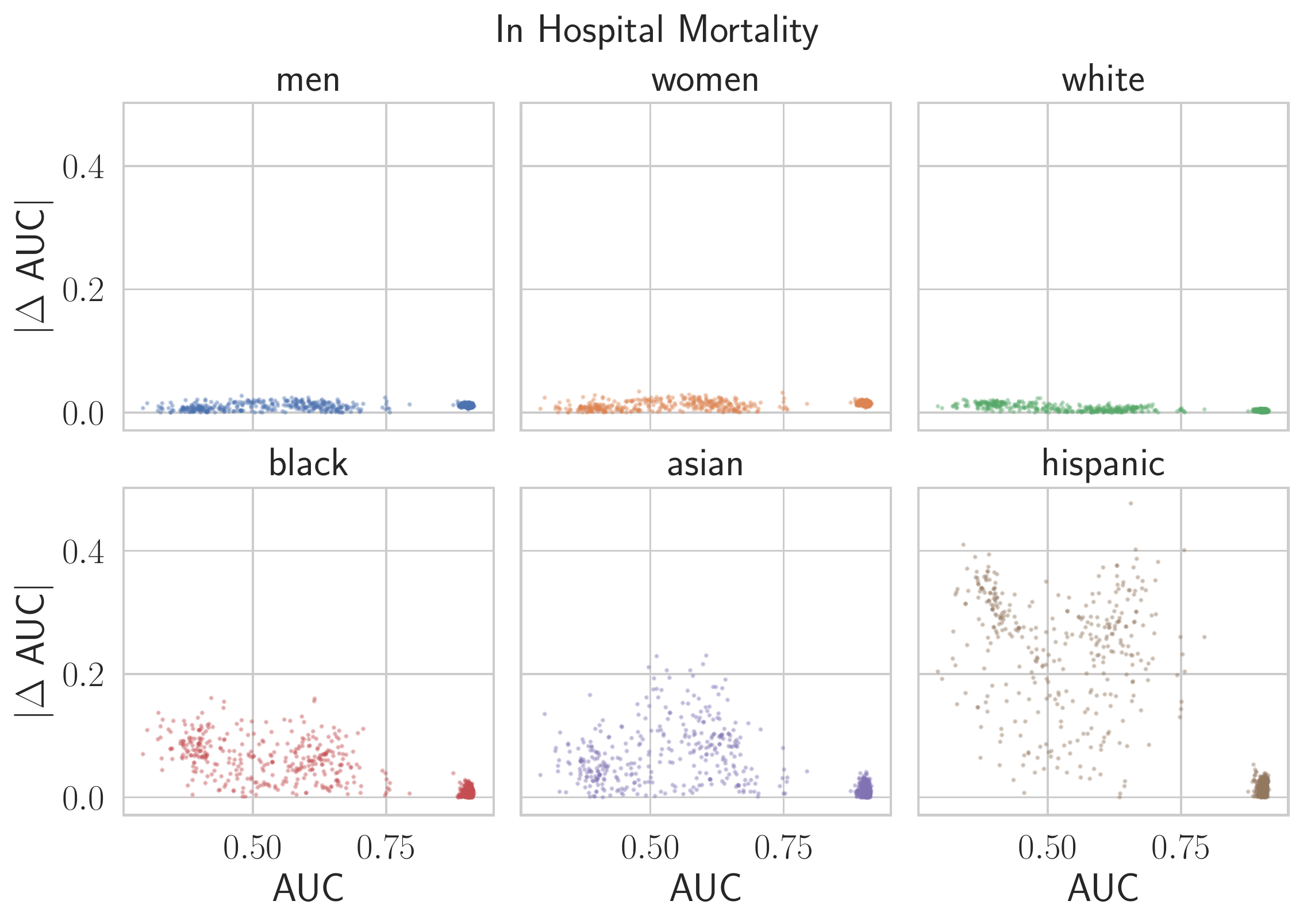}
  \end{minipage}
  \caption{Correlations between overall performance and subgroup performance $\Delta$}
  \label{fig:scatters_3}
\end{figure*}

\begin{figure*}
  \begin{minipage}[t]{0.32\textwidth}
    \includegraphics[width=\textwidth]{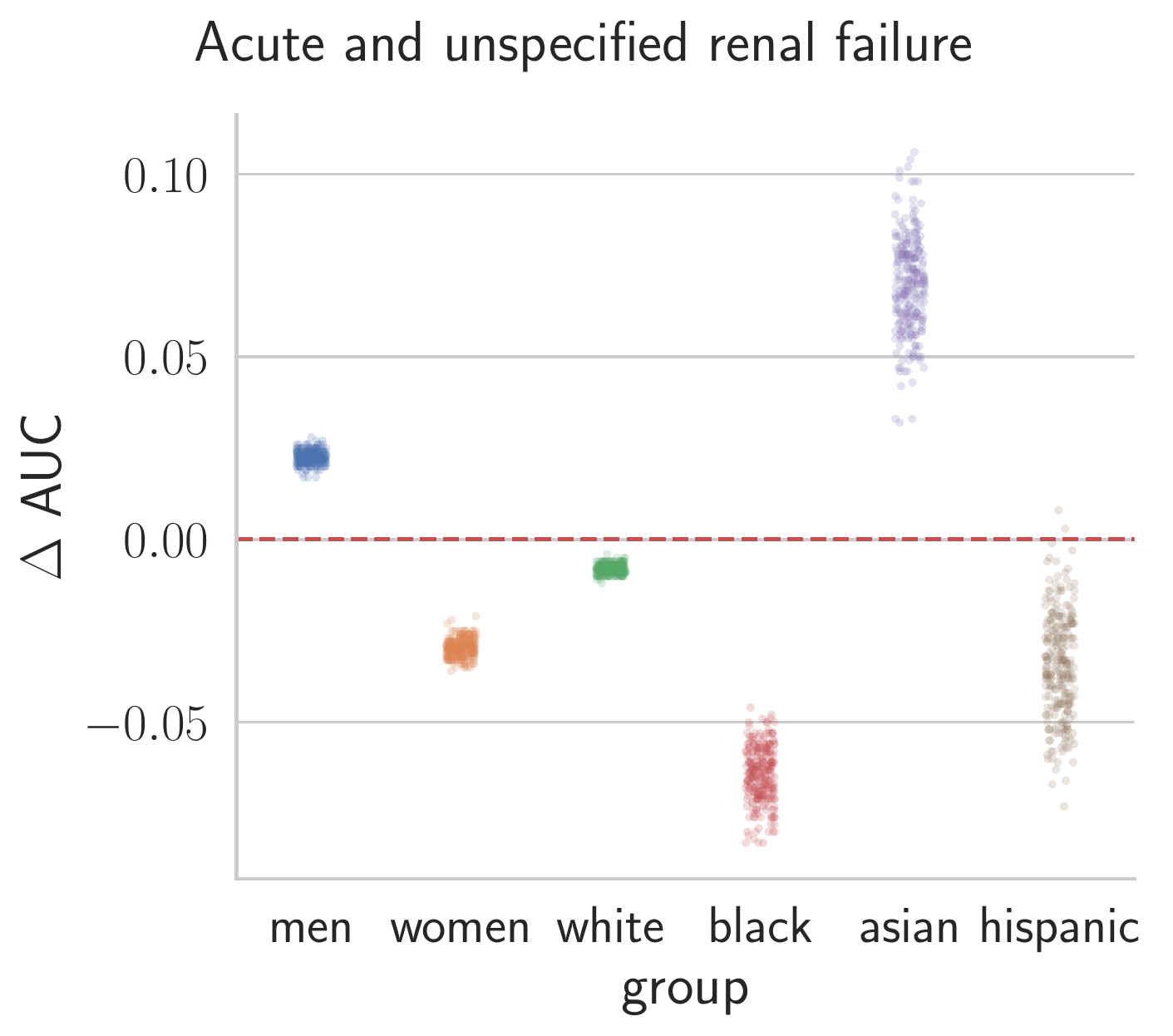}
  \end{minipage}
 \hfill
  \begin{minipage}[t]{0.32\textwidth}
    \includegraphics[width=\textwidth]{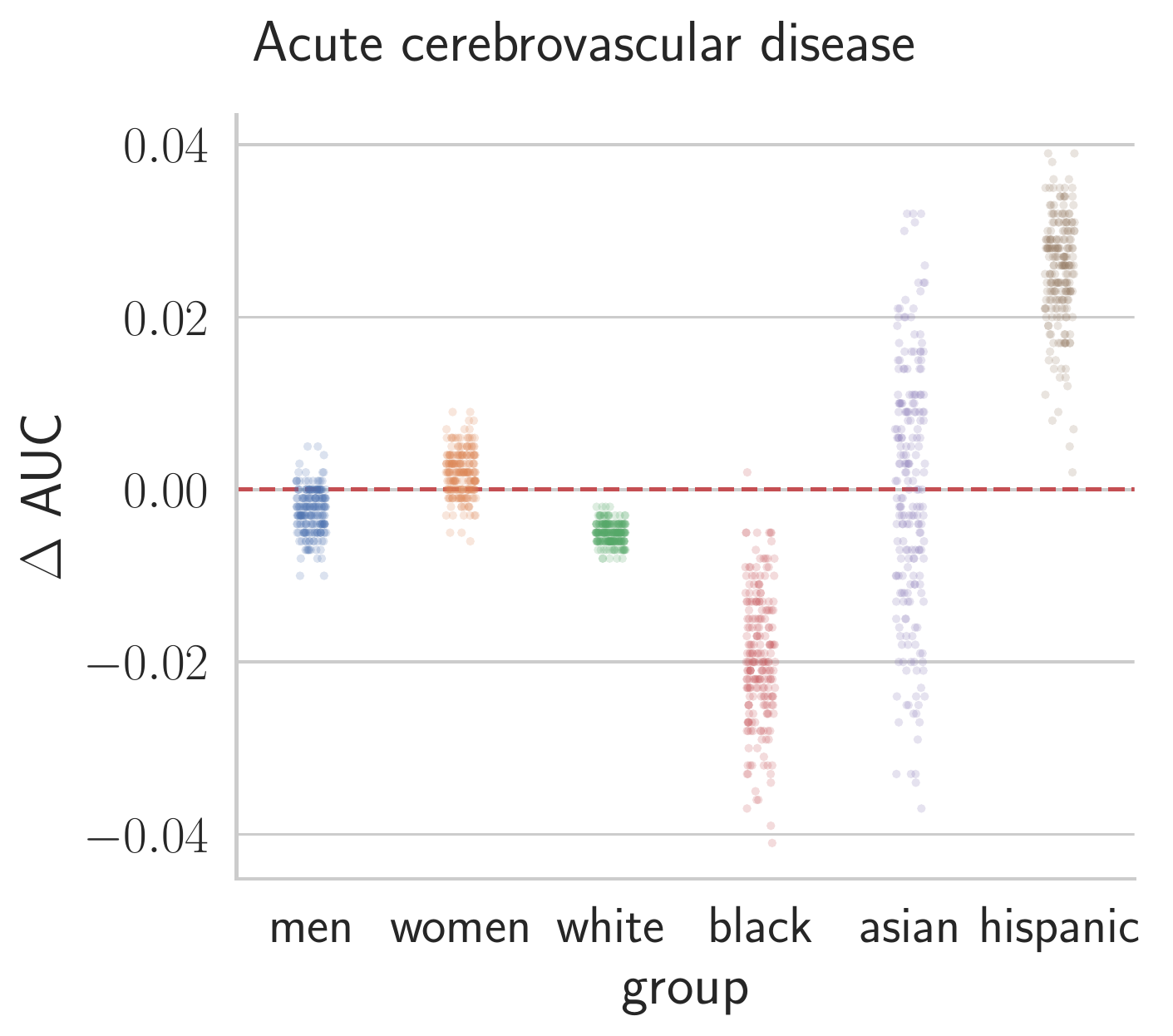}
  \end{minipage}
  \hfill
  \begin{minipage}[t]{0.32\textwidth}
    \includegraphics[width=\textwidth]{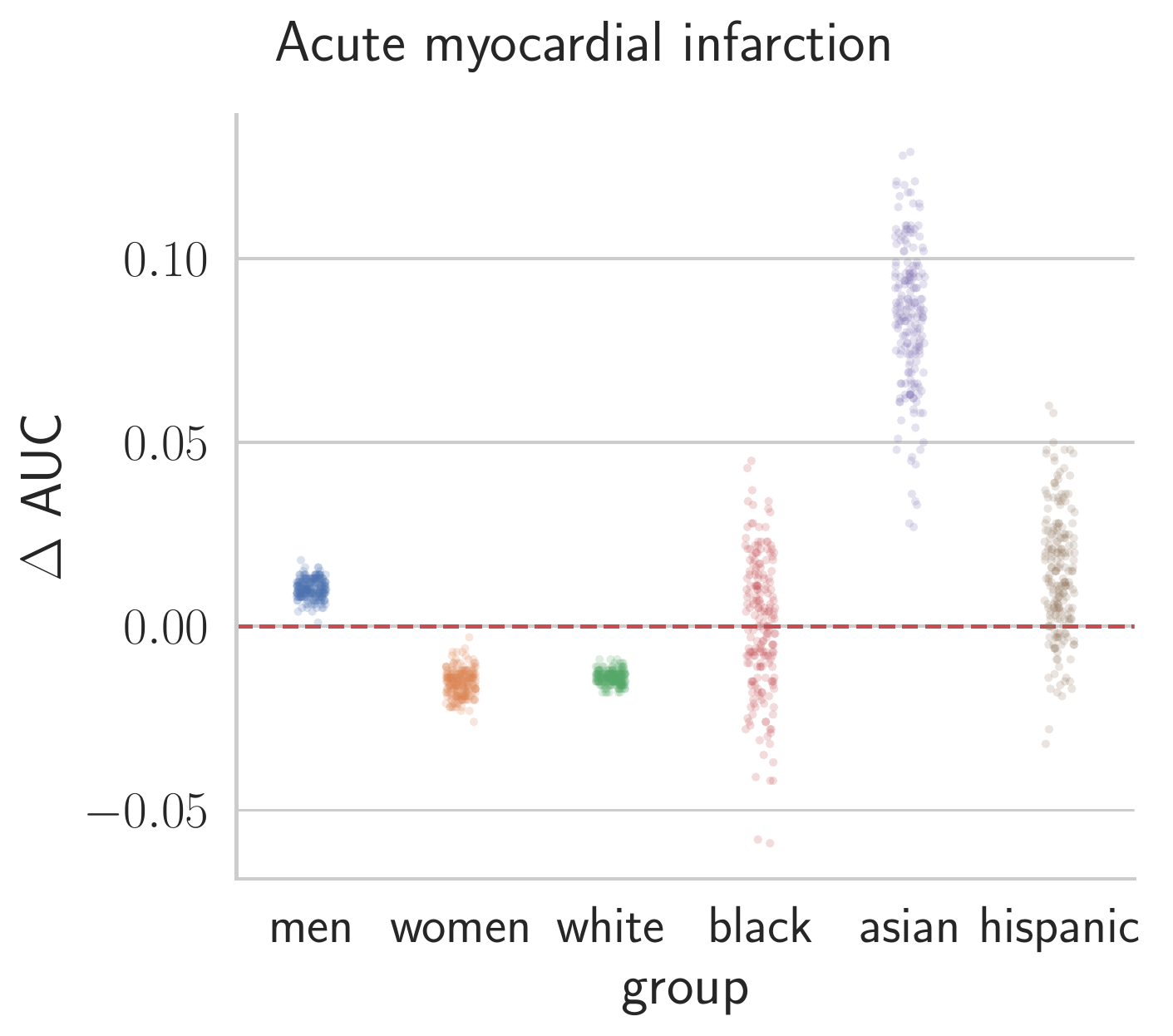}
  \end{minipage}
  \hfill
  \begin{minipage}[t]{0.32\textwidth}
    \includegraphics[width=\textwidth]{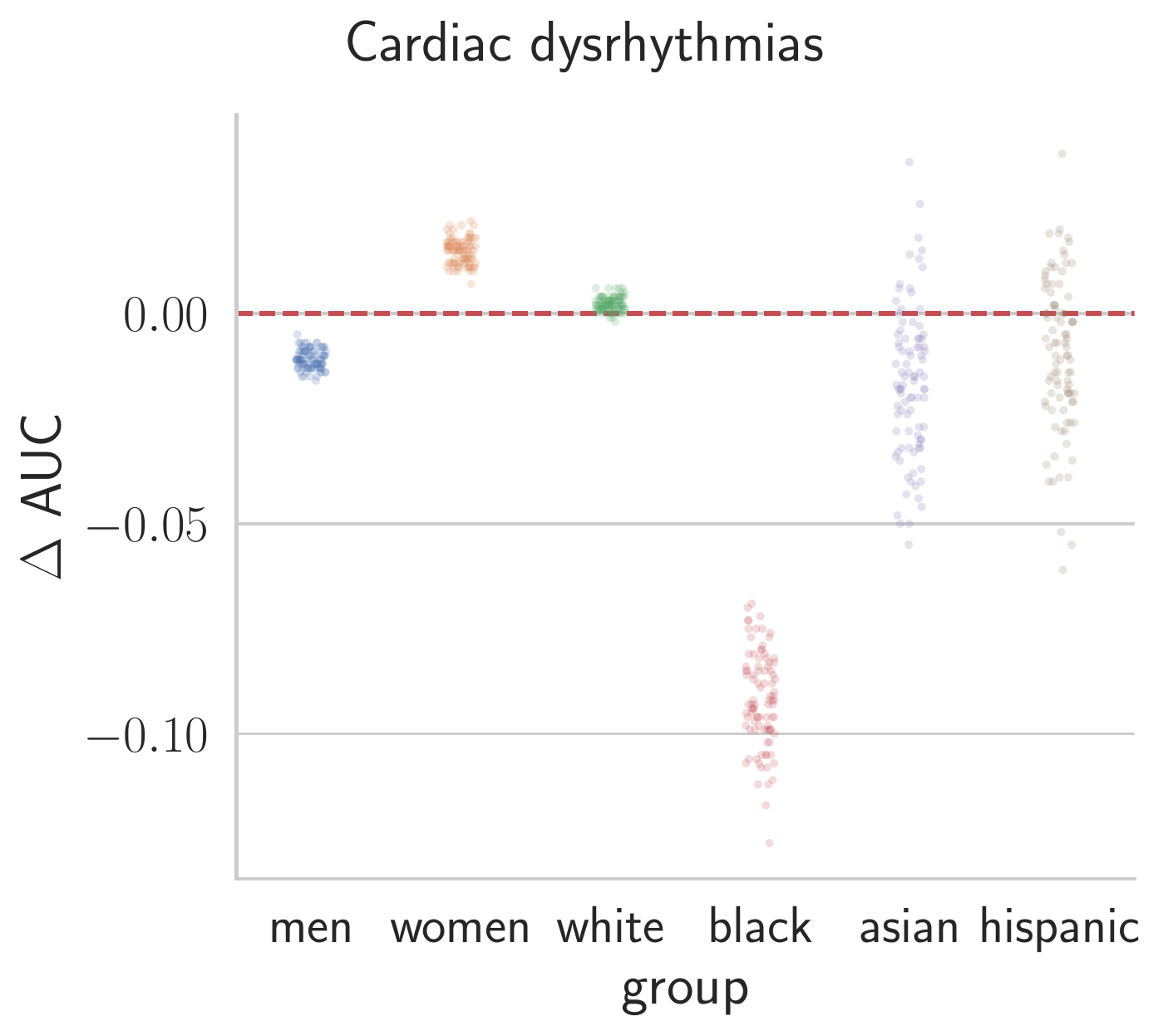}
  \end{minipage}
 \hfill
  \begin{minipage}[t]{0.32\textwidth}
    \includegraphics[width=\textwidth]{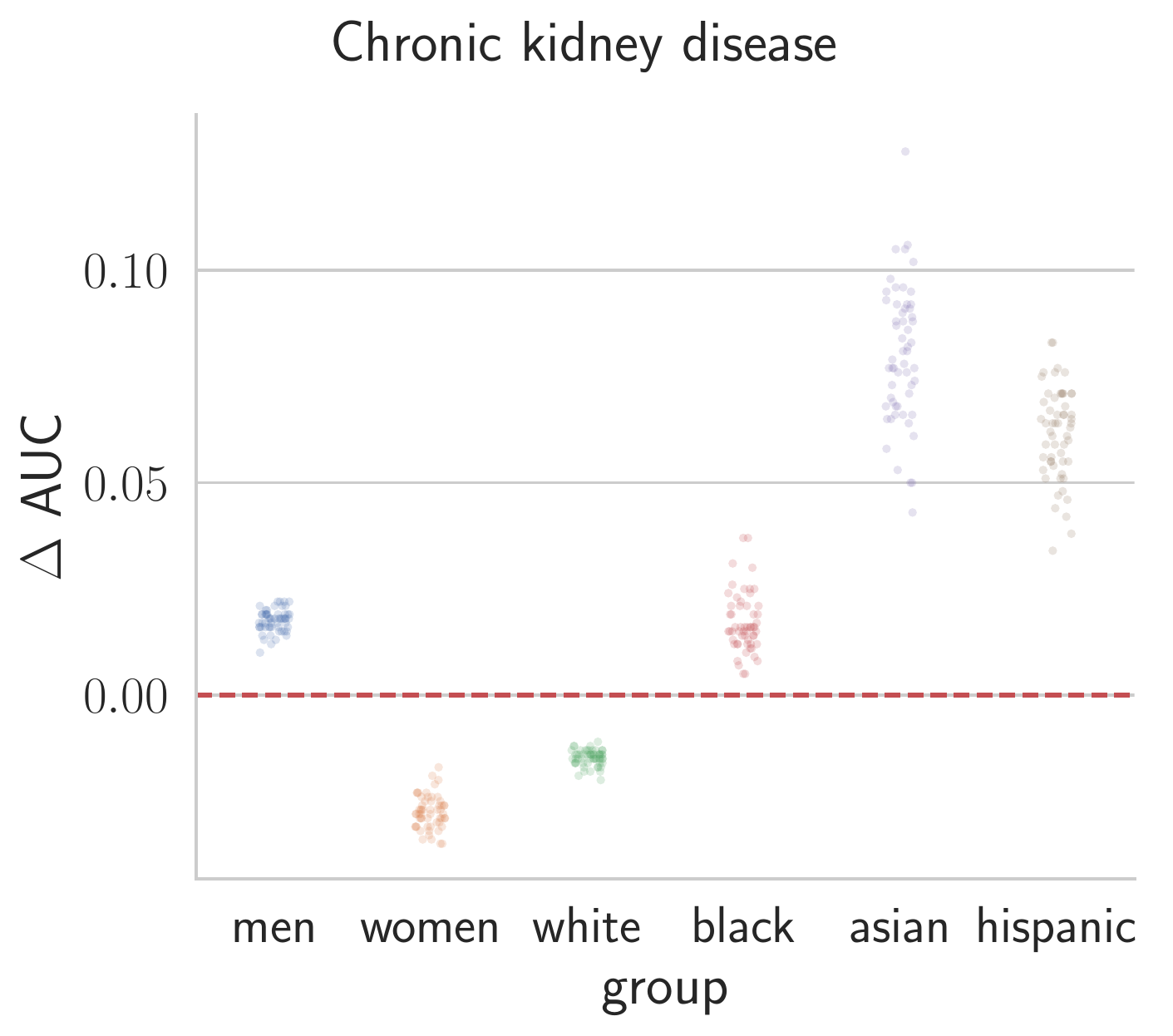}
  \end{minipage}
  \hfill
  \begin{minipage}[t]{0.32\textwidth}
    \includegraphics[width=\textwidth]{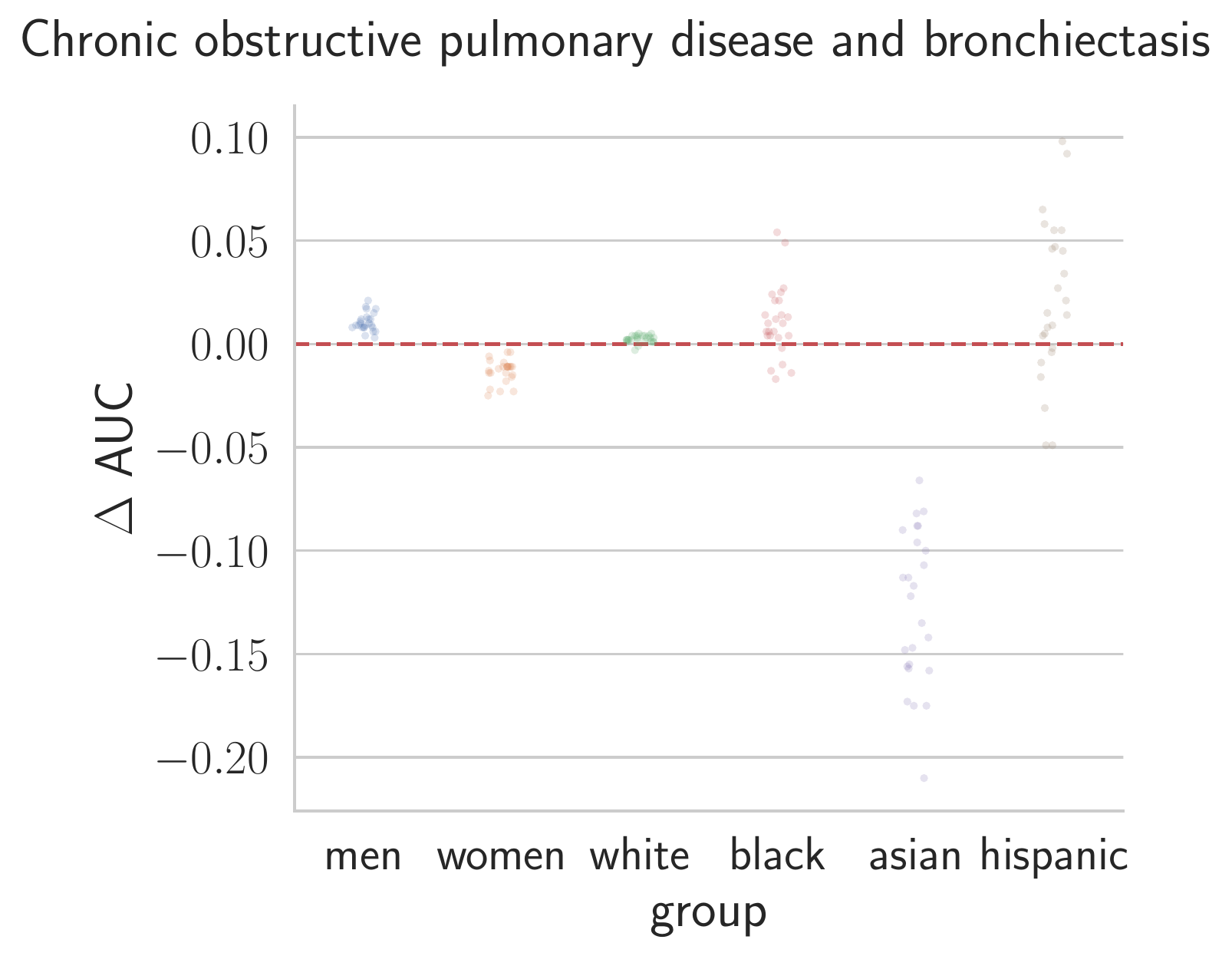}
  \end{minipage}
  \hfill
  \begin{minipage}[t]{0.32\textwidth}
    \includegraphics[width=\textwidth]{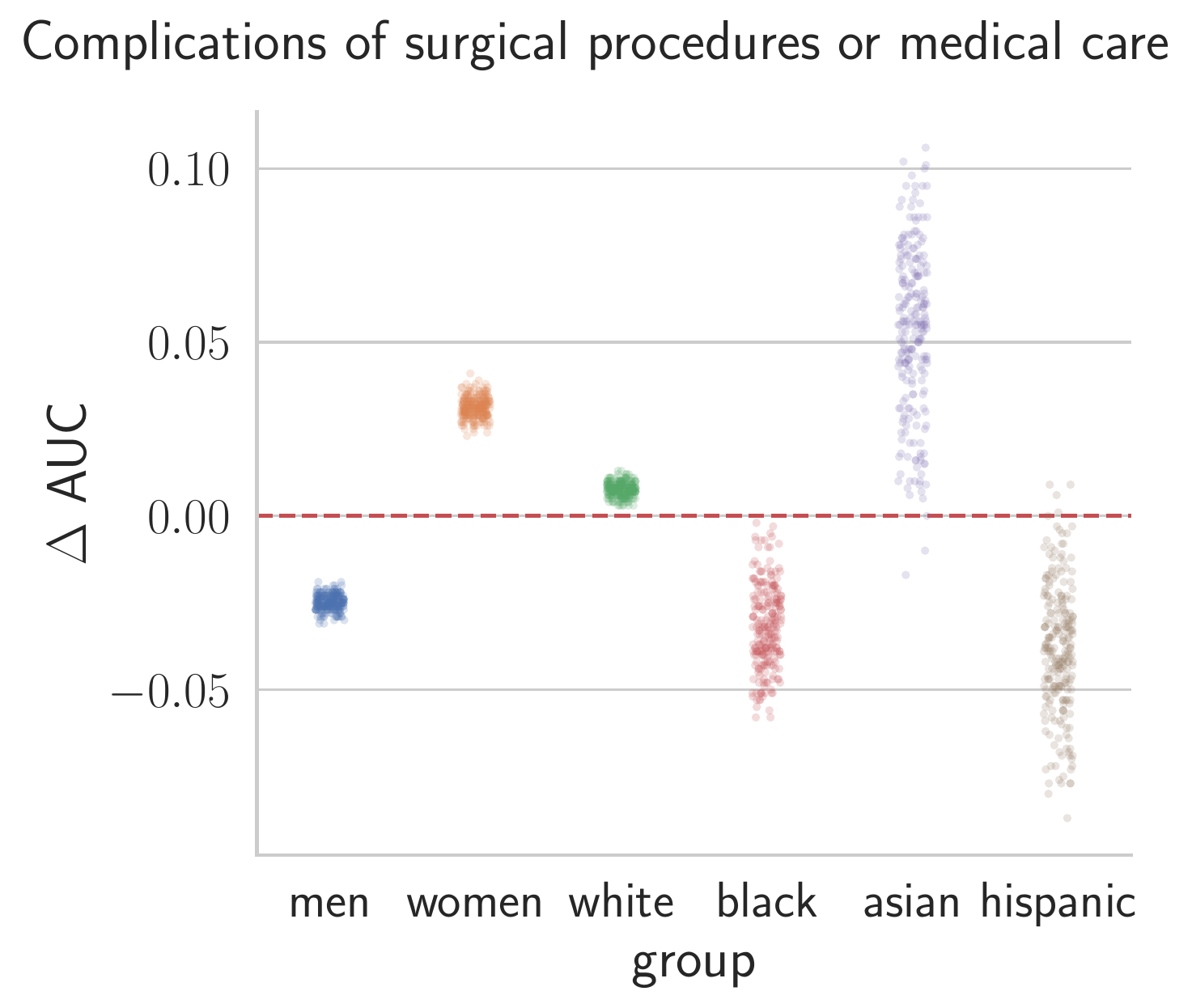}
  \end{minipage}
 \hfill
  \begin{minipage}[t]{0.32\textwidth}
    \includegraphics[width=\textwidth]{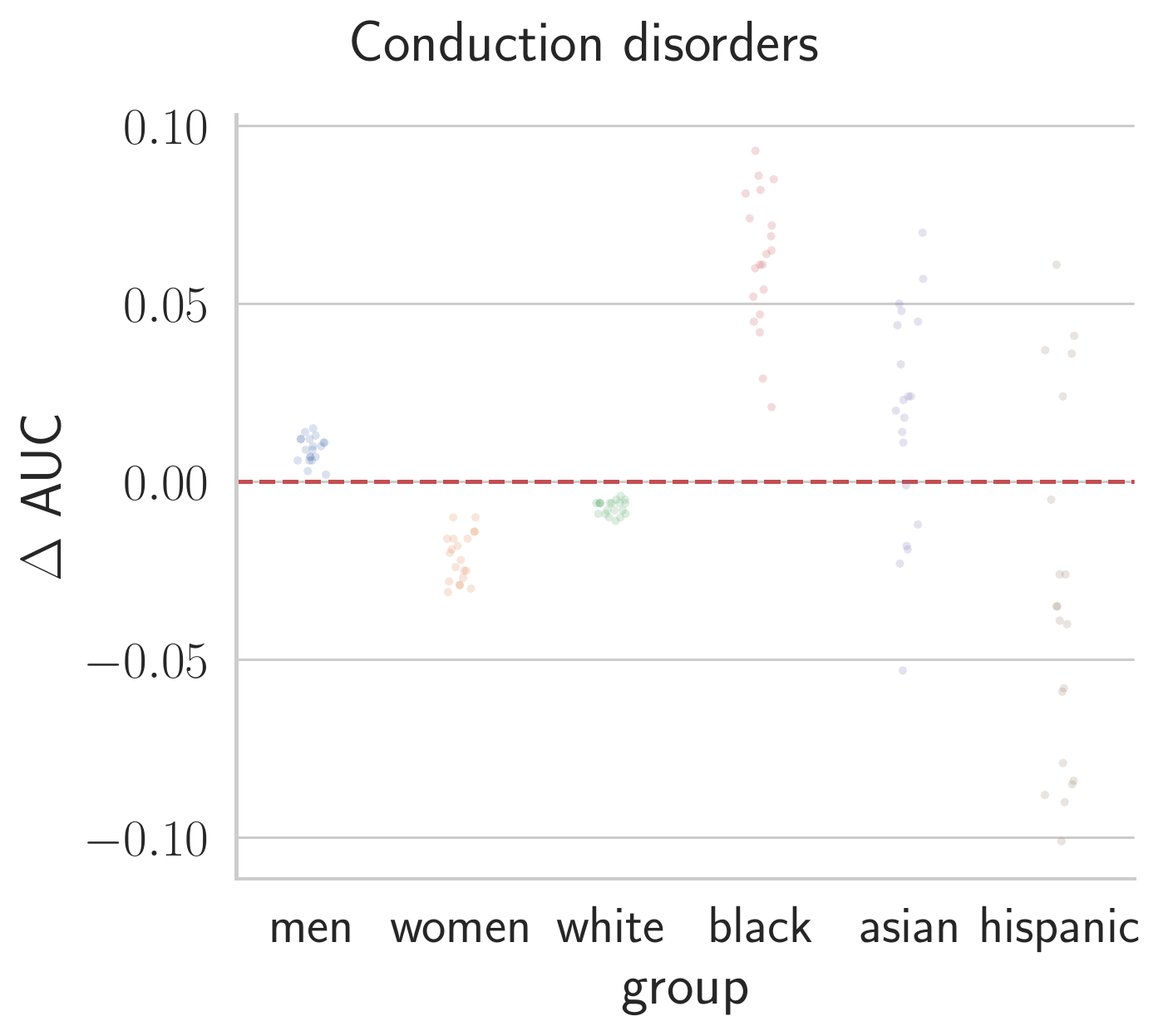}
  \end{minipage}
  \hfill
  \begin{minipage}[t]{0.32\textwidth}
    \includegraphics[width=\textwidth]{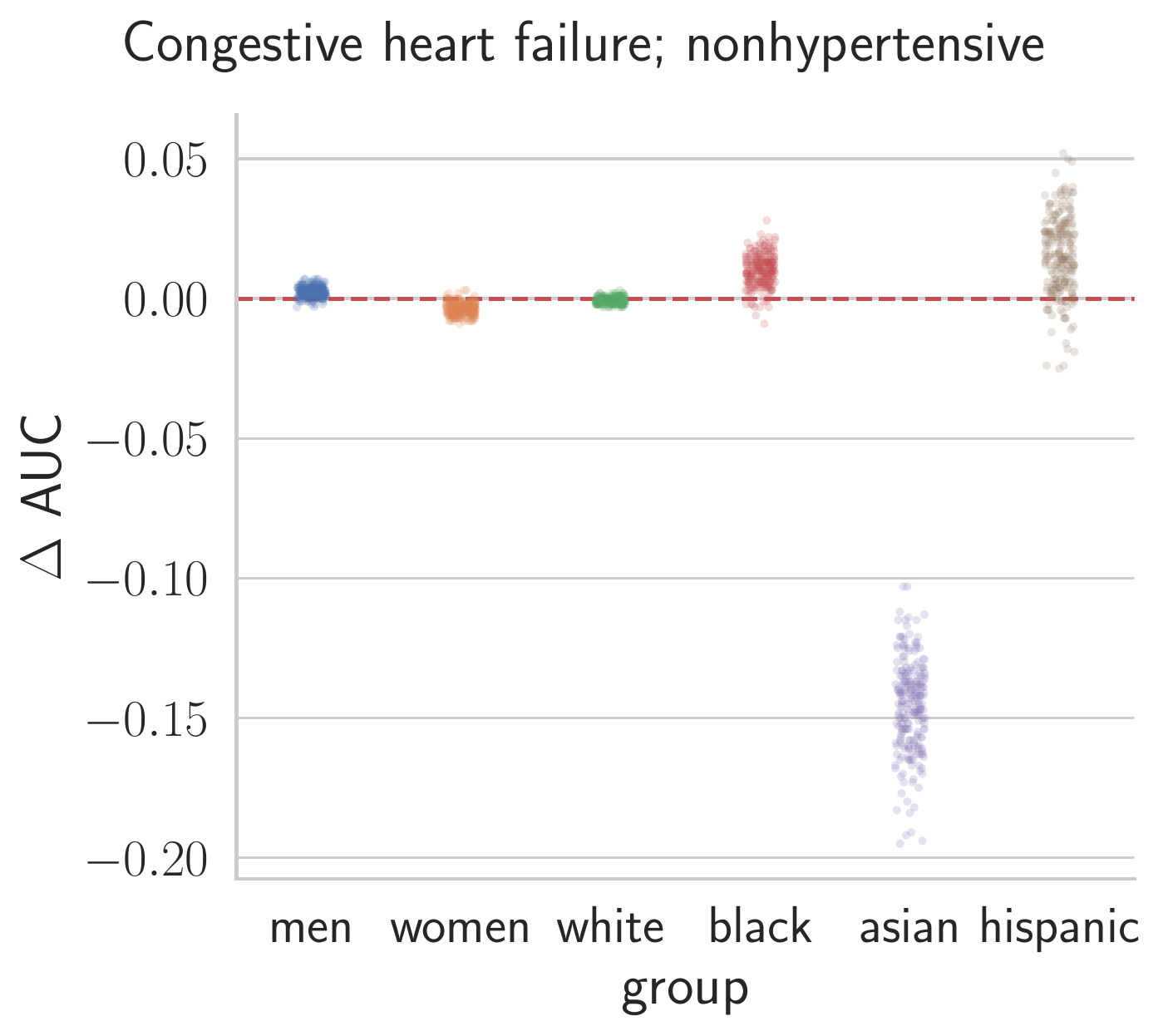}
  \end{minipage}
  \hfill
  \begin{minipage}[t]{0.32\textwidth}
    \includegraphics[width=\textwidth]{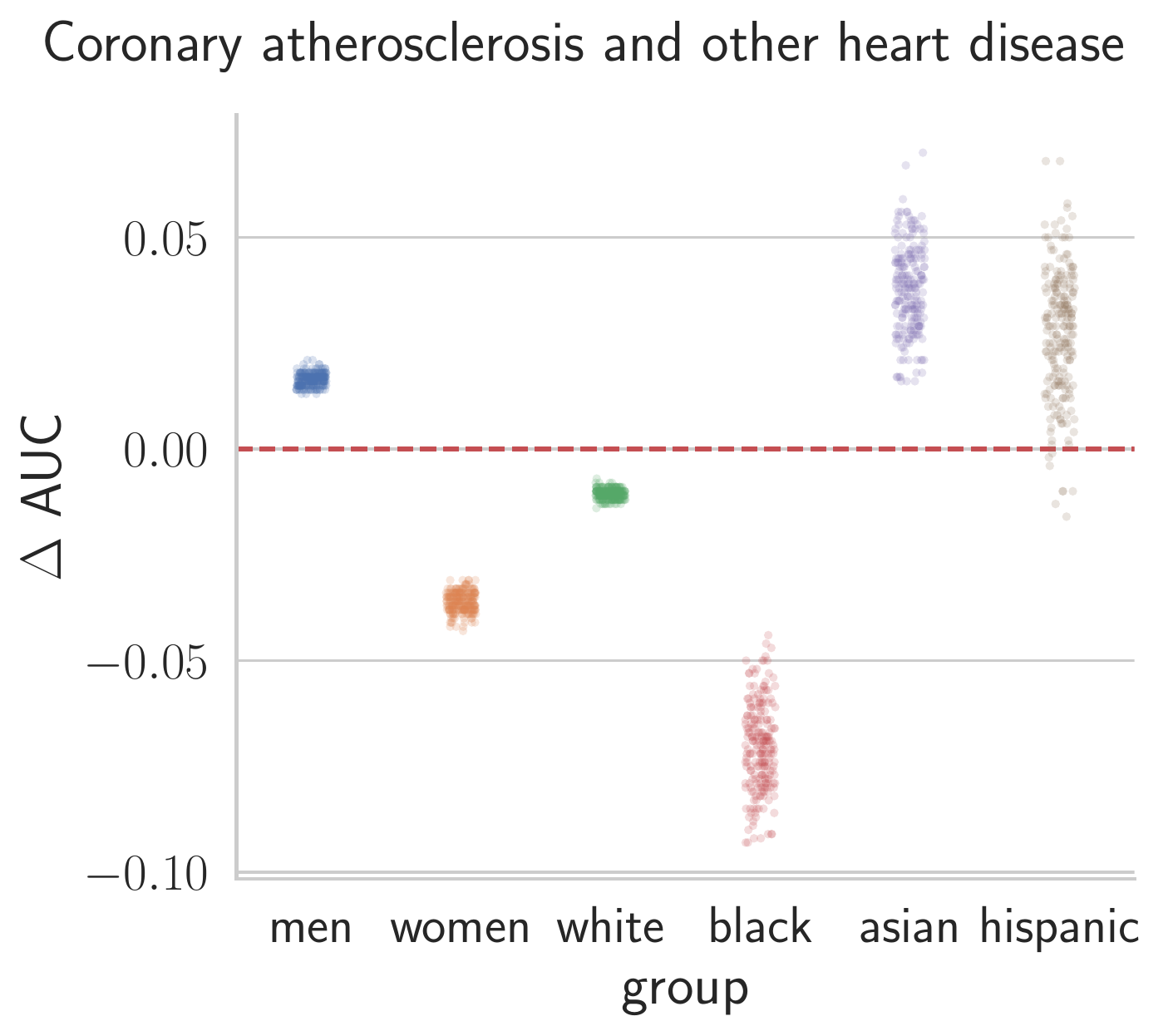}
  \end{minipage}
 \hfill
  \begin{minipage}[t]{0.32\textwidth}
    \includegraphics[width=\textwidth]{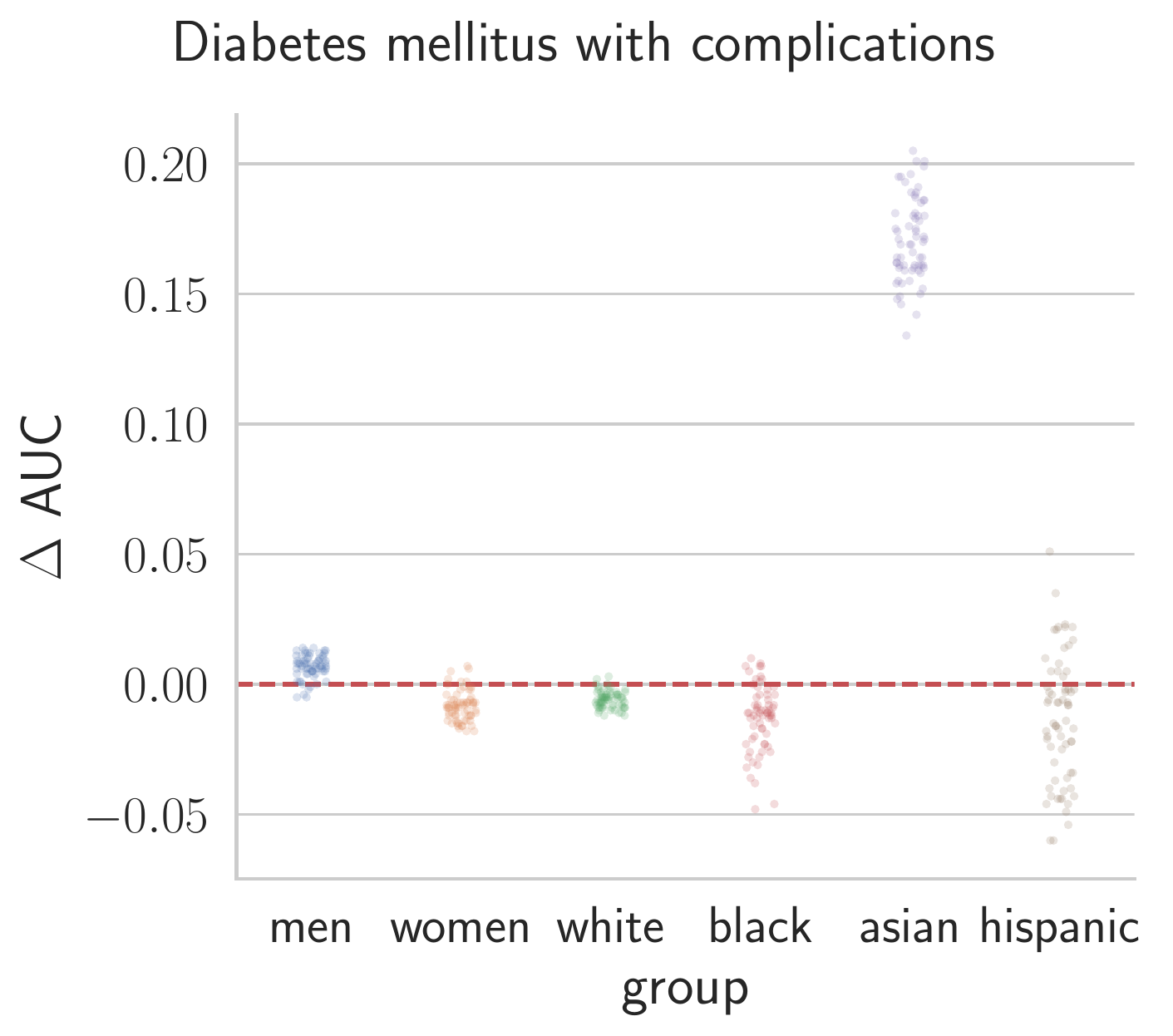}
  \end{minipage}
  \hfill
  \begin{minipage}[t]{0.32\textwidth}
    \includegraphics[width=\textwidth]{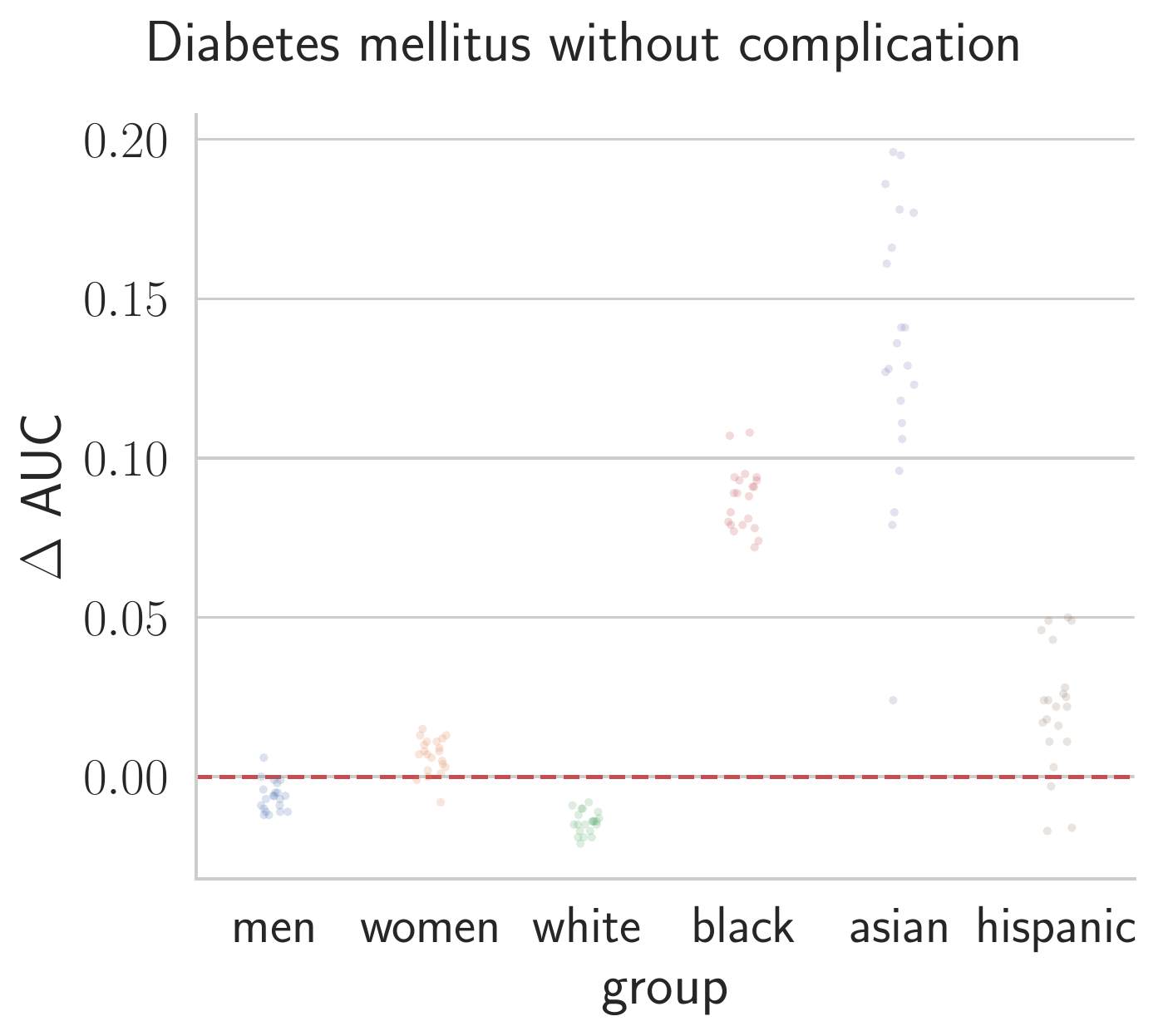}
  \end{minipage}
  \hfill
  \begin{minipage}[t]{0.32\textwidth}
    \includegraphics[width=\textwidth]{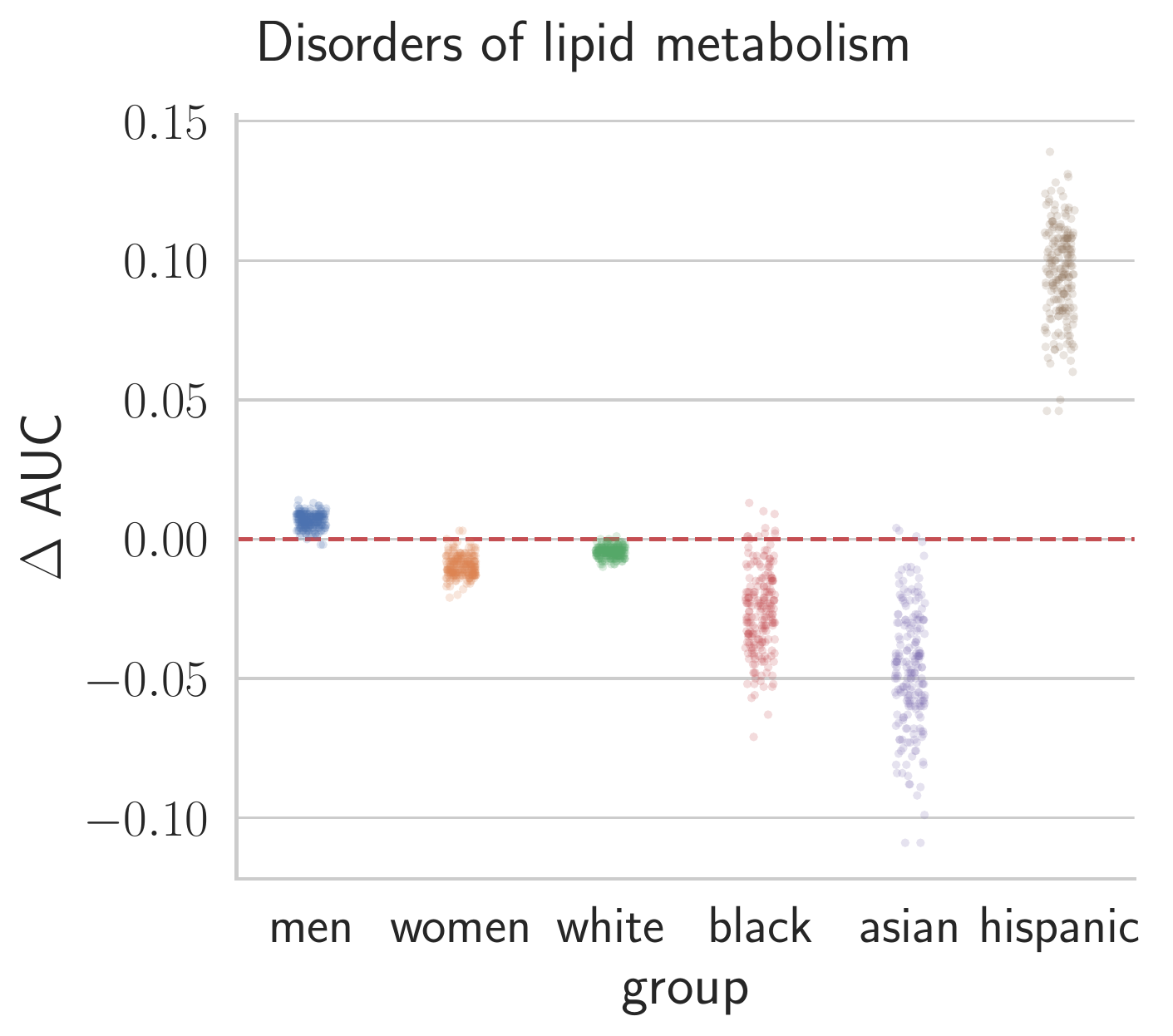}
  \end{minipage}
 \hfill
  \begin{minipage}[t]{0.32\textwidth}
    \includegraphics[width=\textwidth]{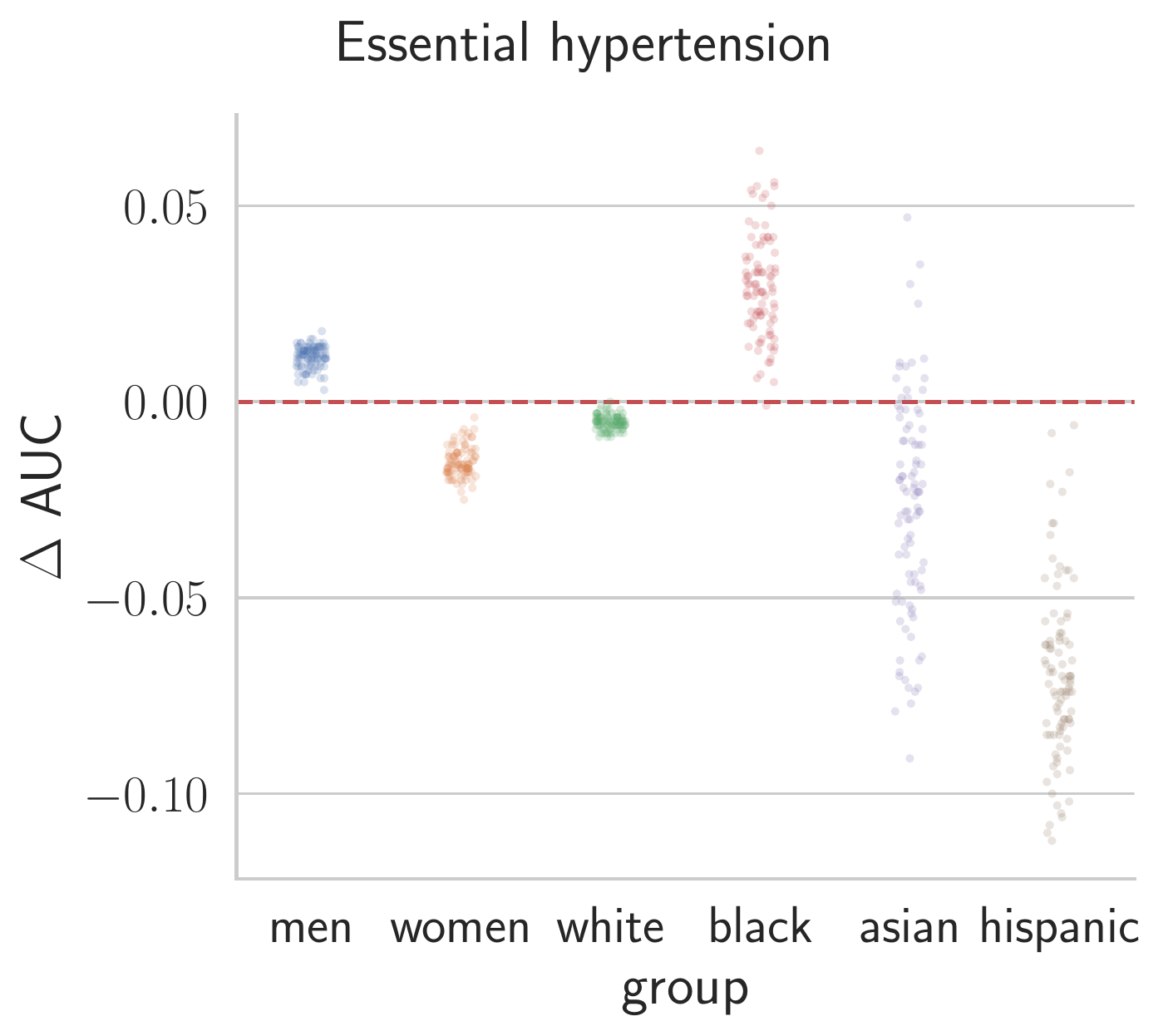}
  \end{minipage}
  \hfill
  \begin{minipage}[t]{0.32\textwidth}
    \includegraphics[width=\textwidth]{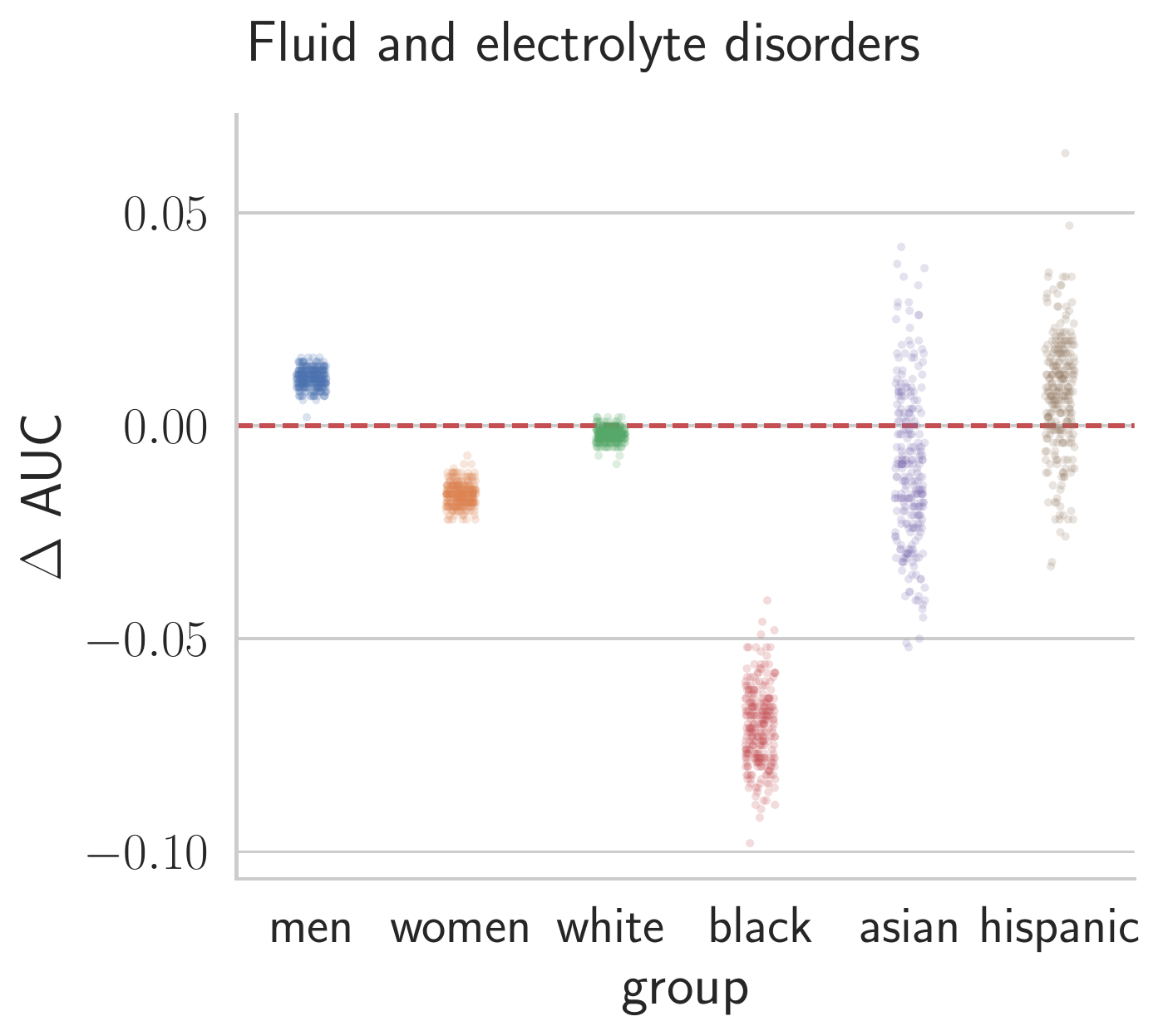}
  \end{minipage}
  \hfill
  \caption{Differences relative to overall performance as a function of random seeds for each subgroup. Each point represents a run for a pair of seeds with validation performance similar to that of the best seeds. }
  \label{fig:underspec_1}
\end{figure*}

\begin{figure*}
  \begin{minipage}[t]{0.32\textwidth}
    \includegraphics[width=\textwidth]{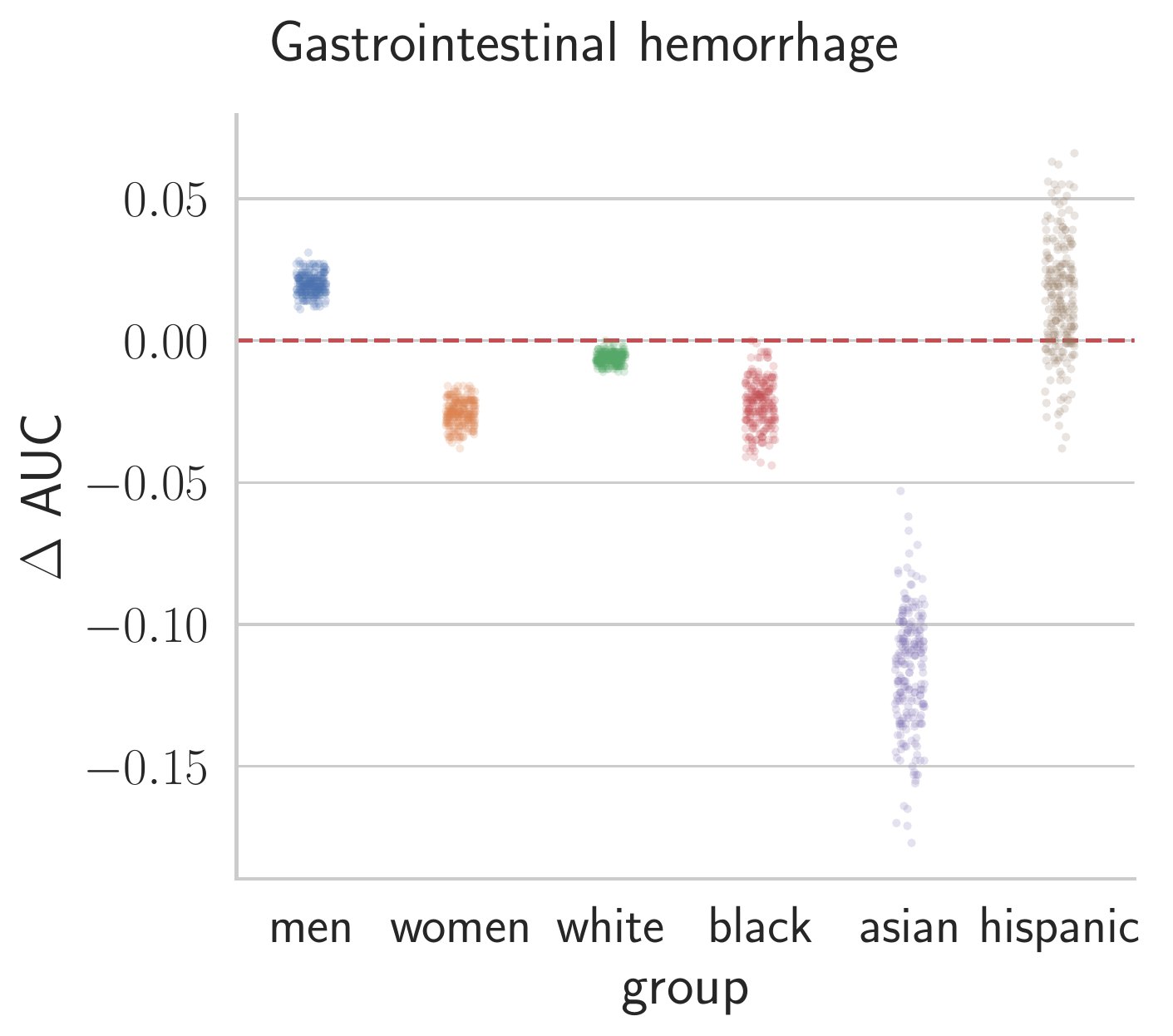}
  \end{minipage}
 \hfill
  \begin{minipage}[t]{0.32\textwidth}
    \includegraphics[width=\textwidth]{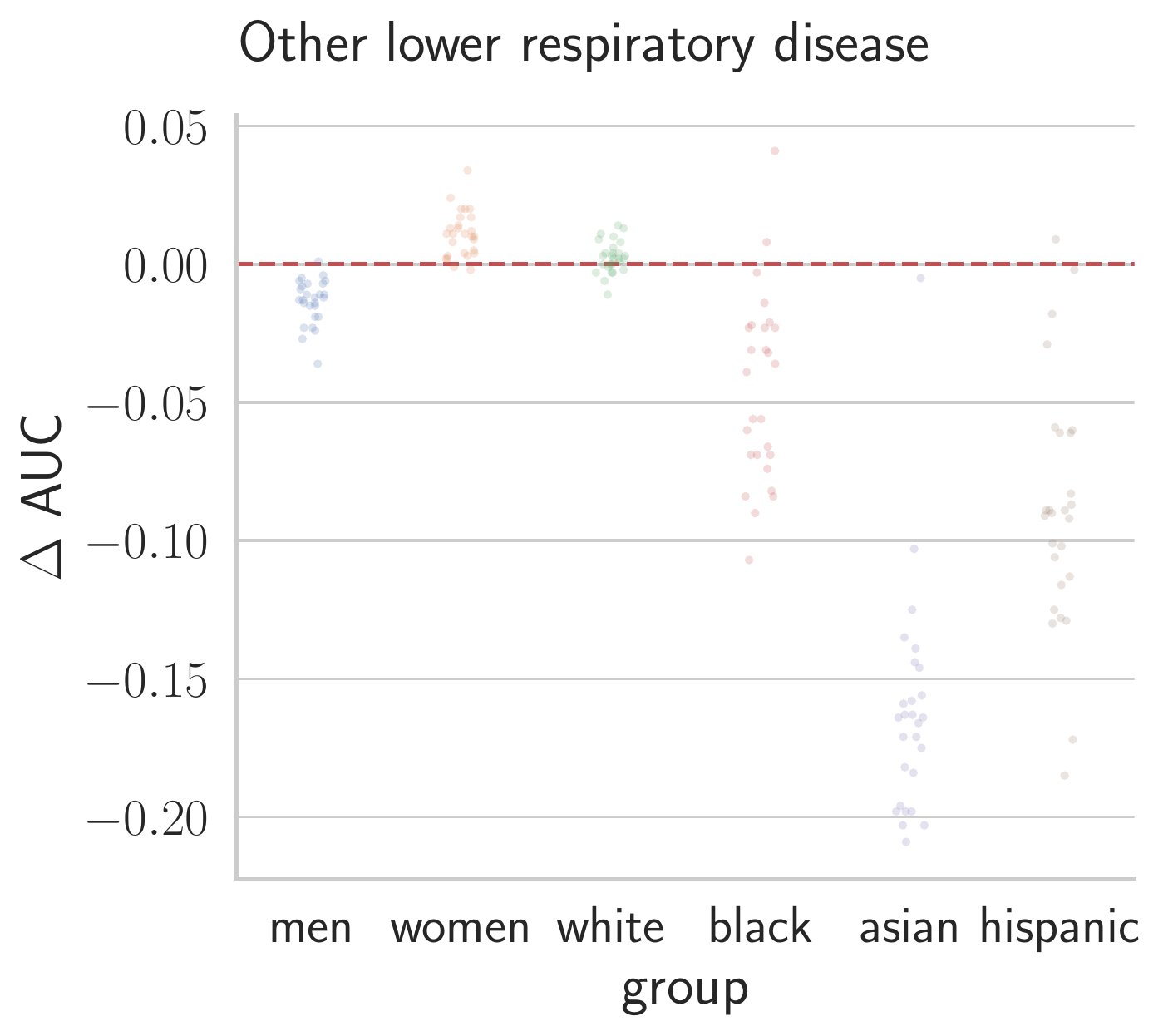}
  \end{minipage}
 \hfill
  \begin{minipage}[t]{0.32\textwidth}
    \includegraphics[width=\textwidth]{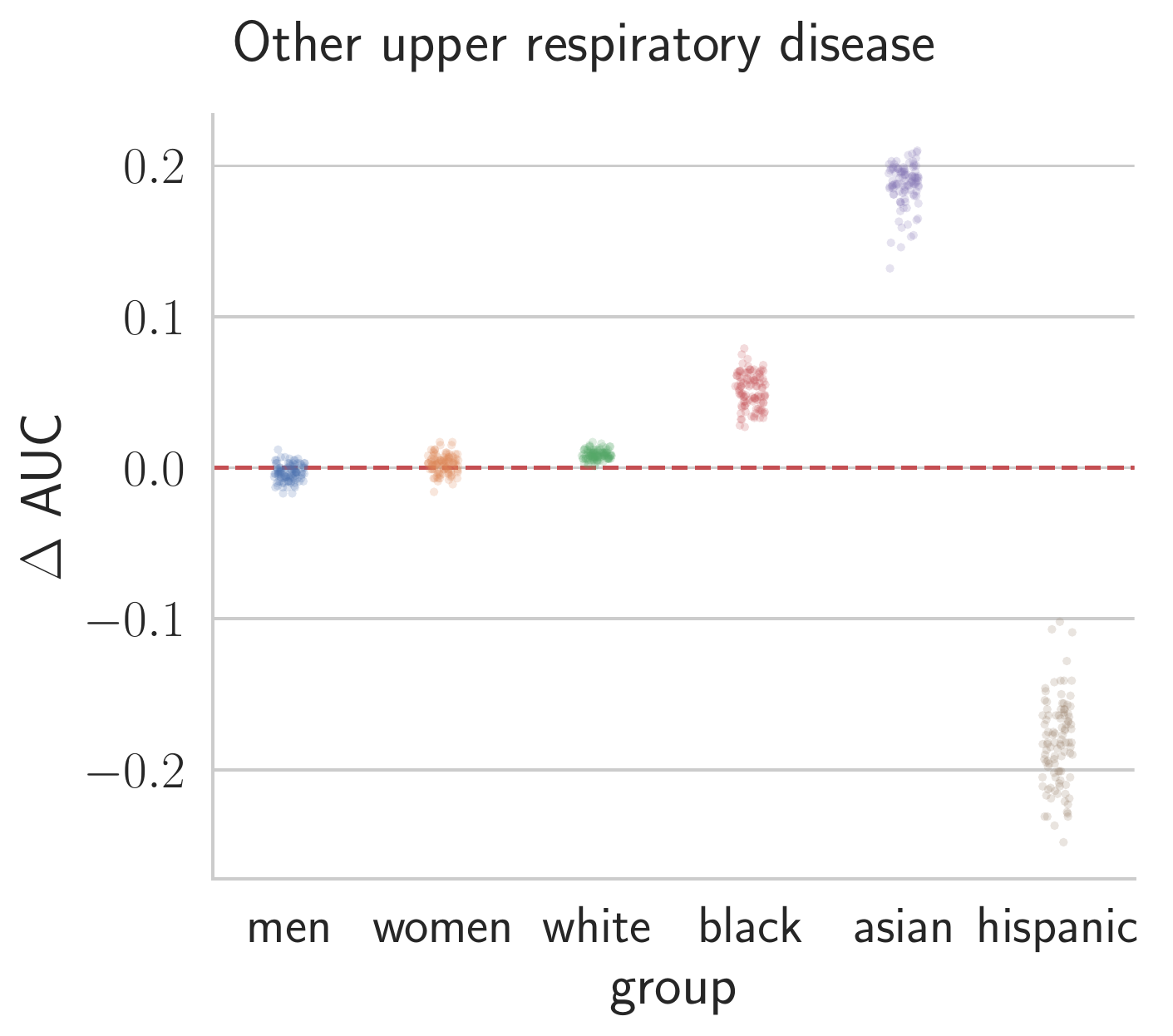}
  \end{minipage}
  \hfill
  \begin{minipage}[t]{0.32\textwidth}
    \includegraphics[width=\textwidth]{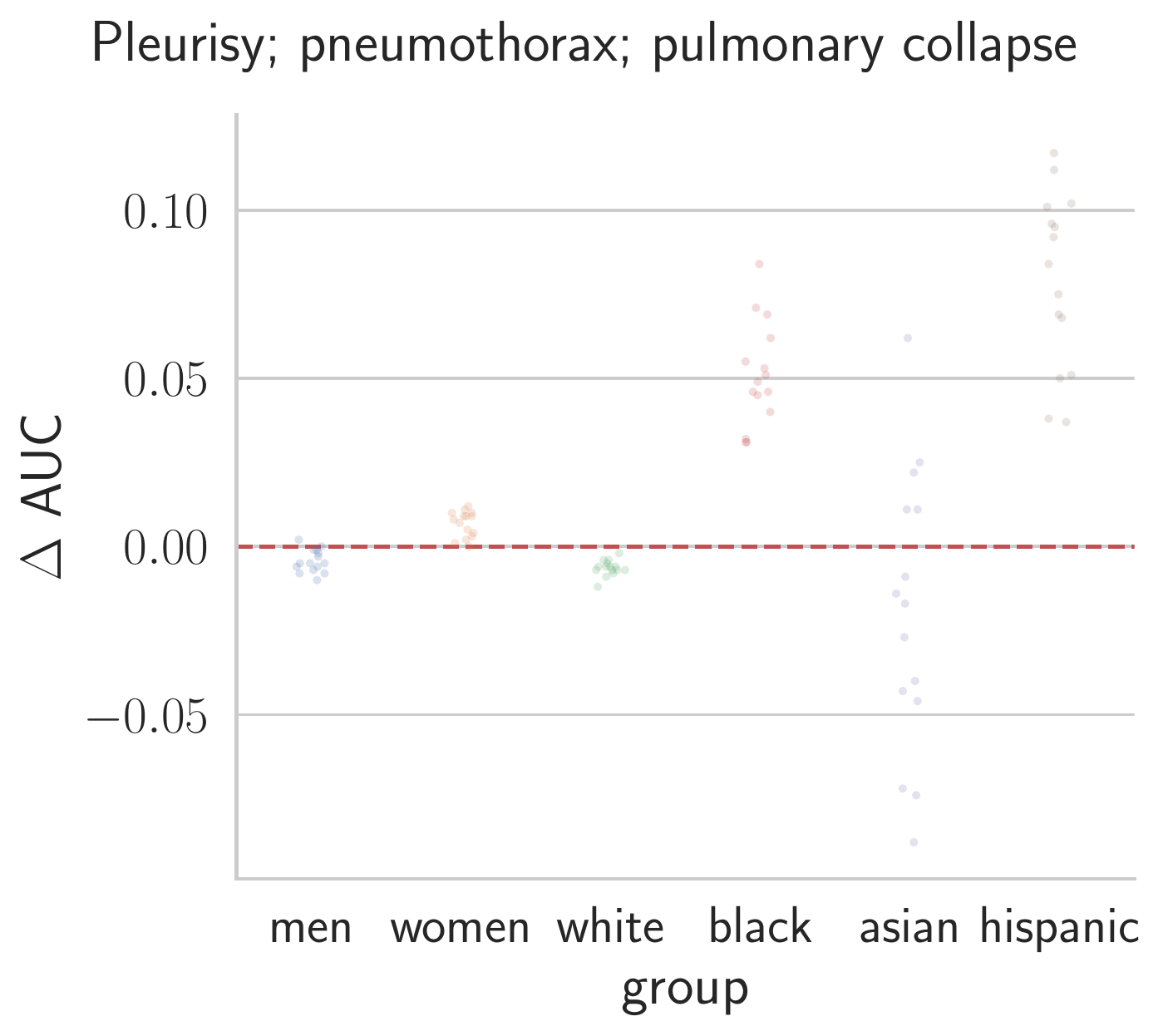}
  \end{minipage}
  \hfill
  \begin{minipage}[t]{0.32\textwidth}
    \includegraphics[width=\textwidth]{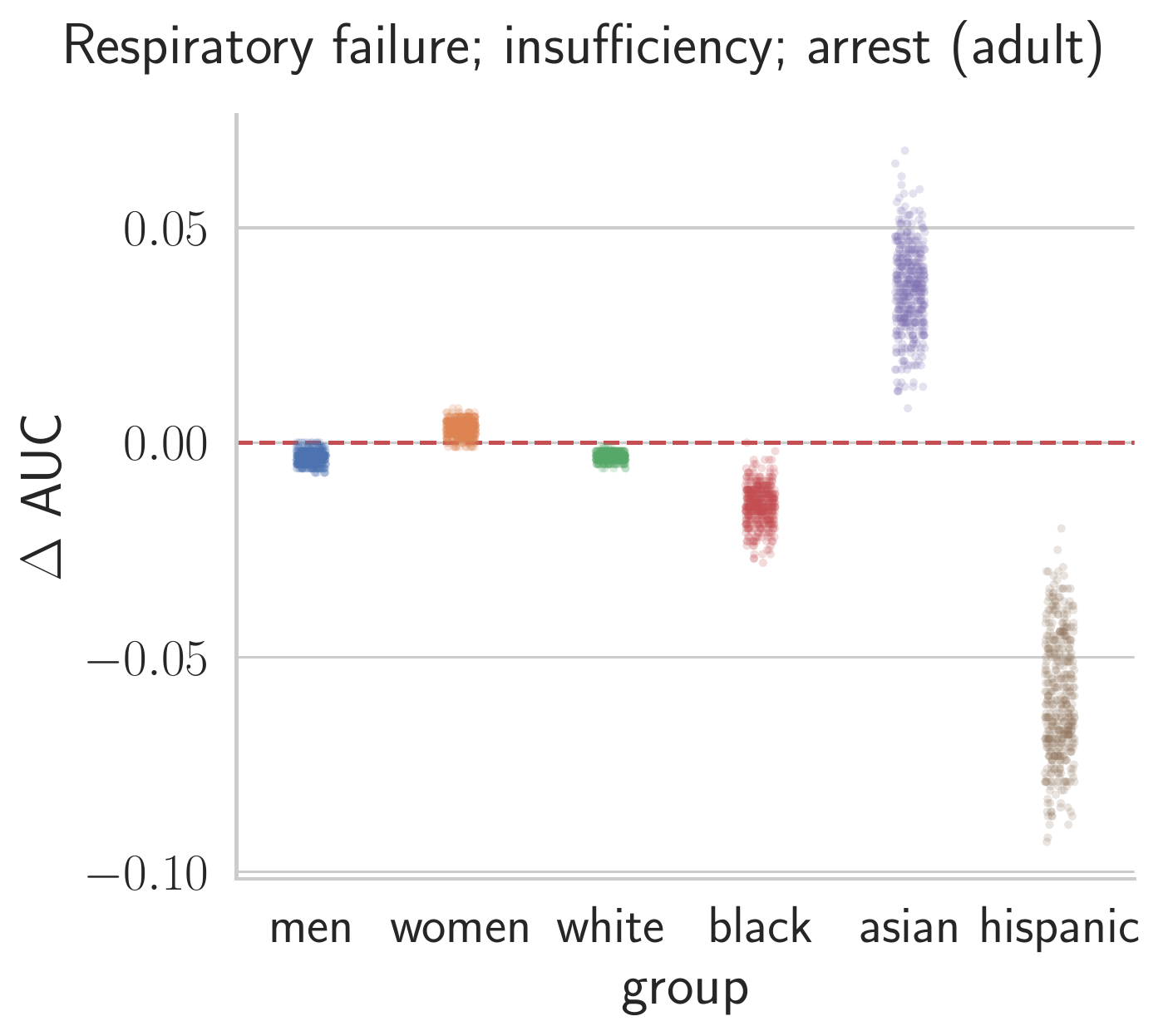}
  \end{minipage}
  \hfill
  \begin{minipage}[t]{0.32\textwidth}
    \includegraphics[width=\textwidth]{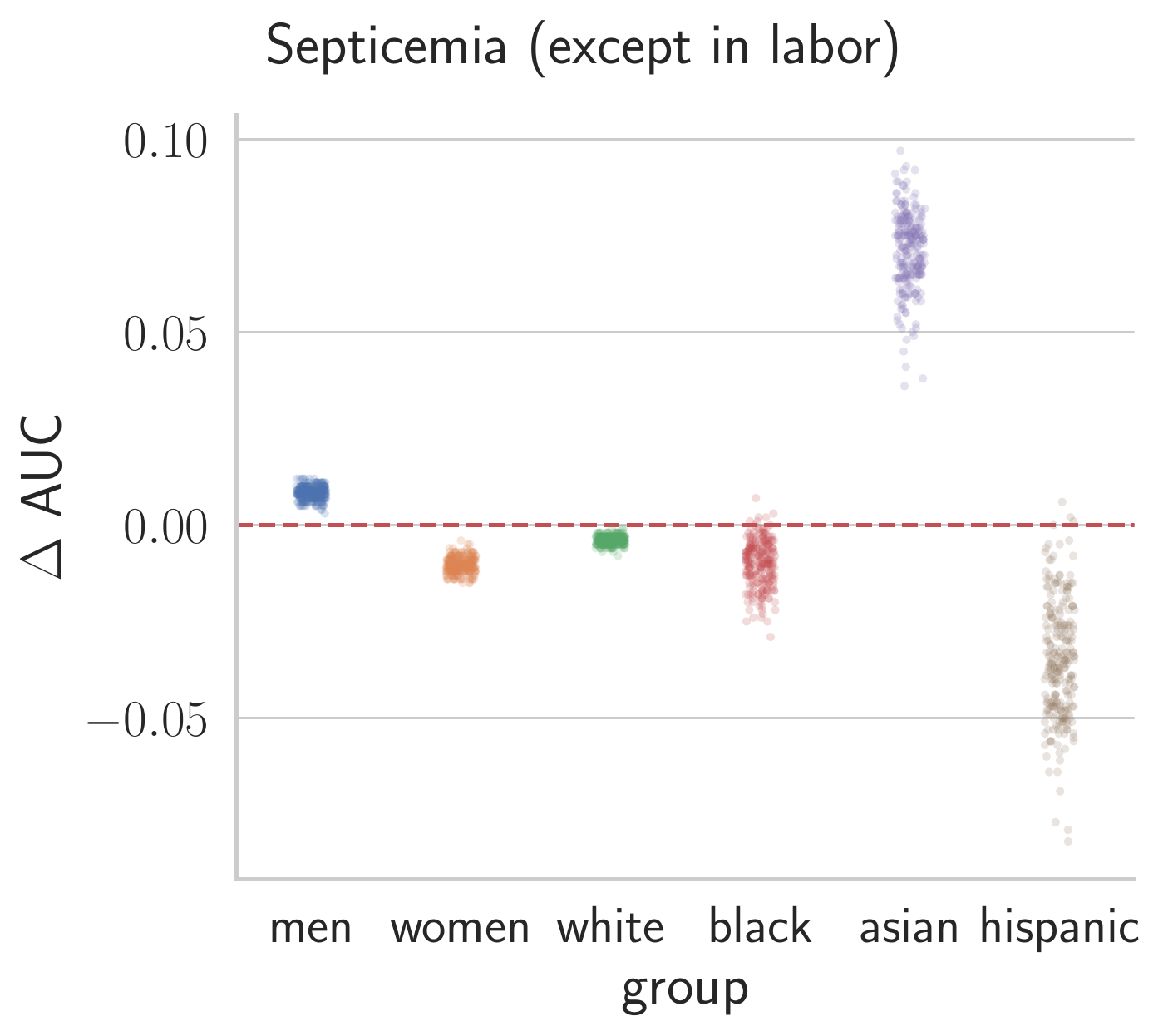}
  \end{minipage}
  \hfill
  \begin{minipage}[t]{0.32\textwidth}
    \includegraphics[width=\textwidth]{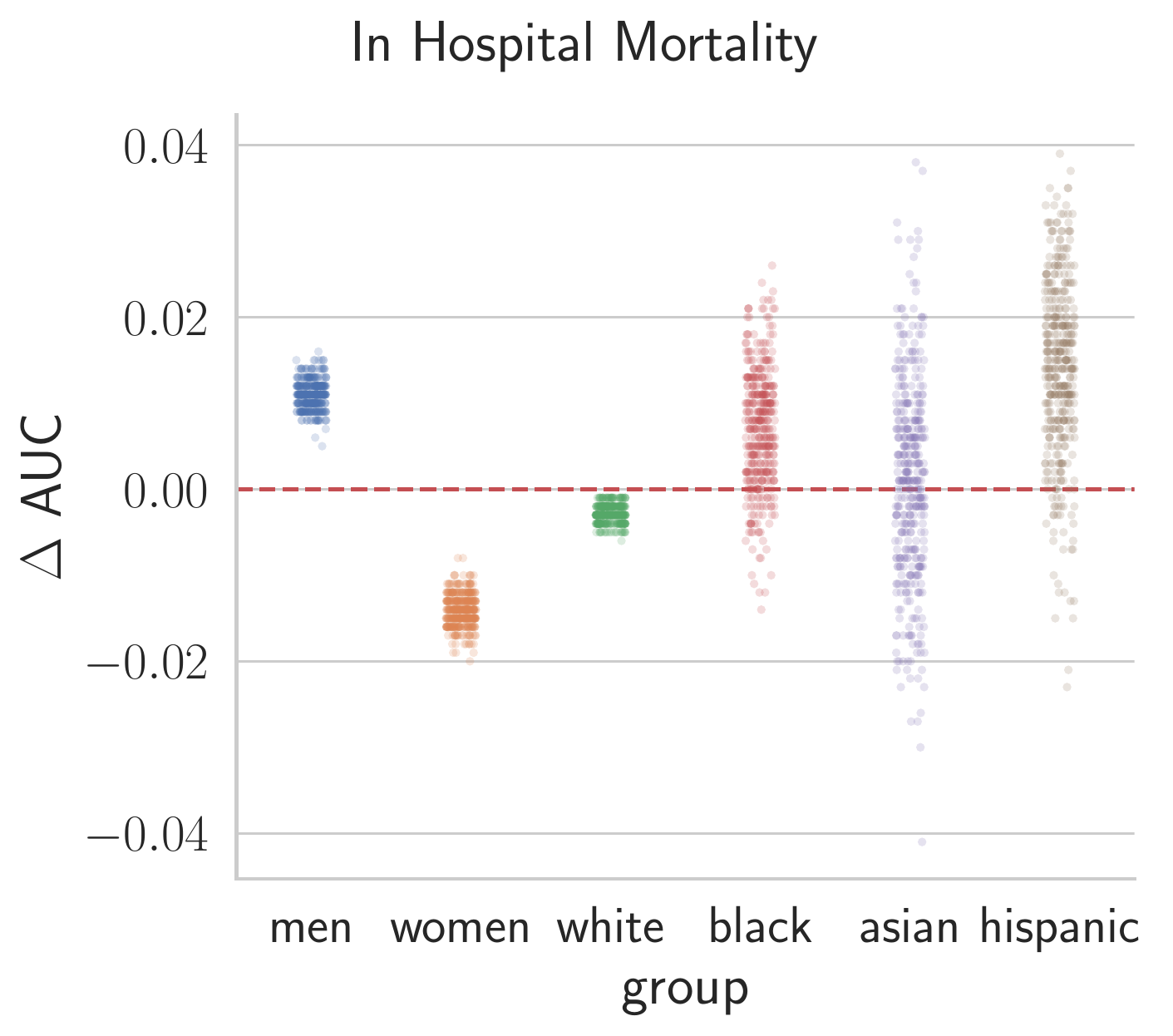}
  \end{minipage}
 \hfill 
 \begin{minipage}[t]{0.42\textwidth}
    \includegraphics[width=\textwidth]{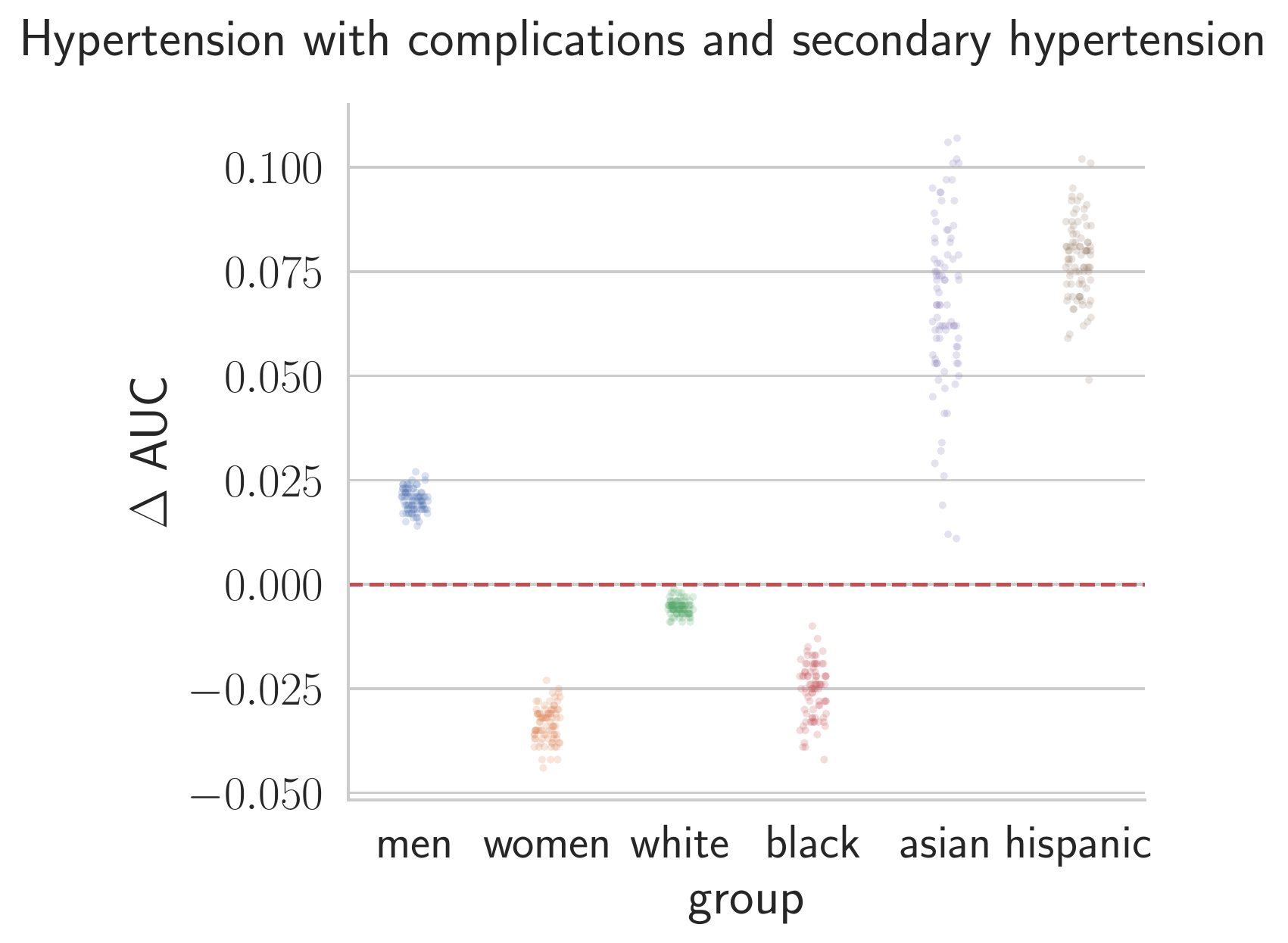}
  \end{minipage}
  \hfill
    \begin{minipage}[t]{0.52\textwidth}
    \includegraphics[width=\textwidth]{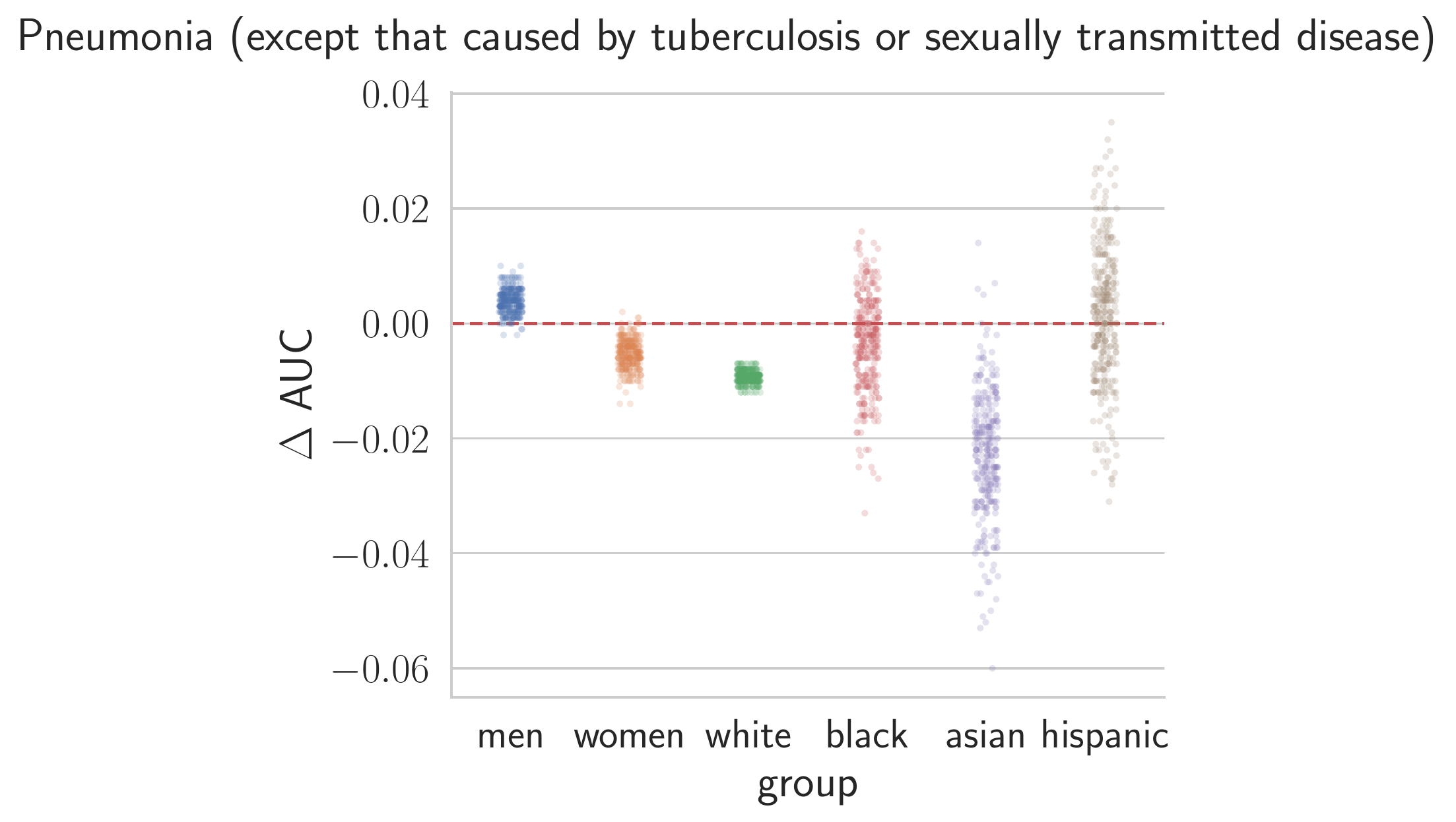}
  \end{minipage}
 \caption{Differences relative to overall performance as a function of random seeds for each subgroup. Each point represents a run for a pair of seeds with validation performance similar to that of the best seeds. }
 \label{fig:underspec_2}
\end{figure*}

Table \ref{tab:tasks} shows all the clinical prediction tasks and the respective prevalence. The following plots present the results for the Phenotyping classification tasks. Figures \ref{fig:deltas_1} and \ref{fig:deltas_2} show the performance $\Delta$s for each subgroup and the overall performance, as a function of the random seeds. Figures \ref{fig:scatters_1} to \ref{fig:scatters_3} show the overall performance against the $\Delta$ for each subgroup. Figures \ref{fig:underspec_1} and \ref{fig:underspec_2} show the subgroup performance $\Delta$s for pairs of seeds with validation performance similar to that of the best seeds.

\begin{table}[htb]
\centering
\scalebox{0.70}{
\begin{tabular}{lp{14em}r}
    \toprule
    {\bf Task} & {\bf Description} & {\bf Prevalence}\\
    \midrule
    IHM & In-Hospital Mortality & $0.13$ \\
    \midrule
    AAURF & Acute and unspecified renal failure & $0.21$ \\
    ACD & Acute cerebrovascular disease & $0.07$ \\
    AMI & Acute myocardial infarction & $0.11$ \\
    CD & Cardiac dysrhythmias & $0.32$ \\
    CKD & Chronic kidney disease & $0.13$ \\
    COPDAB & Chronic obstructive pulmonary disease and bronchiectasis & $0.13$ \\
    COSPOMC & Complications of surgical procedures or medical care & $0.2$ \\
    CD-2 & Conduction disorders & $0.07$ \\
    CHFN & Congestive heart failure; nonhypertensive & $0.28$ \\
    CAAOHD & Coronary atherosclerosis and other heart disease & $0.33$ \\
    DMWC & Diabetes mellitus with complications & $0.1$ \\
    DMWC-2 & Diabetes mellitus without complication & $0.19$ \\
    DOLM & Disorders of lipid metabolism & $0.27$ \\
    EH & Essential hypertension & $0.41$ \\
    FAED & Fluid and electrolyte disorders & $0.25$ \\
    GH & Gastrointestinal hemorrhage & $0.07$ \\
    HWCASH & Hypertension with complications and secondary hypertension & $0.13$ \\
    OLD & Other liver diseases & $0.08$ \\
    OLRD & Other lower respiratory disease & $0.04$ \\
    OURD & Other upper respiratory disease & $0.04$ \\
    PPPC & Pleurisy; pneumothorax; pulmonary collapse & $0.08$ \\
    PTCBTOSTD & Pneumonia (except that caused by tuberculosis or sexually transmitted disease) & $0.14$ \\
    RFIA & Respiratory failure; insufficiency; arrest (adult) & $0.18$ \\
    SIL & Septicemia (except in labor) & $0.14$ \\
    S & Shock & $0.07$ \\
    \bottomrule
    \end{tabular}
    }
    \caption{Clinical prediction tasks along with the respective prevalence}
    \label{tab:tasks}
\end{table}

\begin{figure}
  \centering
  \includegraphics[trim=0 0 0 30, clip, width=0.45\textwidth]{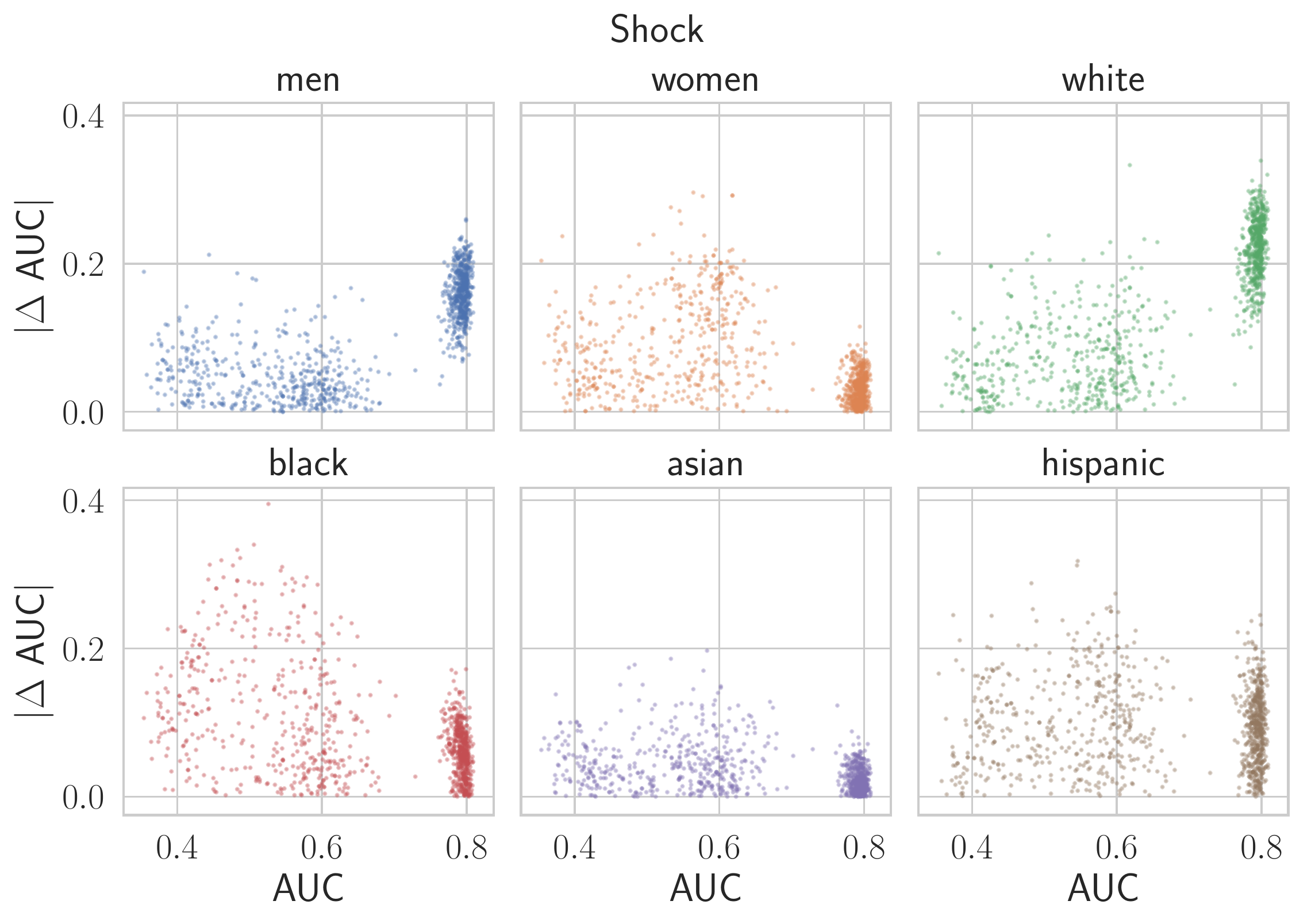}
  \vspace{-0.5\baselineskip}
  \caption{Correlations between overall performance and subgroup performance on the \emph{Shock} phenotype classification task, evaluated on a subset of the test data with equal sample sizes for all demographic subgroups.} 
  \label{fig:subsample_scatters}
\end{figure}

\section{Fine-tuning Experiments}
\label{sec:finetunings}

To illustrate the impact that random seeds can have on measures of algorithmic fairness, we replicate the experiments concerning the fairness of fine-tuned clinical classifiers reported in \cite{zhang2020hurtful}\footnote{see Section 5.2 and Tables 1 and 4.}. 
The study compares 3 measures of multi-group fairness (recall gap, parity gap and specificity gap) with respect to protected attributes such as the gender, language, ethnicity and insurance status. The experiments were conducted over 57 downstream classification tasks: including the IHM and PC tasks we considered in this paper, 3 additional tasks derived from logical ORs on subsets of the PC tasks, and a variation of PC using only the first note. Table 4 in \cite{zhang2020hurtful} reports the number of tasks with statistically significant gaps and the percentage of significant tasks which favor a subgroup.

We use the same experimental setup and implementation\footnote{\url{https://github.com/MLforHealth/HurtfulWords}} to replicate the experiments for IHM and PC using only the first note (29/57 tasks). We repeat each experiment with 100 different random seeds (using the same seed to shuffle the data and initialize parameters) and compute the mean and standard deviation of each measurement across seeds (Table \ref{tab:finetunings}). We find that in general the number of tasks with significant differences is roughly half of those reported by \cite{zhang2020hurtful}, which was expected since we considered half of the tasks. However, we also observe that changes in a single random seed can affect the disparities across protected groups, both in terms of the number and the magnitude of the gaps. We see that there can be variations of up to two tasks with significant gaps and differences of up $31\%$ in the percentage of tasks favoring a specific group.

\begin{table*}[bh!]
\centering
\scalebox{0.68}{
\begin{tabular}{r|r|r|r|r}
\multicolumn{2}{c|}{} & \multicolumn{1}{c|}{\bf Recall Gap} & \multicolumn{1}{c|}{\bf Parity Gap} & \multicolumn{1}{c}{\bf Specificity Gap} \\
 \hline
 {\bf Gender} & Male vs. Female (\% Tasks Favoring Male) & $3.0\pm 1.2$ ($67.8\pm 25.2\%$) & $11.2\pm 2.0$ ($39.0\pm 8.1$\%) & $9.5\pm 2.0$ ($76.8\pm 10.3$\%) \\
\hline {\bf Language} & English vs. Other (\% Tasks Favoring Other)  & $3.1\pm 1.4$ ($50.3\pm 28.9\%$) & $6.9\pm 1.9$ ($5.4\pm 7.0$\%) & $4.2\pm 1.5$ ($88.9\pm 14.8$\%) \\
\hline {\bf Ethnicity} & White vs. Other (\% Tasks Favoring White) & $2.5\pm 1.5$ ($93.0\pm 22.0\%$) & $7.3\pm 1.7$ ($92.6\pm 10.7$\%) & $4.9\pm 1.5$ ($11.1\pm 12.1$\%) \\
& Black vs. Other (\% Tasks Favoring Black) & $3.8\pm 1.5$ ($37.9\pm 21.2\%$) & $6.6\pm 1.7$ ($65.6\pm 15.0$\%) & $3.6\pm 1.4$ ($40.8\pm 22.4$\%) \\
& Hispanic vs. Other (\% Tasks Favoring Hispanic) & $5.1\pm 1.6$ ($8.4\pm 10.9\%$) & $7.6\pm 1.8$ ($0.0\pm 0.0$\%) & $9.3\pm 1.8$ ($99.1\pm 8.9$\%) \\
& Asian vs. Other (\% Tasks Favoring Asian) & $6.1\pm 1.5$ ($53.6\pm 15.8\%$) & $2.3\pm 1.3$ ($77.1\pm 31.0$\%) & $3.6\pm 1.5$ ($54.3\pm 25.0$\%) \\
& Other vs. Other (\% Tasks Favoring Other) & $10.0\pm 1.7$ ($6.0\pm 4.9\%$) & $2.6\pm 1.1$ ($1.0\pm 5.0$\%) & $4.1\pm 1.2$ ($94.5\pm 12.5$\%) \\
\hline {\bf Insurance} & Medicare vs. Other (\% Tasks Favoring Medicare) & $15.0\pm 2.0$ ($93.8\pm 9.5\%$) & $25.7\pm 2.5$ ($92.2\pm 8.6$\%) & $23.9\pm 2.6$ ($2.9\pm 2.6$\%) \\
& Private vs. Other (\% Tasks Favoring Private) & $7.1\pm 1.4$ ($10.5\pm 9.0\%$) & $19.5\pm 2.2$ ($4.2\pm 3.3$\%) & $19.7\pm 2.4$ ($95.5\pm 9.0$\%) \\
& Medicaid vs. Other (\% Tasks Favoring Medicaid) & $9.0\pm 1.7$ ($8.7\pm 7.8\%$) & $17.2\pm 2.1$ ($12.0\pm 3.3$\%) & $15.0\pm 2.1$ ($92.8\pm 9.3$\%) \\
\end{tabular}
}
\caption{We replicated \citet{zhang2020hurtful} analysis of multi-group fairness performance gaps for fine-tuned classifiers across gender, language, ethnicity, and insurance status. We evaluated 28 (out of 57) tasks and repeated the experiments with 100 different random seeds. We measured the average and standard deviation of the number of tasks with statistically significant differences, and the percentage of significant tasks which favor a subgroup.}
\label{tab:finetunings}
\end{table*}




\end{document}